\documentclass[journal]{IEEEtran}
\usepackage{color}
\usepackage{times}
\usepackage{epsfig}
\usepackage{graphicx}
\usepackage{amsmath}
\usepackage{amssymb}
\usepackage{color}
\usepackage{bm}
\usepackage{subfigure}
\usepackage[pagebackref=true,breaklinks=true,letterpaper=true,colorlinks,bookmarks=false]{hyperref}
\usepackage{amsmath,amssymb}
\usepackage{algorithm}
\usepackage{algorithmic}
\usepackage{multirow}
\usepackage{multicol}

\begin{document}

\title{Benchmarking Single Image Dehazing and Beyond}

\author{Boyi Li*, Wenqi Ren*,~\IEEEmembership{Member,~IEEE}, Dengpan Fu*, Dacheng Tao,~\IEEEmembership{Fellow,~IEEE}, Dan Feng,~\IEEEmembership{Member,~IEEE},
        Wenjun Zeng,~\IEEEmembership{Fellow,~IEEE}
        and Zhangyang Wang$^\dag$,~\IEEEmembership{Member,~IEEE}
\thanks{Boyi Li is with the Computer Science Department, Cornell University, USA. Email: boyilics@gmail.com.}
\thanks{Wenqi Ren is with State Key Laboratory of Information Security, Institute of Information Engineering, Chinese Academy of Sciences. Email: rwq.renwenqi@gmail.com.}
\thanks{Dengpan Fu is with Department of Electronic Engineering and Information Science,
University of Science and Technology of China. Email: fdpan@mail.ustc.edu.cn.}
\thanks{Dacheng Tao is with the UBTECH Sydney Artificial Intelligence Centre and the School of Information Technologies, the Faculty of Engineering and Information Technologies, the University of Sydney, 6 Cleveland St, Darlington, NSW 2008, Australia. Email: dacheng.tao@sydney.edu.au.}
\thanks{Dan Feng is with the Wuhan National Laboratory for Optoelectronics, Huazhong University of Science and Technology, Wuhan, China. Email: boyilics@gmail.com, dfeng@hust.edu.cn.}
\thanks{Wenjun Zeng is with Microsoft Research, Beijing, China. Email: wezeng@microsoft.com.}
\thanks{Zhangyang Wang is with the Department of Computer Science and Engineering,
Texas A\&M University, USA. Email: atlaswang@tamu.edu}.
\thanks{The first three authors (Boyi Li, Wenqi Ren, Dengpan Fu) contribute equally. Zhangyang Wang is the main correspondence author. }
\thanks{\textcopyright 2019 IEEE. Personal use of this material is permitted. Permission from IEEE must be obtained for all other uses, in any current or future media, including reprinting/republishing this material for advertising or promotional purposes, creating new collective works, for resale or redistribution to servers or lists, or reuse of any copyrighted component of this work in other works.}}

\markboth{Journal of \LaTeX\ Class Files,~Vol.~14, No.~8, August~2015}%
{Shell \MakeLowercase{\textit{et al.}}: Bare Demo of IEEEtran.cls for IEEE Journals}

\maketitle

\begin{abstract}
We present a comprehensive study and evaluation of existing single image dehazing algorithms, using a new large-scale benchmark consisting of both synthetic and real-world hazy images, called REalistic Single Image DEhazing (RESIDE). RESIDE highlights diverse data sources and image contents, and is divided into five subsets, each serving different training or evaluation purposes. We further provide a rich variety of criteria for dehazing algorithm evaluation, ranging from full-reference metrics, to no-reference metrics, to subjective evaluation and the novel task-driven evaluation. Experiments on RESIDE shed light on the comparisons and limitations of state-of-the-art dehazing algorithms, and suggest promising future directions. 
\end{abstract}


\IEEEpeerreviewmaketitle

\section{Introduction}
\subsection{Problem Description: Single Image Dehazing}
Images captured in outdoor scenes often suffer from poor visibility, reduced contrasts, fainted surfaces and color shift, due to the presence of haze. Caused by aerosols such as dust, mist, and fumes, the existence of haze adds complicated, nonlinear and data-dependent noise to the images, making the haze removal (a.k.a. \textit{dehazing}) a highly challenging image restoration and
enhancement problem. Moreover, many computer vision algorithms can only work well with the scene radiance that is haze-free. However, a dependable vision system must reckon with the entire spectrum of degradations from unconstrained environments. Taking autonomous driving for example, 
hazy and foggy weather
will obscure the vision of on-board cameras and create confusing reflections and glare, leaving state-of-the-art self-driving cars in struggle \cite{Forbes}. Dehazing is thus becoming an increasingly desirable technique for both computational photography
and computer vision tasks, whose advance will immediately benefit many blooming application fields, such as video surveillance and autonomous/assisted driving \cite{tan2008visibility}. 

While some earlier works consider multiple images from the same scene to be available for dehazing~\cite{narasimhan2003contrast,schechner2001instant,treibitz2009polarization,kopf2008deep}, the \textit{single image dehazing} proves to be a more realistic setting in practice, and thus gained the dominant popularity. The \textit{atmospheric scattering model} has been the classical description for the hazy image generation~\cite{mccartney1976optics,narasimhan2000chromatic,narasimhan2002vision}:
\begin{equation}\label{e1}
I\left( x\right) =J\left( x\right) t\left( x\right) +A\left( 1-t\left( x\right) \right) ,
\end{equation}
where $ I\left( x\right)  $ is observed hazy image, $ J\left( x\right)  $ is the haze-free scene radiance
to be recovered. There are two critical parameters: $ A $ denotes the global atmospheric light, and $ t\left( x\right) $ is the transmission matrix defined as:
\begin{equation}\label{e}
t\left( x\right) =e^{-\beta d\left( x\right)},
\end{equation}
where $ \beta $ is the scattering coefficient of the atmosphere, and $ d\left( x\right) $ is the distance between the object and the camera. 

We can re-write the model \eqref{e1} for the clean image as the output:
\begin{equation}\label{e2}
J\left( x\right) =\dfrac{1}{t\left( x\right)} I\left( x\right)-A\dfrac{1}{t\left( x\right)}+A.
\end{equation}
Most state-of-the-art single image dehazing methods exploit the physical model (\ref{e1}), and estimate the key parameters $ A $ and $ t\left( x\right) $ in either physically grounded or data-driven ways. The performance of top methods have continuously improved~\cite{he2011singlecvpr,tarel2009fast,meng2013efficient,chen2016robust,zhu2015fast,berman2016non,jiang2017image,ju2017single}, especially after the latest models embracing deep learning~\cite{cai2016dehazenet,ren2016single,li2017aod}.

\subsection{Existing Methodology: An Overview}
Given the atmospheric scattering model, most dehazing methods follow a similar three-step methodology: (1) estimating the transmission matrix $t\left( x\right)$ from the hazy image $I\left( x\right)$; (2) estimating $A$ using some other (often empirical) methods; (3) estimating the clean image $J\left( x\right)$ via computing \eqref{e2}. 

Usually, the majority of attention is paid to the first step, which can rely on either physically grounded priors or fully data-driven approaches.

A noteworthy portion of dehazing methods exploited natural image priors and depth statistics. \cite{fattal2008single} imposed locally constant constraints of albedo values together with decorrelation of the transmission in local areas, and then estimated the depth value using the albedo estimates and the original image. It did not constrain the scene’s depth structure, thus often leads to the inaccurate estimation of color or depth. 
\cite{he2011single, tang2014investigating} discovered the dark channel prior (DCP) to more reliably calculate the transmission matrix, followed by many successors. However, the prior is found to be unreliable when the scene objects are similar to the atmospheric light \cite{ren2016single}. \cite{meng2013efficient} enforced the boundary constraint and contextual regularization for sharper restorations. \cite{zhu2015fast} developed a color attenuation prior and created a linear model of scene depth for the hazy image, and then learned the model parameters in a supervised way.  \cite{li2015simultaneous} jointly estimated scene depth and recover the clear latent image from a foggy video sequence. \cite{berman2016non} proposed a non-local prior, based on the assumption that each color cluster in the clear image became a haze-line in RGB space.

\begin{figure*}
	\centering
        \includegraphics[width=3.2in,height=2.3in]{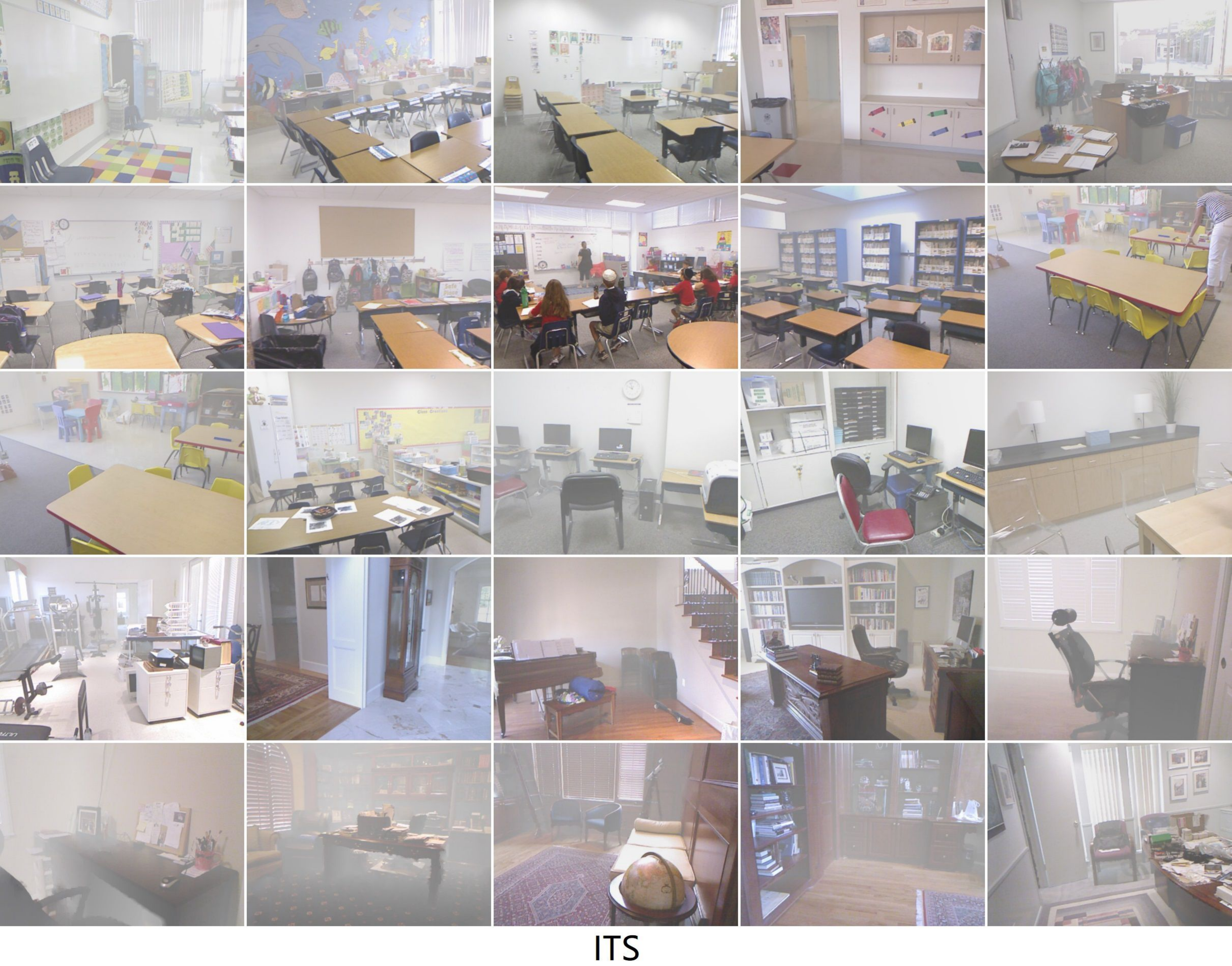} 	
		\includegraphics[width=3.2in,height=2.3in]{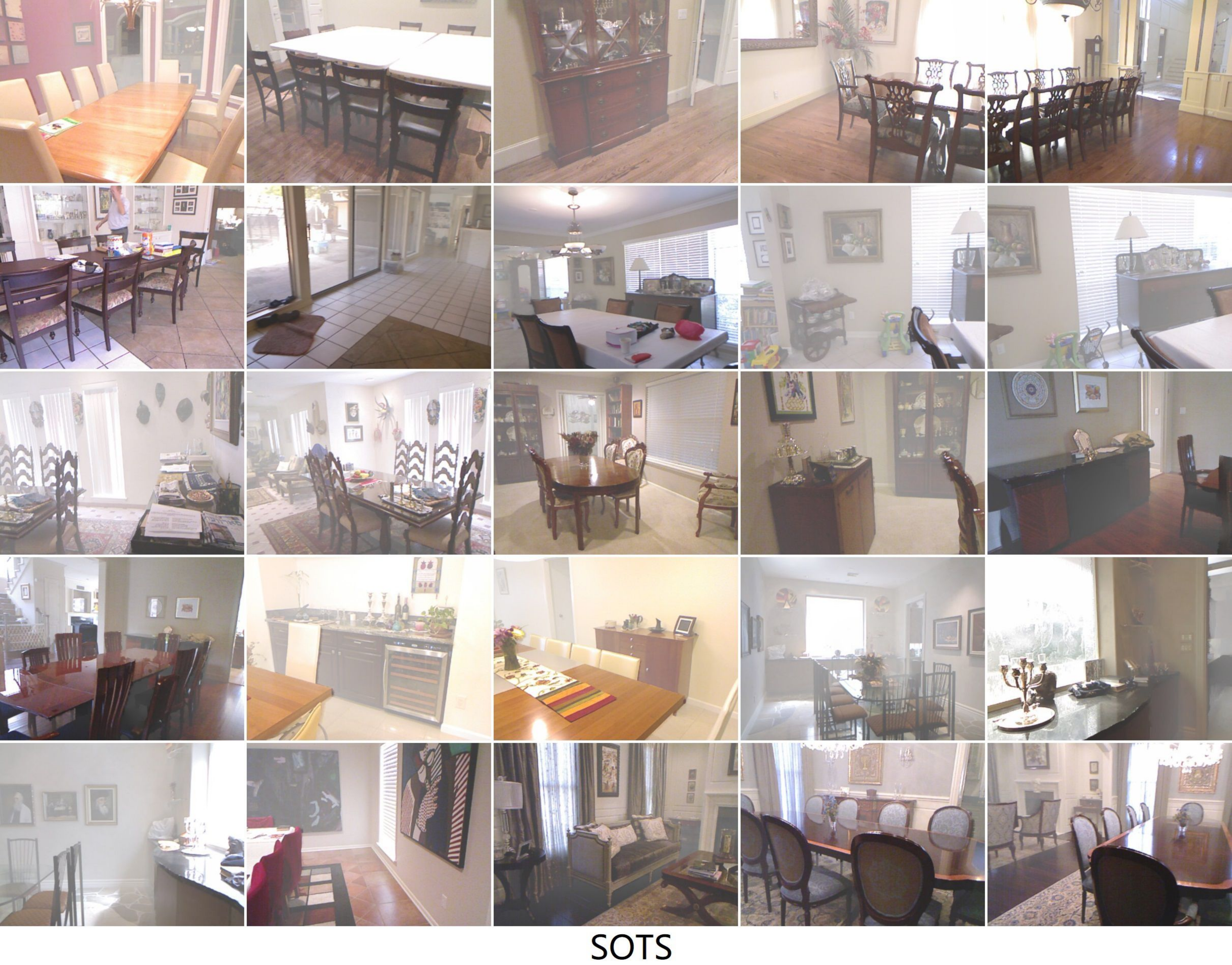}	
    \subfigure[RESIDE.] {
		\includegraphics[width=6.45in,height=1.0in]{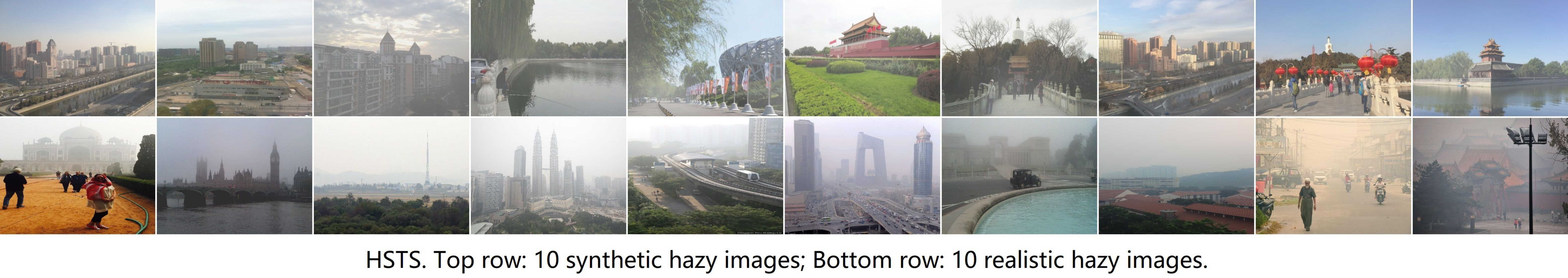}
	}
	\subfigure[RESIDE-$\beta$] {
		\includegraphics[width=3.2in,height=2.3in]{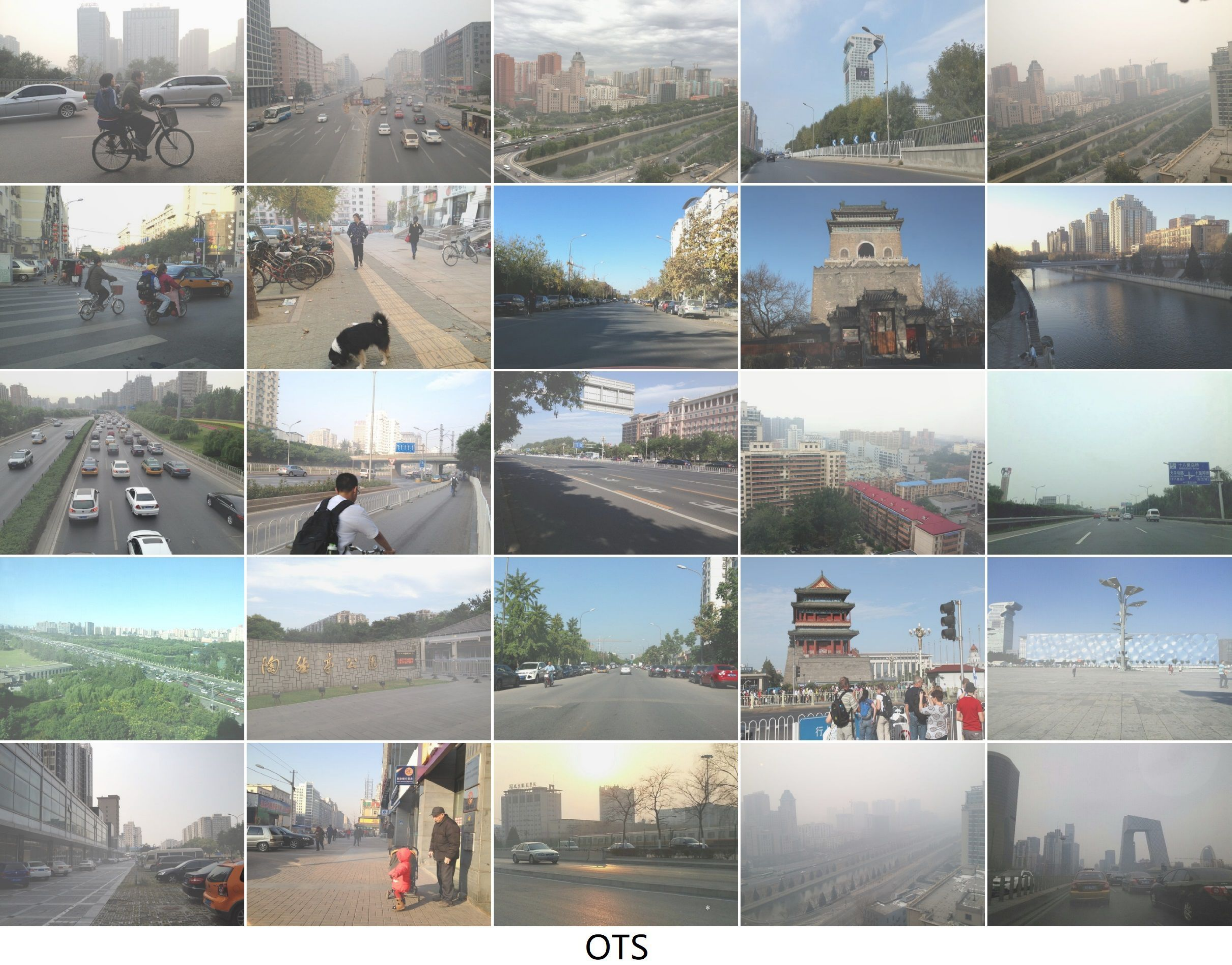}
        \includegraphics[width=3.2in,height=2.3in]{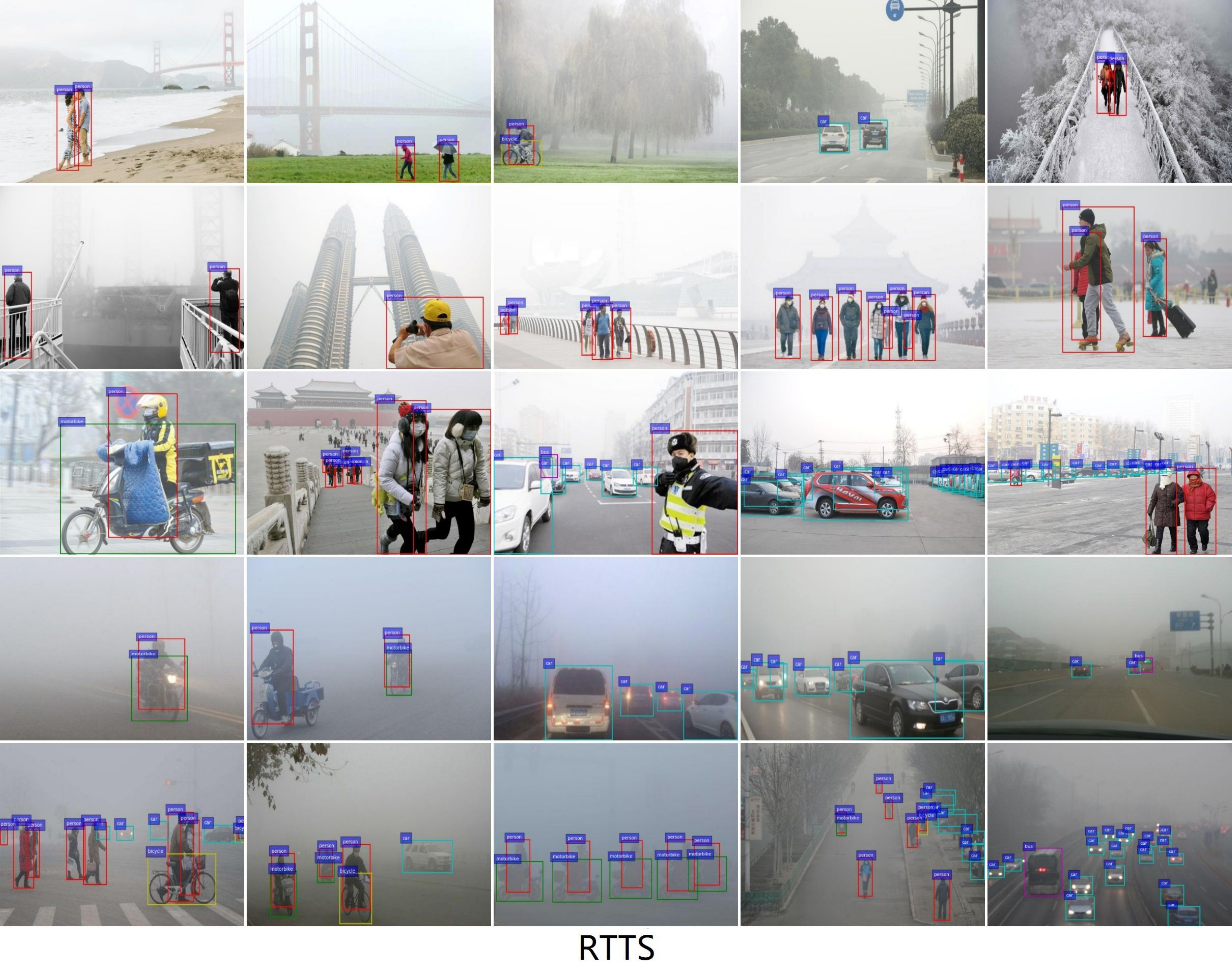}
		\label{example-rtts}
	}
	\caption{Example images from the five sets in RESIDE and RESIDE-$\beta$ (see Table \ref{tab-overview2}.}
	\label{example}
\end{figure*}

In view of the prevailing success of Convolutional Neural Networks (CNNs) in computer vision tasks, several dehazing algorithms have relied on various CNNs to directly learn $t\left( x\right)$ fully from data, in order to avoid the often inaccurate estimation of physical parameters from a single image. DehazeNet~\cite{cai2016dehazenet} proposed a trainable model to estimate the transmission matrix from a hazy image. ~\cite{ren2016single} came up with a multi-scale CNN (MSCNN), that first generated a coarse-scale transmission matrix and gradually refined it. Despite their promising results, \textit{the inherent limitation of training data is becoming a increasingly severe obstacle for this booming trend}: see Section II-1 for more discussions.

Besides, a few efforts have been made beyond the sub-optimal procedure of separately estimating parameters, which will cause accumulated or even amplified errors, when combining them together to calculate \eqref{e2}. They instead advocate simultaneous and unified parameter estimation. Earlier works \cite{kratz2009factorizing,nishino2012bayesian} modeled the hazy image with a factorial Markov random field, where $t\left( x\right)$ and $A$ were two statistically independent latent layers. In addition, some researchers also examined the more challenging night-time dehazing problem \cite{li2015nighttime,zhang2017fast}, which falls beyond the focus of this paper.

Another line of researches~\cite{nair2014effective,zhou2013single} tries to make use of Retinex theory to approximate the spectral properties of object surfaces by the ratio of the reflected light.
Very recently, \cite{li2017aod} presented a re-formulation of \eqref{e} to integrate $t\left( x\right)$ and $A$ into one new variable. As a result. their CNN dehazing model was fully end-to-end: $J\left( x\right)$ was directly generated from $I\left( x\right)$, without any intermediate parameter estimation step. The idea was later extended to video dehazing in \cite{li2017end}.

\subsection{Our Contribution}
Despite the prosperity of single image dehazing algorithms, there have been several hurdles to the further development of this field. There is a lack of benchmarking efforts on state-of-the-art algorithms on a large-scale public dataset. Moreover, current metrics for evaluating and comparing image dehazing algorithms are mostly just PSNR and SSIM, which turn out to be insufficient for characterizing either human perception quality or machine vision effectiveness.

This paper is directly motivated to overcome the above hurdles, and makes three-fold technical contributions:
\begin{itemize}
	\item We introduce a new single image dehazing benchmark, called the \textit{Realistic Single Image Dehazing} (\textbf{RESIDE}) dataset. It features a large-scale synthetic training set, and two different sets designed for objective and subjective quality evaluations, respectively. We further introduce the RESIDE-$\beta$ set, an exploratory and supplementary part of the RESIDE benchmark, including two innovative discussions on the current hurdles on training data content (indoor versus outdoor images) and evaluation criteria (from either human vision or  machine vision perspective), respectively. Particularly in the latter part, we annotate a task-driven evaluation set of 4,322 real-world hazy images with object bounding boxes, which is first-of-its-kind contribution. 
	\item We bring in an innovative set of evaluation strategies in accordance with the new RESIDE and RESIDE-$\beta$ datasets. In RESIDE, besides the widely adopted PSNR and SSIM, we further employ both no-reference metrics and human subjective scores to evaluate the dehazing results, especially for real-world hazy images without clean ground truth. In RESIDE-$\beta$, we recognize that image dehazing in practice usually serves as the preprocessing step for mid-level and high-level vision tasks. We thus propose to exploit the perceptual loss \cite{johnson2016perceptual} as a ``full-reference'' task-driven metric that captures more high-level semantics, and the object detection performance on the dehazed images as a ``no-reference'' task-specific evaluation criterion for dehazing realistic images\cite{li2017aod}.
    \item We conduct an extensive and systematic range of experiments to quantitatively compare nine state-of-the-art single image dehazing algorithms, using the new RESIDE and RESIDE-$\beta$ datasets and the proposed variety of evaluation criteria. Our evaluation and analysis demonstrate the performance and limitations of state-of-the-art algorithms, and bring in rich insights. The findings from these experiments not only confirm what is commonly believed, but also suggest new research directions in single image dehazing.
\end{itemize}
An overview of RESIDE could be found in Table \ref{tab-overview2}.
We note that some of the strategies used in this paper have been previously used in the literature to a greater or smaller extent, such as no-reference metrics in dehazing \cite{ICIP15}, subjective evaluation \cite{sakaridis2017semantic}, and connecting dehazing to high-level tasks \cite{li2017aod}. However, RESIDE is so far the first and only systematic evaluation, that includes a number of dehazing algorithms with multiple criteria on a common large-scale benchmark, which has long been missing from the literature.

The RESIDE dataset is made publicly available for research purposes\footnote{Website: \url{https://sites.google.com/site/boyilics/website-builder/reside}}, and we plan to periodically update our own benchmarking results for noticeable new dehazing algorithms. We also welcome authors to report new results on RESIDE, and to contact us to add their references on the website.

\section{Dataset and Evaluation: Status Quo}
\subsubsection{Training Data}
Many image restoration and enhancement tasks benefit from the continuous efforts for standardized benchmarks to allow for comparison of different proposed methods
under the same conditions, such as \cite{sheikh2006statistical,ntiredata}. In comparison, a common large-scale benchmark has been long missing for dehazing, owing to the significant challenge in collecting or creating realistic hazy images with clean ground truth references. It is generally impossible to capture the same visual scene with and without haze, while all other environment conditions stay identical. Therefore, recent dehazing models \cite{fattal2014dehazing,sakaridis2017semantic} typically generate their training sets by creating synthetic hazy images from clean ones: they first obtain depth maps of the clean images, by either utilizing available depth maps for depth image datasets, or estimating the depth \cite{liu2016learning}; and then generate the hazy images by computing (\ref{e1}). Data-driven dehazing models could then be trained to regress clean images from hazy ones.

\begin{table*}[h]
	\caption{Structure of RESIDE(Standard) and RESIDE-$\beta$}
	\begin{center}{
			\begin{tabular}{c|c|c|c|c}
				\hline
				\multicolumn{5}{c}{\textbf{RESIDE(Standard)}}\\
				\hline
                \hline
				\textit{Subset} & \textit{Number of Images}&\textit{real/synthetic}&\textit{indoor/outdoor}&\textit{annotations}\\
				\hline
				Indoor Training Set (ITS)& 13,990&synthetic&indoor&No\\
				\hline
				Synthetic Objective Testing Set (SOTS)&500&synthetic&indoor&No\\
				\hline
				Hybrid Subjective Testing Set (HSTS)& 20&real&outdoor&No\\
				\hline
  				\hline              
                \multicolumn{5}{c}{RESIDE-$\beta$}\\
                \hline
                \hline
				\textit{Subset} & \textit{Number of Images}&\textit{real/synthetic}&\textit{indoor/outdoor}&\textit{annotations}\\
                \hline
            	Outdoor Training Set (OTS)& 72,135&synthetic&outdoor&No\\
				\hline
                Real-world Task-driven Testing Set (RTTS)& 4,322&real&outdoor&Yes\\
				\hline
		\end{tabular}}
		\label{tab-overview2}
	\end{center}
\end{table*}


Fattal's dataset \cite{fattal2014dehazing} provided 12 synthetic images. FRIDA \cite{tarel2012vision} produced a set of 420 synthetic images, for evaluating the performance of automatic driving systems in various hazy environments. Both of them are too small to train effective dehazing models. To form large-scale training sets, \cite{ren2016single,li2017aod} used the ground-truth images with depth meta-data from the indoor NYU2 Depth Database~\cite{silberman2012indoor} and the Middlebury stereo database \cite{scharstein2003high}. Recently, \cite{sakaridis2017semantic} generated Foggy Cityscapes dataset\cite{cordts2016cityscapes} with 25,000 images from the Cityscapes dataset, using incomplete depth information. 

\subsubsection{Testing Data and Evaluation Criteria}
The testing sets in use are mostly synthetic hazy images with known ground truth too, although some algorithms were also visually evaluated on real hazy images \cite{ren2016single,cai2016dehazenet,li2017aod}.

With multiple dehazing algorithms available, it becomes pivotal to find appropriate evaluation criteria to compare their dehazing results. Most dehazing algorithms rely on the full-reference PSNR and SSIM metrics, with assuming a synthetic testing set with known clean ground truth too. As discussed above, their practical applicability may be in jeopardy even a promising testing performance is achieved, due to the large content divergence between synthetic and real hazy images. To objectively evaluate dehazing algorithms on real hazy images without reference, no-reference image quality assessment (IQA) models \cite{mittal2012no,BLINDS2,SSEQ} are possible candidates. \cite{ICIP15} tested a few no-reference objective IQA models among several dehazing approaches on a self-collected set of 25 hazy images (with no clean ground truth), but did not compare any latest CNN-based dehazing models. A recent work \cite{zhang2017hazerd} collected 14 haze-free images of real outdoor scene and corresponding depth maps, providing a small realistic testing set. 

PSNR/SSIM, as well as other objective metrics, often align poorly with human perceived visual qualities \cite{ICIP15}. Many papers visually display dehazing results, but the result differences between state-of-the-art dehazing algorithms are often too subtle for people to reliably judge. That suggests the necessity of conducting a subjective user study, towards which few efforts have been made so far \cite{chen2014quality,ICIP15}.

All the aforementioned hazy image datasets, as well as RESIDE, are compared in Table \ref{tab-existdata}. As shown, most of the existing datasets are either too small in scale, or lack sufficient real-world images (or annotations) for diverse evaluations.  


\begin{table}[h]
	\caption{Comparison between existing hazy datasets and RESIDE.}
	\begin{center}{
			\begin{tabular}{c|c|c|c|c}
				\hline
				 & \multicolumn{2}{|c|}{Synthetic} & \multicolumn{2}{|c}{Real}  \\
                \hline
                 & indoor & outdoor & outdoor & annotated \\
                \hline
				Fattal \cite{fattal2014dehazing} & 4 & 8 & 31 & -\\
				\hline
				FIRDA \cite{tarel2012vision} & - & 480 & - & - \\
				\hline
				Ma \cite{ICIP15} & 3 & 22 & - & - \\
				\hline
                HazeRD \cite{zhang2017hazerd} & - & 14 & - & - \\
                \hline
                Sakaridis \cite{sakaridis2017semantic} & - & 25,000 & 101 & 101 \\
                \hline
				RESIDE & 14,490 & 72,135 & 9,129 & 4,322\\
				\hline
		\end{tabular}}
		\label{tab-existdata}
	\end{center}
\end{table}

\section{A New Large-Scale Dataset: RESIDE}
We propose the \textit{REalistic Single Image DEhazing} (\textbf{RESIDE}) dataset, a new large-scale dataset for fairly evaluating and comparing single image dehazing algorithms. A distinguishing feature of RESIDE lies in the diversity of its evaluation criterion, ranging from traditional full-reference metrics, to more practical no-reference metrics, and to the desired human subjective ratings. A novel set of task-driven evaluation options will be discussed later in this paper.

\subsection{Dataset Overview}
The REISDE training set contains 13, 990 synthetic hazy images, generated using 1, 399 clear images from existing indoor depth datasets NYU2~\cite{silberman2012indoor} and Middlebury stereo \cite{scharstein2003high}. We synthesize 10 hazy images for each clear image. An optional split of 13, 000 for training and 990 for validation is provided. We set different atmospheric lights $A$, by choosing each channel uniformly randomly between $[0.7,1.0]$, and select $\beta$ uniformly at random between $[0.6,1.8]$. It thus contains paired clean and hazy images, where a clean ground truth image can lead to multiple pairs whose hazy images are generated under different parameters $A$ and $\beta$.

The REISDE testing set is composed of \textit{Synthetic Objective Testing Set} (SOTS) and the \textit{Hybrid Subjective Testing Set} (HSTS), designed to manifest a diversity of evaluation viewpoints. SOTS selects 500 indoor images from NYU2~\cite{silberman2012indoor} (non-overlapping with training images), and follow the same process as training data to synthesize hazy images. We specially create challenging dehazing cases for testing, e.g., white scenes added with heavy haze. HSTS picks 10 synthetic outdoor hazy images generated in the same way as SOTS, together with 10 real-world hazy images collected real world outdoor scenes \cite{BeijingImages}\footnote{Image Source: \url{http://www.tour-beijing.com/real_time_weather_photo/}}, combined for human subjective review.

\subsection{Evaluation Strategies}
\subsubsection{From Full-Reference to No-Reference}
Despite the popularity of the full-reference PSNR/SSIM metrics for evaluating dehazing algorithms, they are inherently limited due to the unavailability of clean ground truth images in practice, as well as their often poor alignment with human perception quality \cite{ICIP15}. We thus refer to two no-reference IQA models: spatial-spectral entropy-based quality (SSEQ) \cite{SSEQ}, and blind image integrity notator using DCT statistics (BLIINDS-II) \cite{BLINDS2}, to complement the shortness of PSNR/SSIM. 
Note that the score of SSEQ and BLIINDS2 used in~\cite{SSEQ} and \cite{BLINDS2} are range from 0 (best) to 100 (worst), and we reverse the score to make the correlation consistent to full-reference metrics.

We will apply PSNR, SSIM, SSEQ, and BLIINDS-II, to the dehazed results on SOTS, and examine how consistent their resulting ranking of dehazing algorithms will be. We will also apply the four metrics on HSTS (PSNR and SSIM are only computed on the 10 synthetic images), and further compare those objective measures with subjective ratings. 

\subsubsection{From Objective to Subjective}
\cite{ICIP15} investigated various choices of full-reference and no-reference IQA models, and found them to be limited in predicting the quality of dehazed images. We then conduct a subjective user study on the quality of dehazing results produced by different algorithms, from which we gain more useful observations. Ground-truth images are also included when they are available as references. 

In the previous survey \cite{ICIP15,li2017haze} a participant scored each dehazing result image with an integer from 1 to 10 that best reflects its perceptual quality. We adopt a different pipeline: (1) asking participants to give pairwise comparisons rather than individual ratings, the former often believed to be more robust and consistent in subjective surveys, which has also be adopted by \cite{wang2015learning,sakaridis2017semantic}; (2) decomposing the perceptual quality into two dimensions: the dehazing \textit{Clearness} and \textit{Authenticity}, the former defined as how thoroughly the haze has been removed, and the latter defined as how realistic the dehazed image looks like. Up to our best knowledge, such two disentangled dimensions have not been explored before in similar literature. They are motivated by our observations that some algorithms produce naturally-looking results but are unable to fully remove haze, while some others remove the haze at the price of unrealistic visual artifacts. 

During the survey, each participant is shown a set of dehazed result pairs obtained using two different algorithms for the same hazy image. For each pair, a participant needs to independently decide which one is better than the other in terms of Clearness, and then which one is better for Authenticity. The image pairs are drawn from all the competitive methods randomly, and the images winning the pairwise comparison will be compared again in the next round \cite{pulkkinen1994bayesian}, until the best one is selected. We fit a Bradley-Terry \cite{bradley1952rank} model to estimate the subjective scores for each dehazing algorithm so that they can be ranked.

As the same for peer benchmarks \cite{yang2014single,lai2016comparative}, the subjective survey is not ``automatically'' scalable to new results. However, it is extremely important to study the correlation between human perception and objective metrics, which helps analyze the effectiveness of the latter. We are preparing to launch a leaderboard, where we will accept selective result submissions, and periodically run new subjective reviews.
\begin{table*}[!ht]
	\caption{Average full- and no-reference evaluations results of dehazed results on SOTS.}	
	\begin{center}\footnotesize{
			\begin{tabular}{c|c|c|c|c|c|c|c|c|c}
				\hline
				& DCP~\cite{he2011singlecvpr}  & FVR~\cite{tarel2009fast} & BCCR~\cite{meng2013efficient} & GRM~\cite{chen2016robust} & CAP~\cite{zhu2015fast} & NLD~\cite{berman2016non} & DehazeNet~\cite{cai2016dehazenet} & MSCNN~\cite{ren2016single} & AOD-Net~\cite{li2017aod}\\
				\hline
				PSNR & 16.62  & 15.72  & 16.88  & 18.86  &  \textcolor{blue}{19.05}  & 17.29  & \textcolor{red}{21.14}  &  17.57  & \textcolor{cyan}{19.06}  \\
				\hline
				SSIM &  0.8179 & 0.7483   &  0.7913  & \textcolor{red}{0.8553}  & 0.8364 &  0.7489 &  \textcolor{blue}{0.8472}  &  0.8102  &  \textcolor{cyan}{0.8504} \\
				\hline
				SSEQ &  64.94  &  \textcolor{red}{67.75}   &  65.83  &  63.30   &  64.69   &  \textcolor{blue}{67.46} &  65.46 & 65.31   &  \textcolor{cyan}{67.65}  \\
				\hline
				BLIINDS-II & 74.41  &  \textcolor{cyan}{75.63}  &   74.45 & 73.46 & 73.41  & \textcolor{blue}{74.85} & 71.71  &  74.34  &   \textcolor{red}{79.02} \\
			\hline
		\end{tabular}}
		\label{tab-full-subjective}
	\end{center}
\end{table*}
\begin{table*}[!ht]
	\caption{Average full-evaluations results of dehazed results on SOTS with different haze level.}	
	\begin{center}\footnotesize{
			\begin{tabular}{c|c|c|c|c|c|c|c|c|c}
                \hline
				& DCP~\cite{he2011singlecvpr}  & FVR~\cite{tarel2009fast} & BCCR~\cite{meng2013efficient} & GRM~\cite{chen2016robust} & CAP~\cite{zhu2015fast} & NLD~\cite{berman2016non} & DehazeNet~\cite{cai2016dehazenet} & MSCNN~\cite{ren2016single} & AOD-Net~\cite{li2017aod}\\
				\hline
                \multicolumn{10}{c}{$\beta\in[0.6,0.9]$}\\
                \hline
				PSNR & 16.10  & 17.18  & 16.91  & 18.64  &  \textcolor{blue}{20.88}  & 17.52  & \textcolor{red}{24.24}  &  19.72  & \textcolor{cyan}{22.40}  \\
				\hline
				SSIM &  0.8158 & 0.7682   &  0.7978  & 0.8528  & \textcolor{blue}{0.8597} &  0.7558 &  \textcolor{red}{0.9044}  &  0.8489  & \textcolor{cyan}{0.8980} \\
				\hline
                \multicolumn{10}{c}{$\beta\in[1.0,1.4]$}\\
                \hline
				PSNR & 16.58  & 16.00  & 17.07  & 18.74  &  \textcolor{cyan}{19.68}  & 17.37  & \textcolor{red}{22.02}  &  17.25  & \textcolor{blue}{19.61}  \\
				\hline
				SSIM &  0.8210 & 0.7538   &  0.7942  & \textcolor{blue}{0.8576}  & 0.8450 &  0.7487 &  \textcolor{red}{0.8870}  &  0.8110  &  \textcolor{cyan}{0.8616} \\
			\hline
                \multicolumn{10}{c}{$\beta\in[1.5,1.8]$}\\
                \hline
				PSNR & 17.15  & 14.42  & 17.14  & \textcolor{red}{19.11}  &  \textcolor{blue}{17.21}  & 17.06  & \textcolor{cyan}{18.67}  &  15.10  & 16.16  \\
				\hline
				SSIM &  \textcolor{blue}{0.8259} & 0.7289   &  0.7906  & \textcolor{red}{0.8555}  & 0.8120 &  0.7438 & \textcolor{cyan}{0.8454}  &  0.7723  & 0.8064\\
                \hline
		\end{tabular}}
		\label{tab-full-subjective-vary}
	\end{center}
\end{table*}

\begin{table*}[h]
	\caption{Average subjective scores, as well as full- and no-reference evaluations results, of dehazing results on HSTS.}
	\begin{center}\footnotesize{
			\begin{tabular}{c|c|c|c|c|c|c|c|c|c}
				\hline
				& DCP~\cite{he2011singlecvpr}  & FVR~\cite{tarel2009fast} & BCCR~\cite{meng2013efficient} & GRM~\cite{chen2016robust} & CAP~\cite{zhu2015fast} & NLD~\cite{berman2016non}& DehazeNet~\cite{cai2016dehazenet}& MSCNN~\cite{ren2016single} & AOD-Net~\cite{li2017aod}\\
				\hline
				&	\multicolumn{9}{c}{Synthetic images} \\
				\hline
				\textbf{Clearness} & \textcolor{red}{1.26} &  0.18 &  0.62 & 0.75 & 0.50  & \textcolor{blue}{1}  & 0.29 & \textcolor{cyan}{1.22}  & 0.86\\
				\hline
				\textbf{Authenticity} &  0.78 &  0.14 & 0.50 & 0.95 & 0.86  & \textcolor{blue}{1}  & \textcolor{red}{1.94} &  0.54 &\textcolor{cyan}{1.41} \\
				\hline
				PSNR & 14.84 & 14.48 & 15.08 & 18.54 & \textcolor{blue}{21.53} & 18.92 & \textcolor{red}{24.48} & 18.64 & \textcolor{cyan}{20.55}\\
				\hline
				SSIM & 0.7609  & 0.7624 & 0.7382 & 0.8184 & \textcolor{blue}{0.8726} & 0.7411 & \textcolor{red}{0.9153} & 0.8168 & \textcolor{cyan}{0.8973} \\
				\hline
				SSEQ &  \textcolor{blue}{86.15}  & 85.68  & 85.60  & 78.43  & 85.32  & \textcolor{cyan}{86.28}  & 86.01  & 85.56  & \textcolor{red}{86.75}  \\
				\hline
				BLIINDS-II &  \textcolor{cyan}{90.70}  & 87.65  & \textcolor{red}{91.05}  & 82.30  & 85.75  & 85.30  & 87.15  & \textcolor{blue}{88.70}  & 87.50 \\
				\hline
				&	\multicolumn{9}{c}{Real-world images} \\
				\hline
				\textbf{Clearness} & 0.39 &  0.46 &  0.45 & 0.75 & 1  & 0.54  & \textcolor{cyan}{1.16} & \textcolor{red}{1.29}  & \textcolor{blue}{1.05}\\
				\hline
				\textbf{Authenticity} &  0.17 &  0.20 & 0.18 & 0.62 & 1  & 0.15  & \textcolor{blue}{1.03} &  \textcolor{red}{1.27} &\textcolor{cyan}{1.07} \\
				\hline
				SSEQ &  \textcolor{blue}{68.65}   & 67.75 &  66.63 & \textcolor{red}{70.19}  &  67.67  & 67.96  &  68.34 & 68.44   &  \textcolor{cyan}{70.05}  \\
				\hline
				BLIINDS-II &  69.35  &  \textcolor{blue}{72.10}   &  68.55  & \textcolor{red}{79.60}  &  63.55 & 70.80  & 60.35  & 62.65  & \textcolor{cyan}{74.75} \\
				\hline
		\end{tabular}}
		\label{tab-clearness_authenticity}
	\end{center}
\end{table*}
%


\section{Algorithm Benchmarking}
Based on the rich resources provided by RESIDE, we evaluate 9 representative state-of-the-art algorithms:
Dark-Channel Prior (DCP)~\cite{he2011singlecvpr},
Fast Visibility Restoration (FVR)~\cite{tarel2009fast},
Boundary Constrained Context Regularization (BCCR)~\cite{meng2013efficient},
Artifact Suppression via Gradient Residual Minimization (GRM)~\cite{chen2016robust},
Color Attenuation Prior (CAP)~\cite{zhu2015fast},
Non-local Image Dehazing (NLD)~\cite{berman2016non},
DehazeNet~\cite{cai2016dehazenet},
Multi-scale CNN (MSCNN)~\cite{ren2016single},
and All-in-One Dehazing Network (AOD-Net)~\cite{li2017aod}. The last three belong to the latest CNN-based dehazing algorithms. For all data-driven algorithms, they are trained on the same RESIDE training set.

\subsection{Objective Comparison on SOTS}

We first compare the dehazed results on SOTS using two full-reference (PSNR, SSIM) and two no-reference metrics (SSEQ, BLIINDS-II). Table~\ref{tab-full-subjective} displays the detailed scores of each algorithm in terms of each metric.\footnote{We highlight the top-3 performances using red, cyan and blue, respectively.}

In general, since learning-based methods~\cite{cai2016dehazenet,zhu2015fast,ren2016single,li2017aod} are optimized by directly minimizing the mean-square-error (MSE) loss between output and ground truth pairs or maximizing
the likelihood on large-scale data, they clearly outperform earlier algorithms based on natural or statistical priors~\cite{he2011singlecvpr,meng2013efficient,tarel2009fast,chen2016robust,berman2016non} in most cases, in terms of PSNR and SSIM.  
Especially, 
DehazeNet~\cite{cai2016dehazenet} achieves the highest PSNR value, AOD-Net~\cite{li2017aod} and CAP \cite{zhu2015fast} obtain the suboptimal and third PSNR score. 
Although GRM \cite{chen2016robust} achieves the highest SSIM score,  AOD-Net~\cite{li2017aod} and DehazeNet~\cite{cai2016dehazenet} still obtain the similar SSIM values.

However, when it comes to no-reference metrics, the results become less consistent. AOD-Net~\cite{li2017aod} still maintains competitive performance by obtaining the best BLIINDS-II result on indoor images, thanks to end-to-end pixel correction.
On the other hand, several prior-based methods, such as 
FVR~\cite{tarel2009fast}
and NLD~\cite{berman2016non} also show competitiveness: FVR~\cite{tarel2009fast} ranks first in term of SSEQ,
and NLD~\cite{berman2016non} achieves the suboptimal SSEQ and BLIINDS-II.
We visually observe the results, and find that DCP \cite{he2011singlecvpr}, BCCR~\cite{meng2013efficient} and NLD~\cite{berman2016non} tend to produce sharp edges and highly contrasting colors, which explains why they are preferred by BLIINDS-II and SSEQ. Such an inconsistency between full- and no-reference evaluations aligns with the previous argument \cite{ICIP15} that existing objective IQA models are very limited in providing proper quality predictions of dehazed images. 

We have further conducted an experiment using standard evaluation metrics, with different haze concentration levels (i.e.,$\beta$ values), to detail the suitability of each method for each distinct haze density. As show in Table \ref{tab-full-subjective-vary}, we split the SOTS dataset into three groups according to the ranges of $\beta$. It makes clear that DehazeNet is consistently the best for light and medium haze, and GRM achieves the highest PSNR and SSIM  for thick haze.

\subsection{Subjective Comparison on HSTS}

We recruit 100 participants from different educational backgrounds for the subjective, using HSTS which contains 10 synthetic outdoor and 10 real-world hazy images. We fit a Bradley-Terry \cite{bradley1952rank} model to estimate
the subjective score for each method so that they can be ranked.
In the Bradley-Terry model, the probability that an object $X$ is favored over $Y$ is assumed to be
\begin{equation}
\label{eq:bt}
    p(X \succ Y) = \frac{e^{s_X}}{e^{s_X}+e^{s_Y}} = \frac{1}{1+e^{s_Y-s_X}} ,
\end{equation}
where $s_X$ and $s_Y$ are the subjective scores for $X$ and $Y$.
The scores $\mathbf{s}$ for all the objects can be jointly estimated
by maximizing the log likelihood of the pairwise comparison observations:
\begin{equation}
\label{eq:bt_mle}
    \max_{\mathbf{s}} \sum_{i, j} w_{ij}\log \left(\frac{1}{1+e^{s_j-s_i}}\right) ,
\end{equation}
where $w_{ij}$ is the $(i, j)$-th element in the winning matrix $\mathbf{W}$,
representing the number of times when method $i$ is favored over method $j$.
We use the Newton-Raphson method to solve Eq.~\eqref{eq:bt_mle}. 
Note that for a synthetic image, we have a $10 \times 10$ winning matrix $\mathbf{W}$, including the ground truth and nine dehazing methods’ results. For a real-world image, its winning matrix $\mathbf{W}$ is $9 \times 9$ due to the absence of ground truth. For synthetic images, we set the score for ground truth method as 1 to normalization scores.

\begin{figure*}
	\centering
    \includegraphics[width=3.2in,height=1.7in]{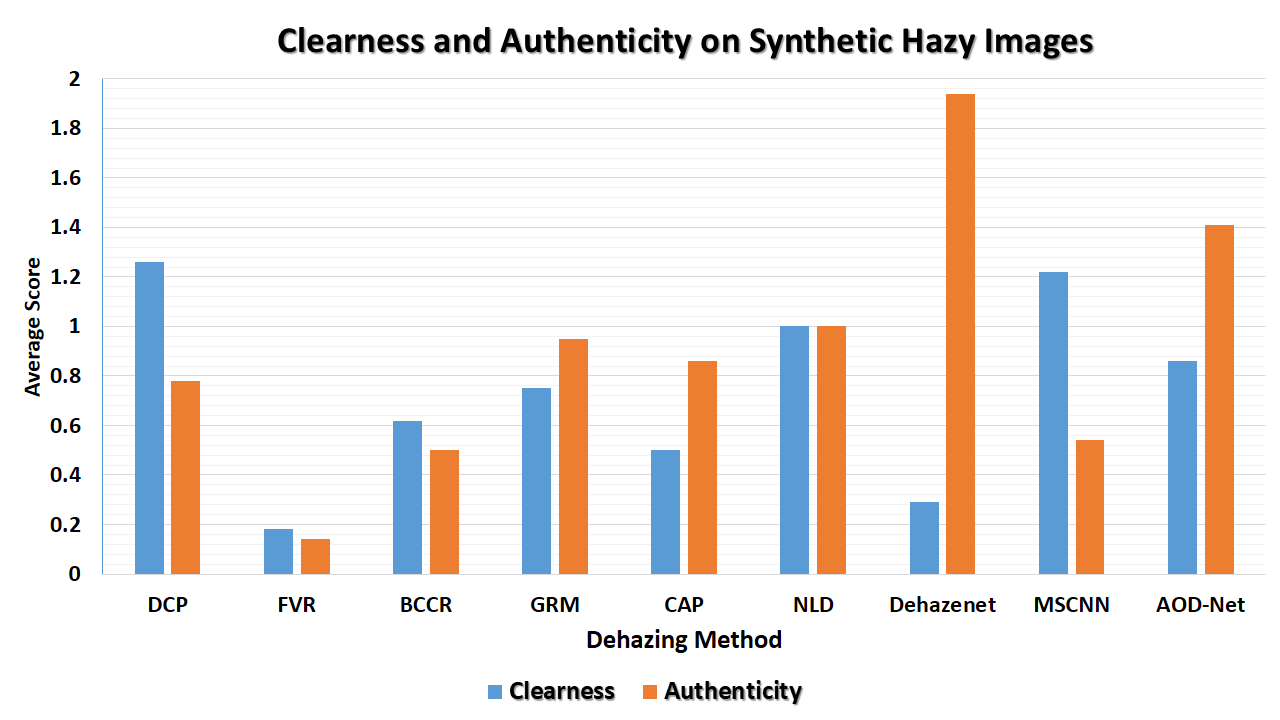}
	\includegraphics[width=3.2in,height=1.7in]{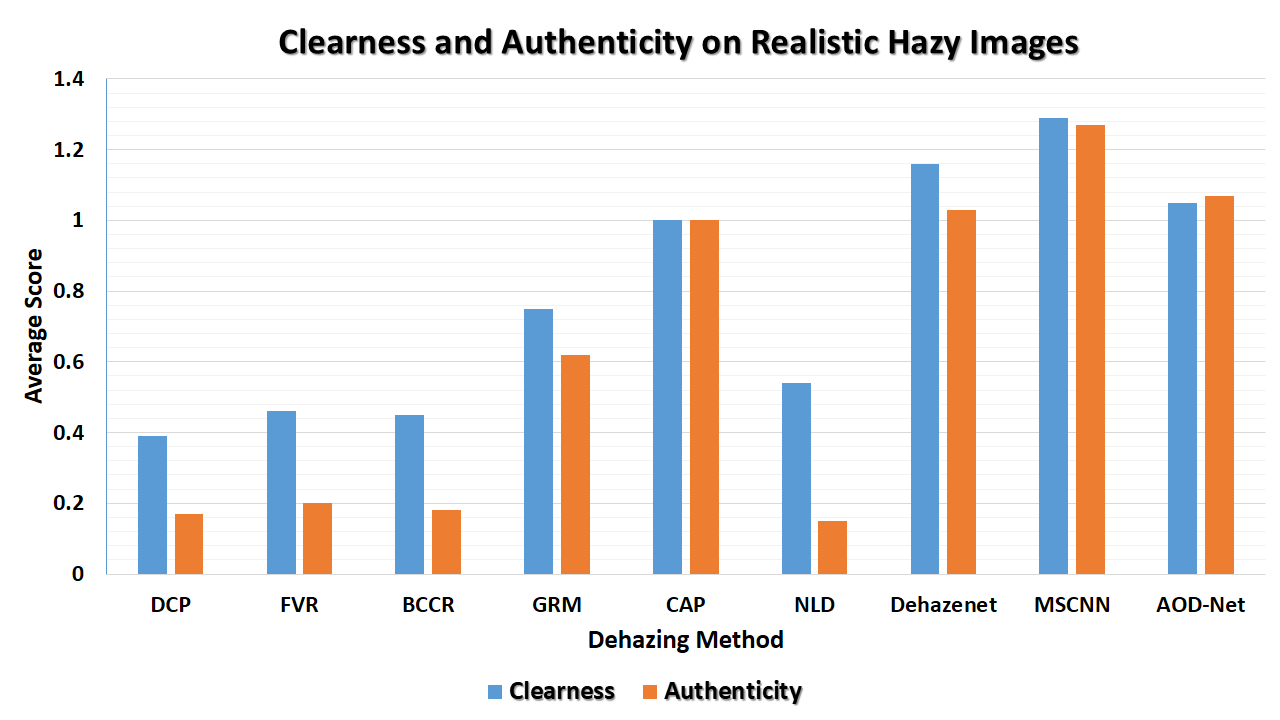}
	\caption{Averaged clearness and authenticity scores: (a) on 10 synthetic images in HSTS; and (b) on real-world images in HSTS.}
	\label{fig:realsyn_clearness_authenticity}
\end{figure*}

\begin{figure*}[t]
	\centering
	\centering
    \begin{minipage}{0.15\textwidth}
    	\centering
        \subfigure[Clean Image] {
        	\includegraphics[width=1.0\textwidth,height=1.5in]{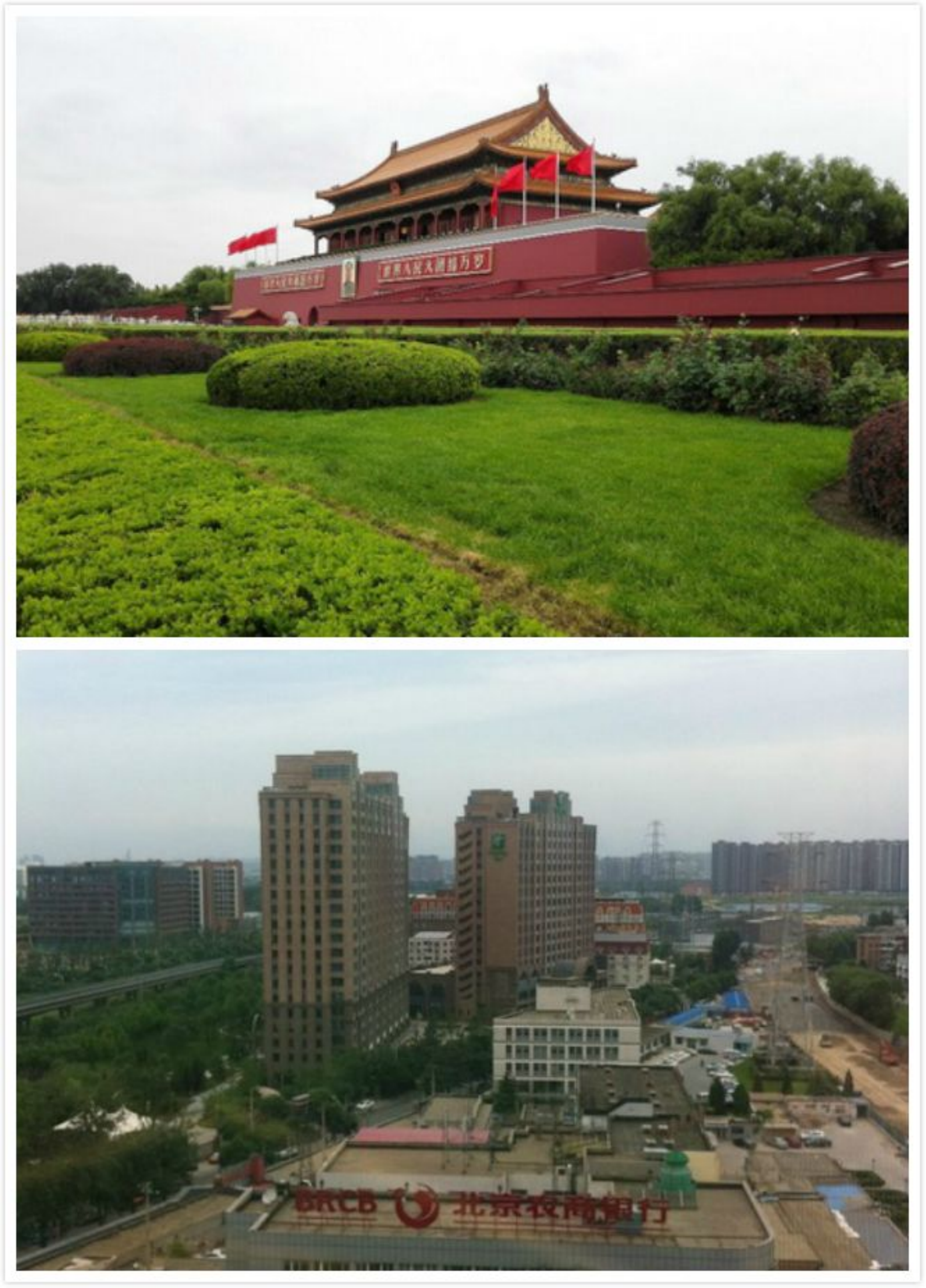}}     
    \end{minipage}
	\begin{minipage}{0.15\textwidth}
		\centering \subfigure[Hazy Image] {
			\includegraphics[width=\textwidth,height=1.5in]{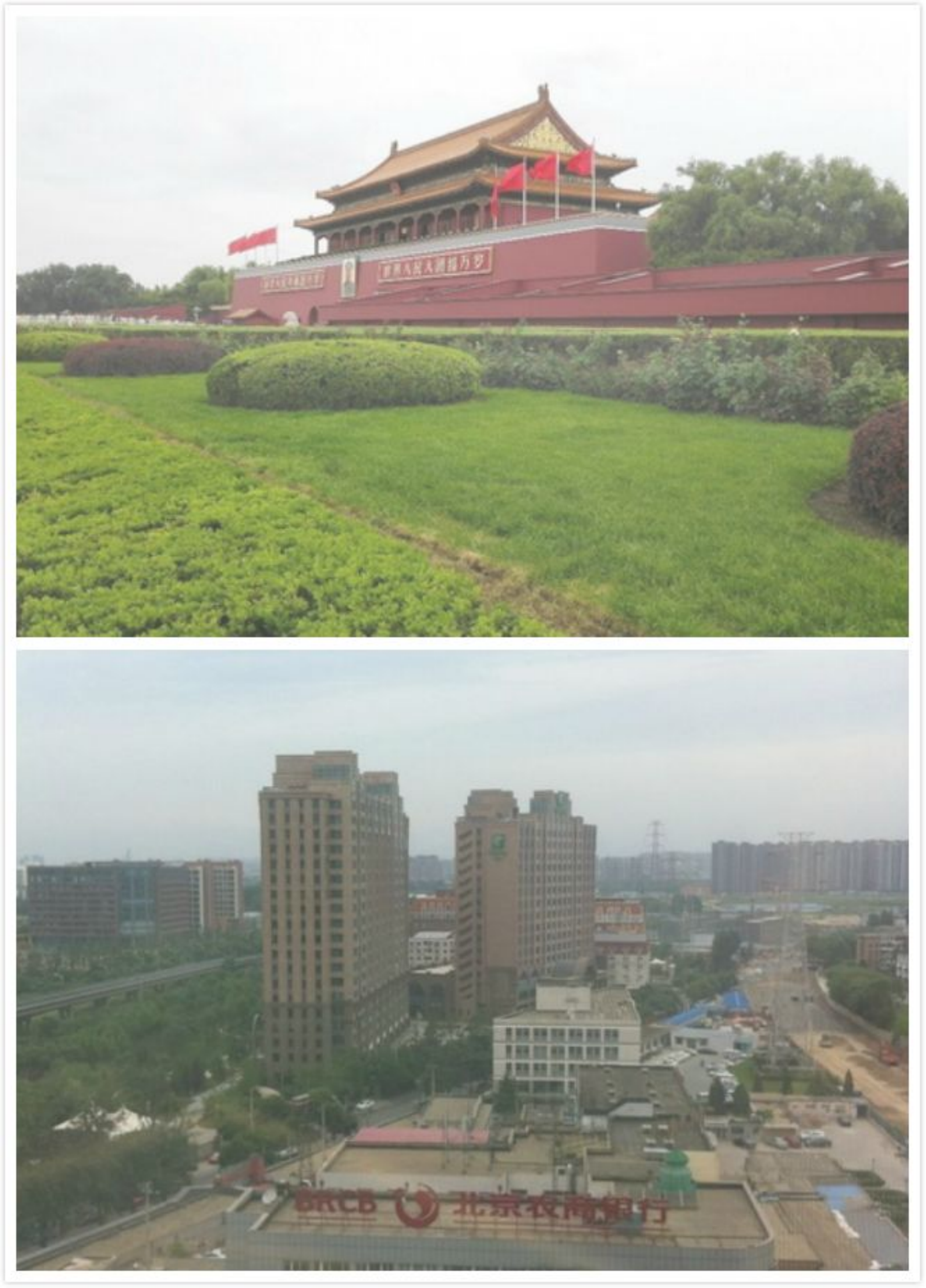}		}\\
		\subfigure[DCP] {
			\includegraphics[width=\textwidth,height=1.5in]{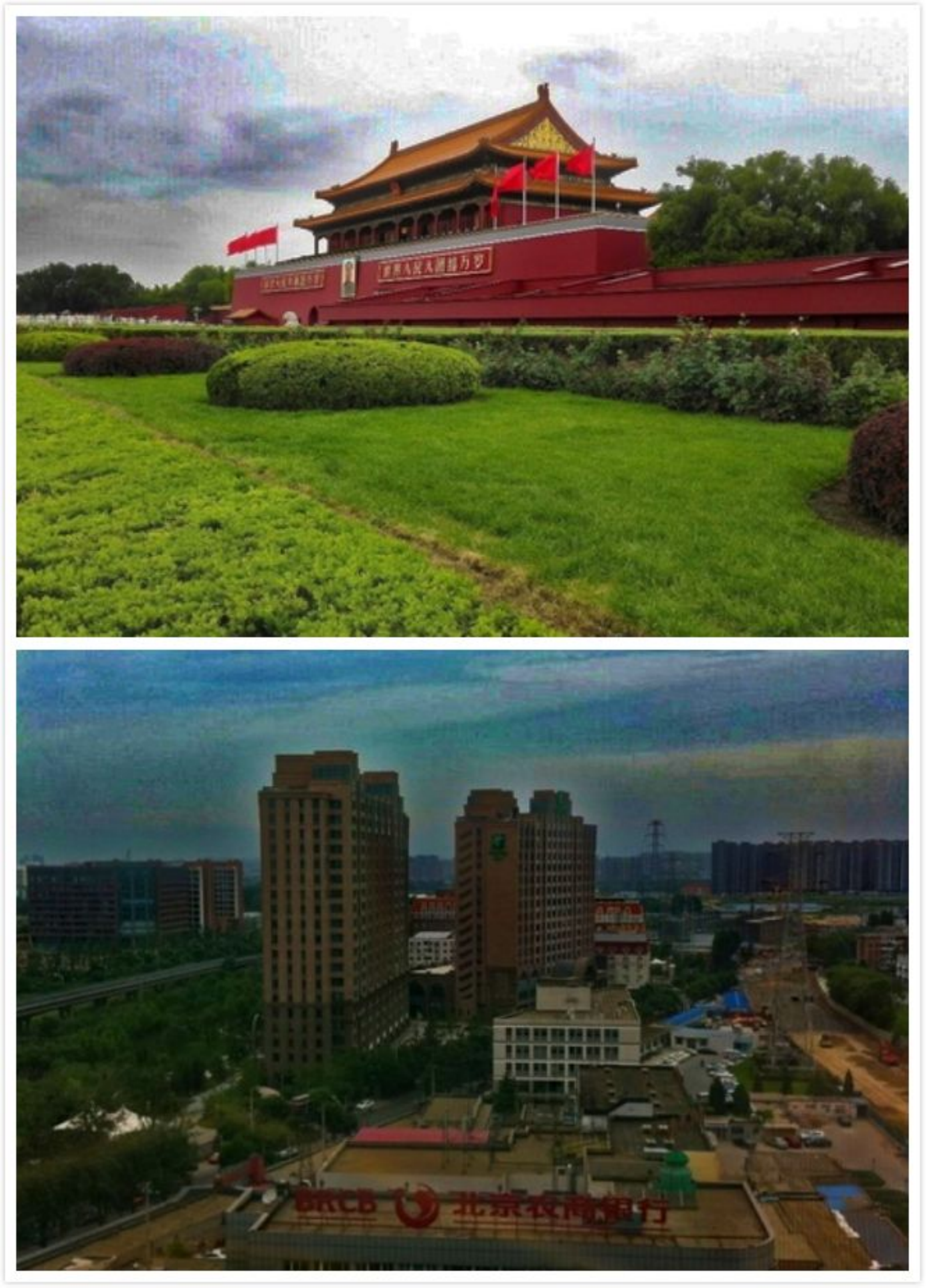}		}\end{minipage}
	\begin{minipage}{0.15\textwidth}
		\centering \subfigure[FVR] {
			\includegraphics[width=\textwidth,height=1.5in]{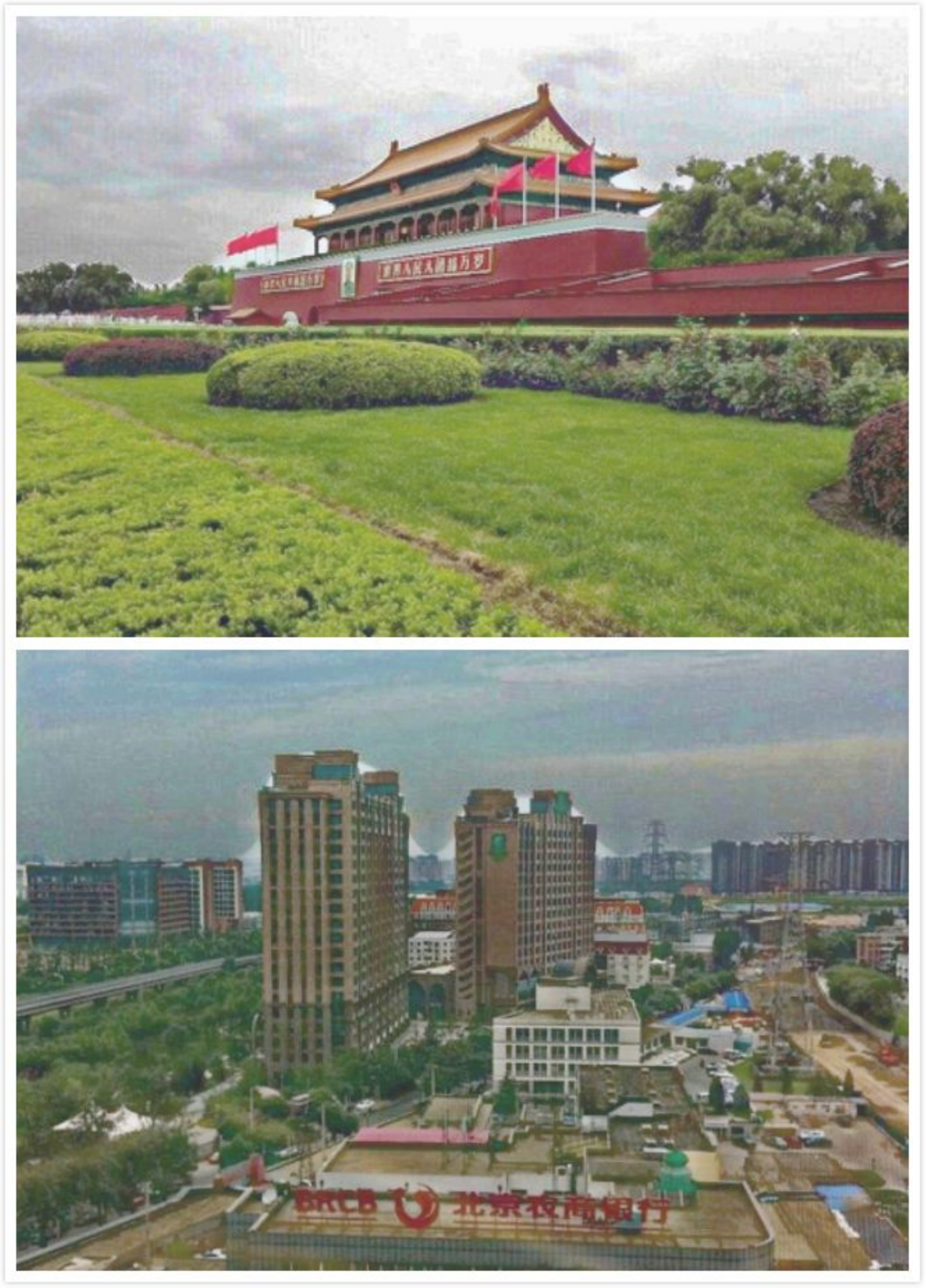}		}\\
		\subfigure[BCCR] {
			\includegraphics[width=\textwidth,height=1.5in]{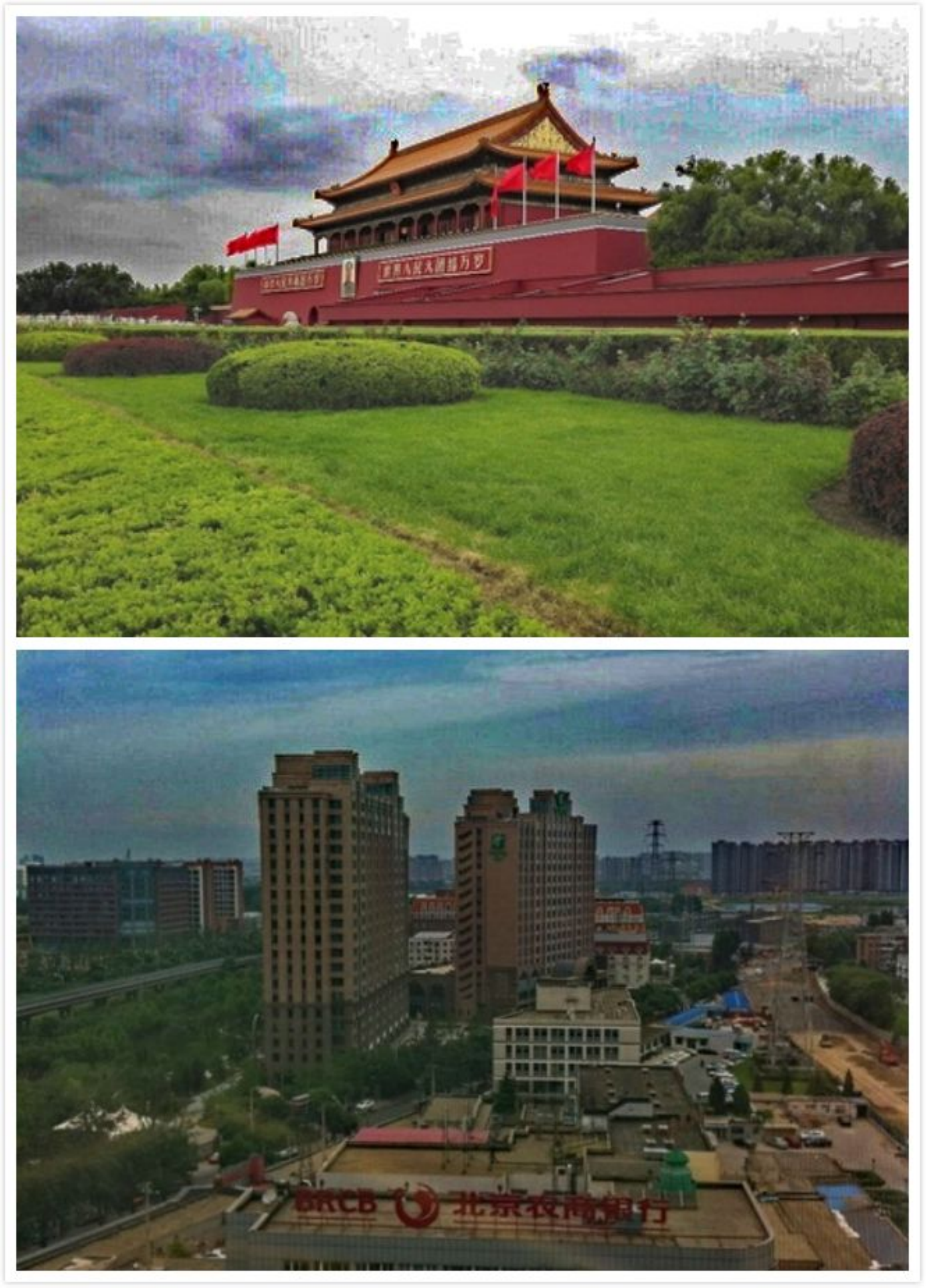}	}\end{minipage}
	\begin{minipage}{0.15\textwidth}
		\centering \subfigure[GRM] {
			\includegraphics[width=\textwidth,height=1.5in]{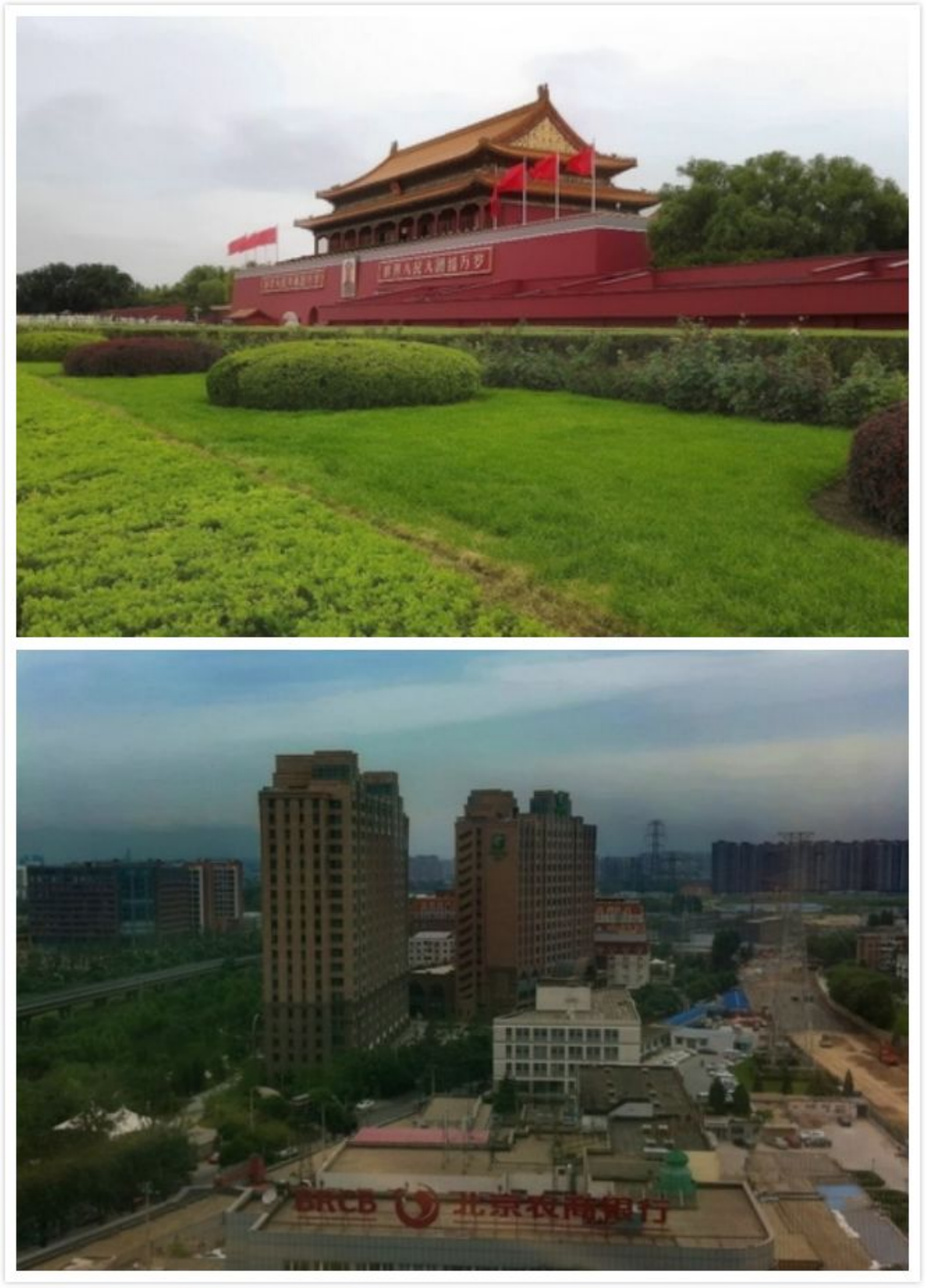}		}\\
		\subfigure[CAP] {
			\includegraphics[width=\textwidth,height=1.5in]{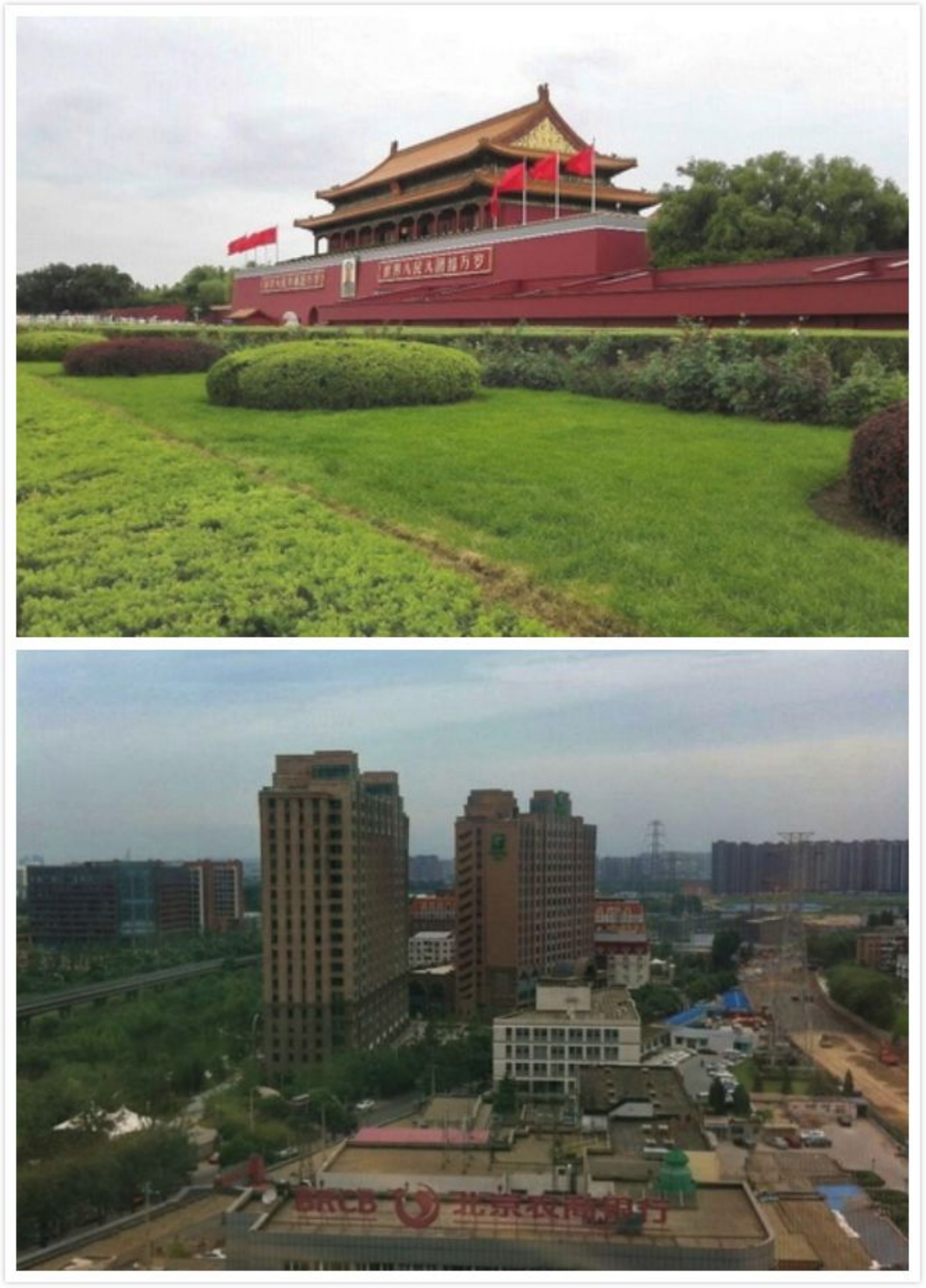}	}\end{minipage}
	\begin{minipage}{0.15\textwidth}
		\centering \subfigure[NLD] {
			\includegraphics[width=\textwidth,height=1.5in]{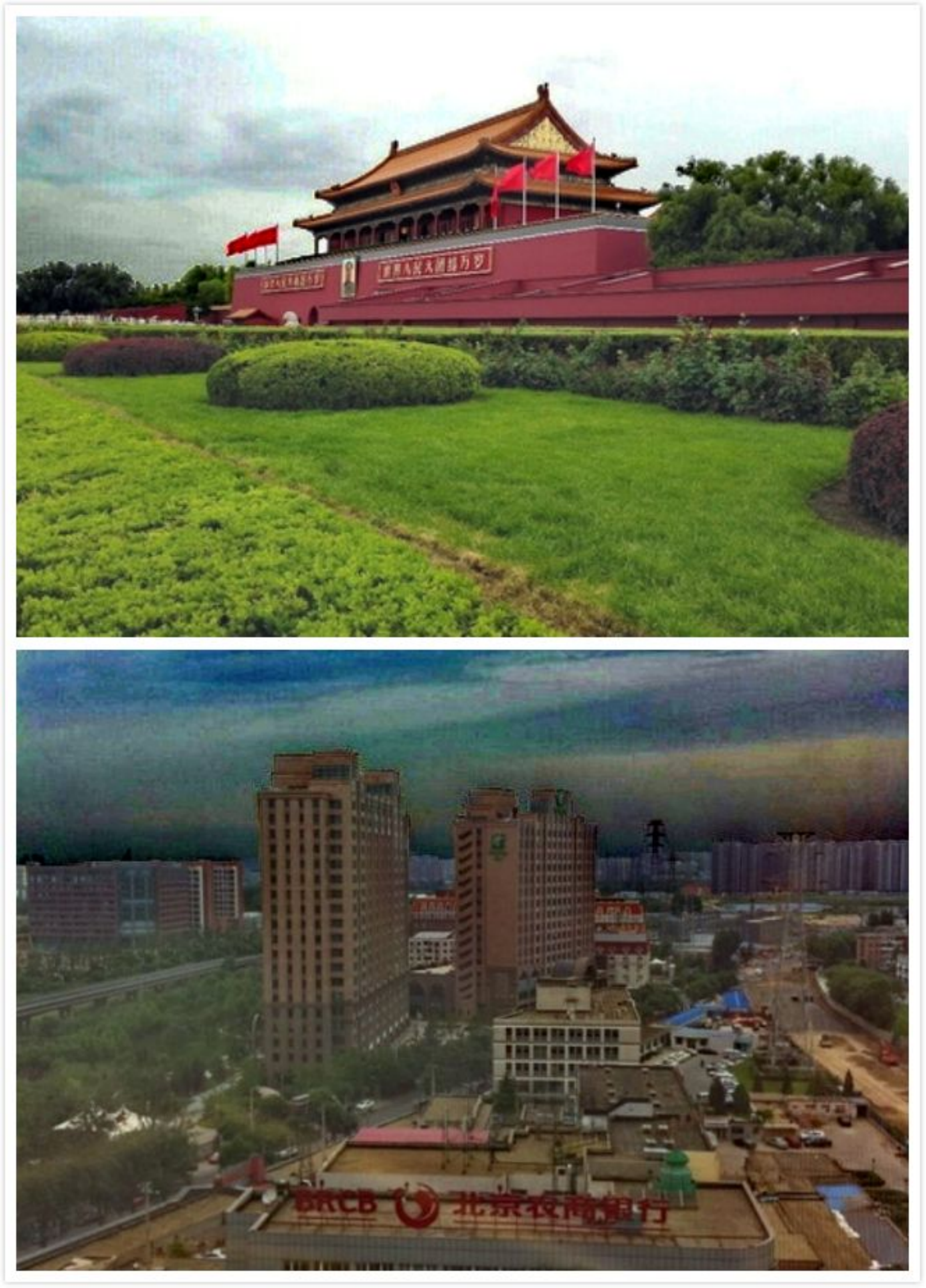}		}\\
		\subfigure[DehazeNet] {
			\includegraphics[width=\textwidth,height=1.5in]{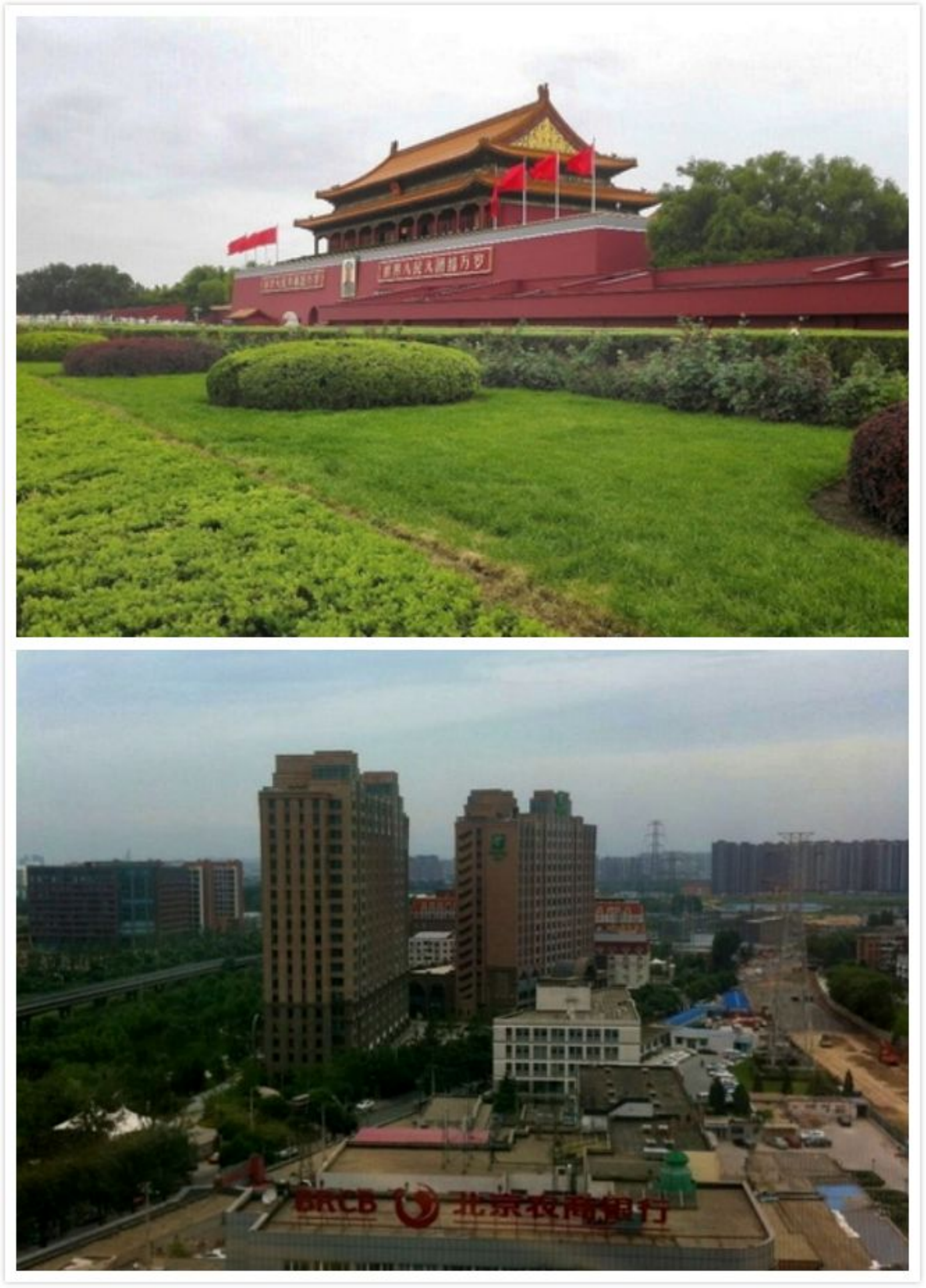}	}\end{minipage}
	\begin{minipage}{0.15\textwidth}
		\centering \subfigure[MSCNN] {
			\includegraphics[width=\textwidth,height=1.5in]{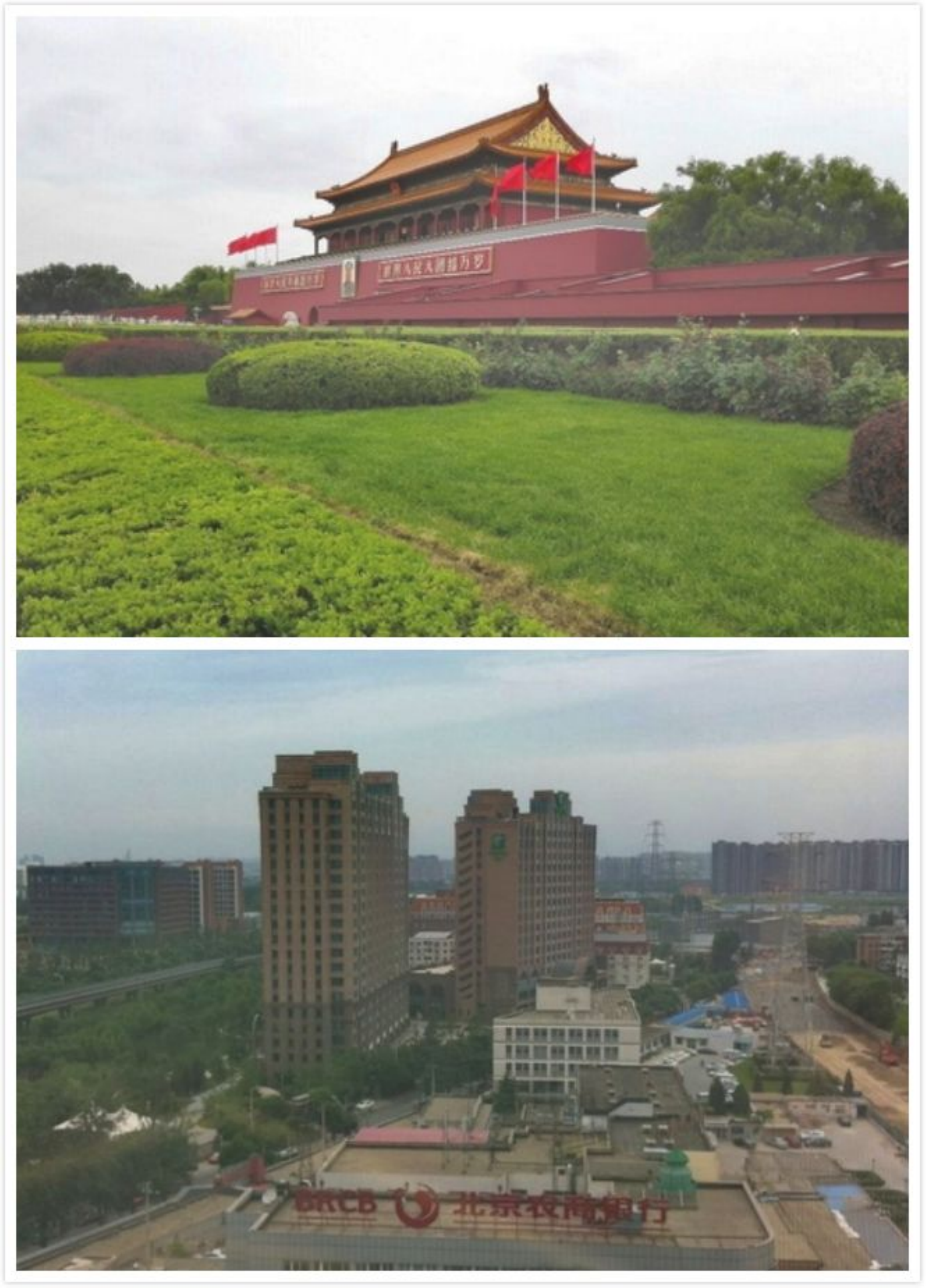}		}\\
		\subfigure[AOD-Net] {
			\includegraphics[width=\textwidth,height=1.5in]{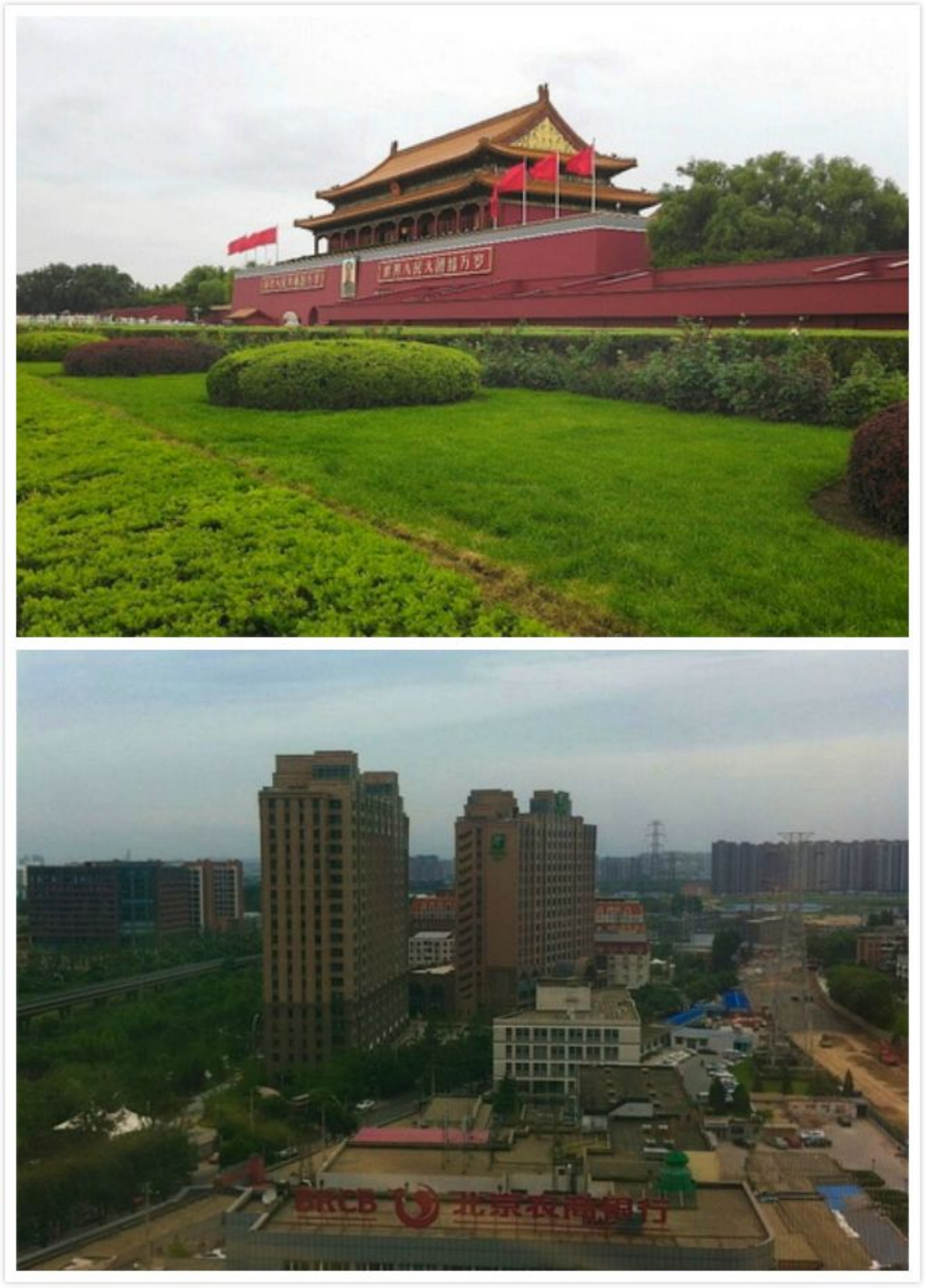}	}\end{minipage}
	\caption{Examples of dehazed results on a synthetic hazy image from HSTS.}
	\label{fig:sub_img1}
\end{figure*}

\begin{figure*}[t]
	\centering
	\centering
	\begin{minipage}{0.18\textwidth}
		\centering \subfigure[Hazy Image] {
			\includegraphics[width=\textwidth]{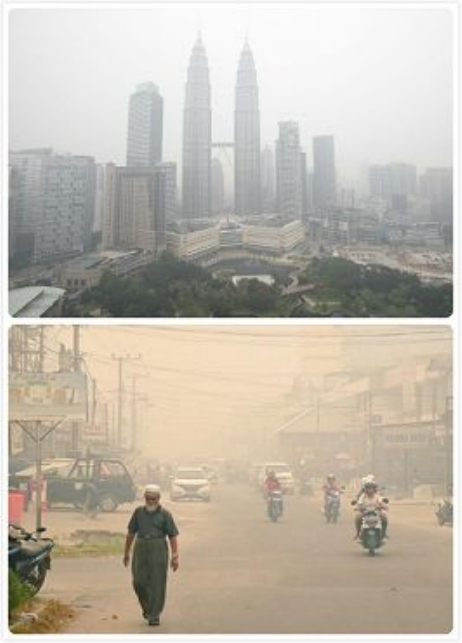}		}\\
		\subfigure[DCP] {
			\includegraphics[width=\textwidth]{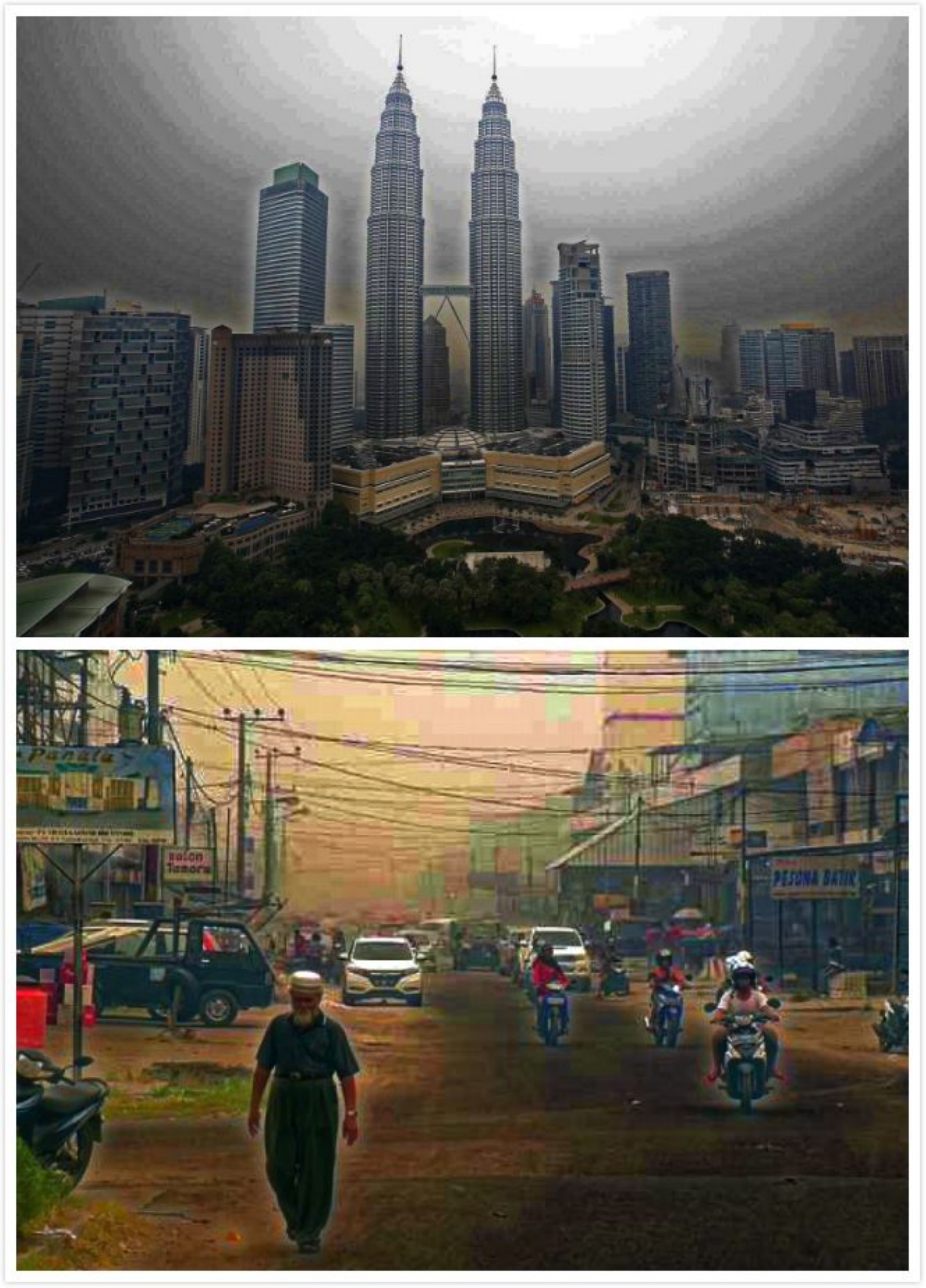}		}\end{minipage}
	\begin{minipage}{0.18\textwidth}
		\centering \subfigure[FVR] {
			\includegraphics[width=\textwidth]{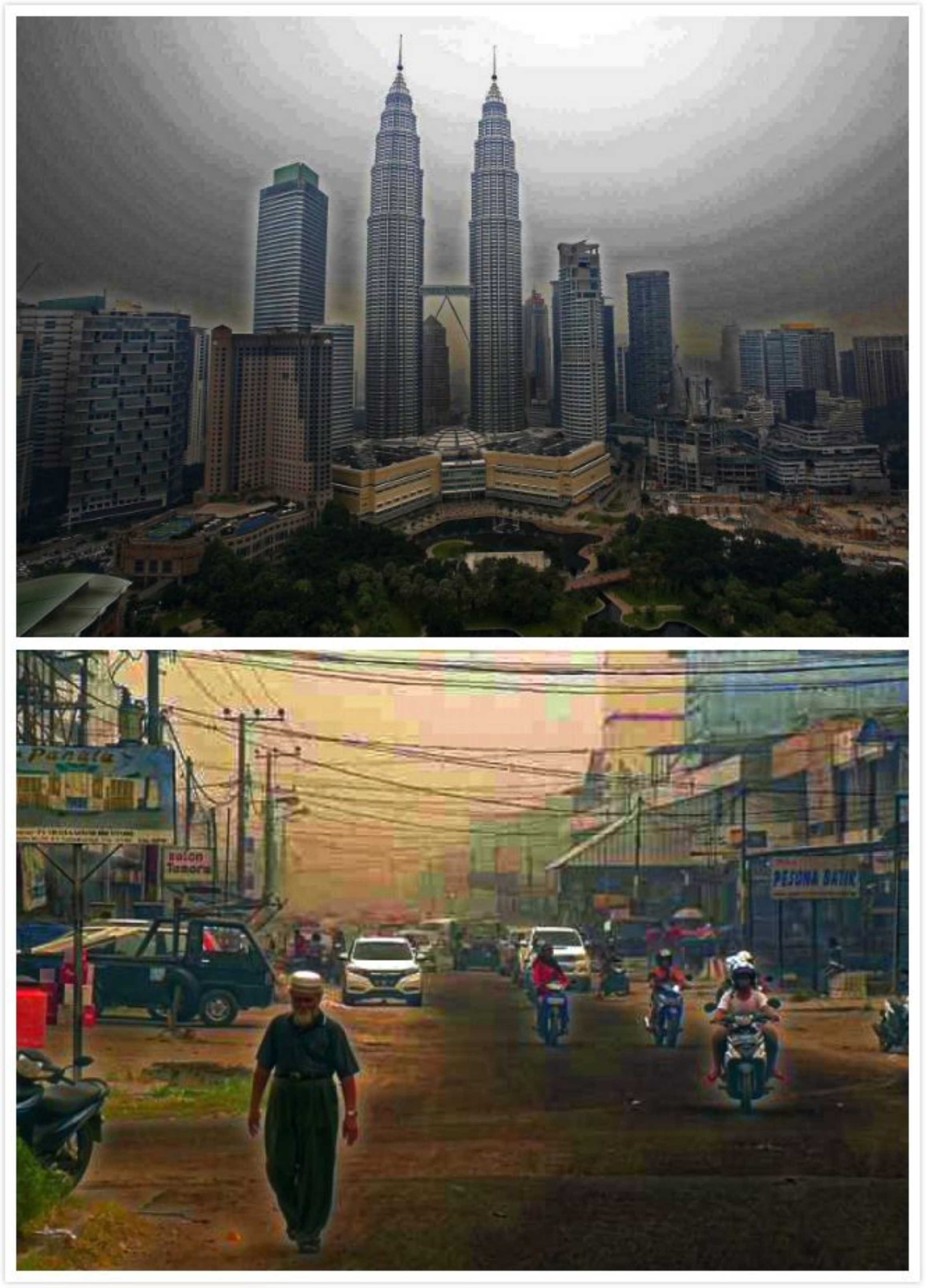}		}\\
		\subfigure[BCCR] {
			\includegraphics[width=\textwidth]{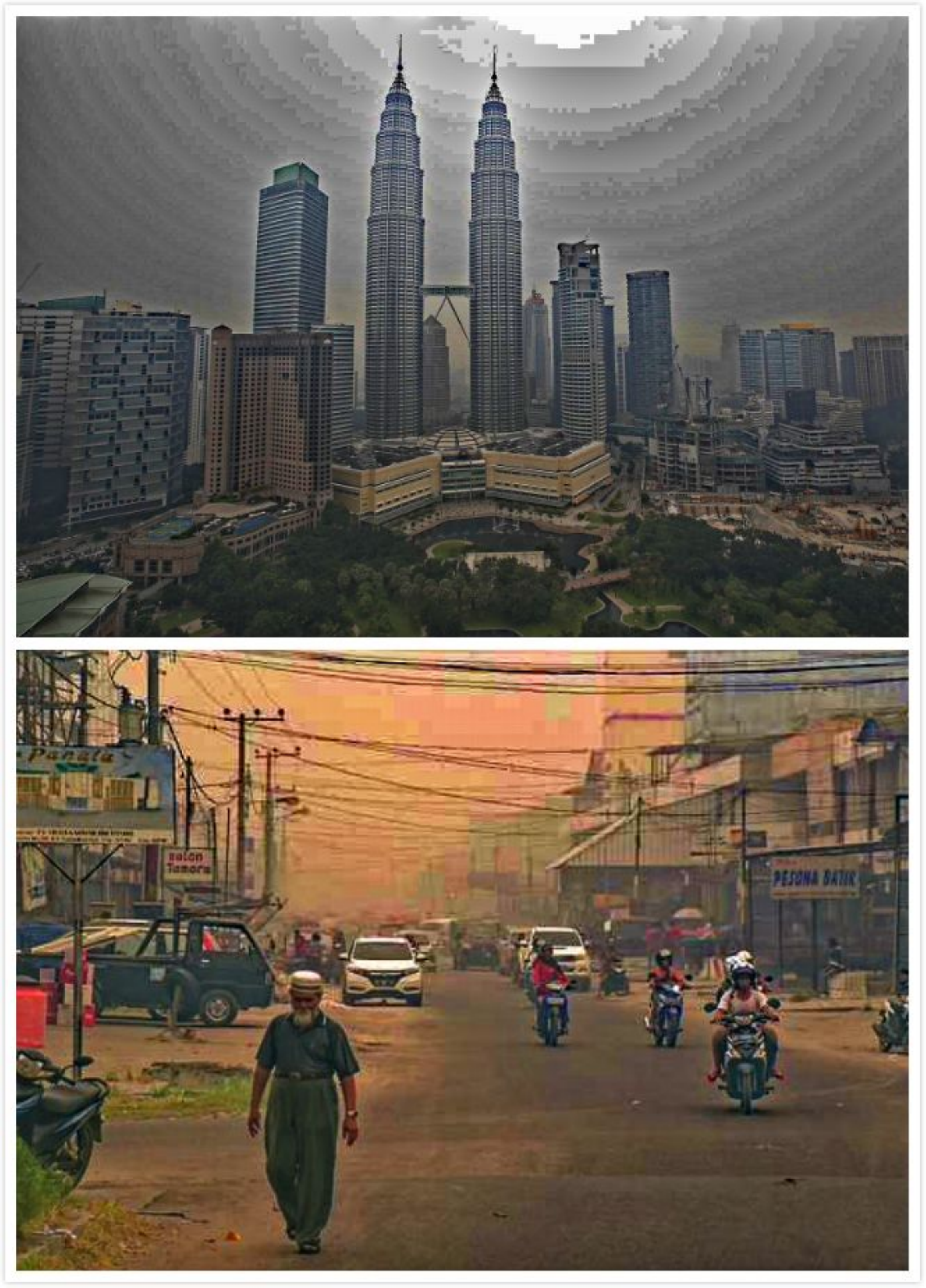}	}\end{minipage}
	\begin{minipage}{0.18\textwidth}
		\centering \subfigure[GRM] {
			\includegraphics[width=\textwidth]{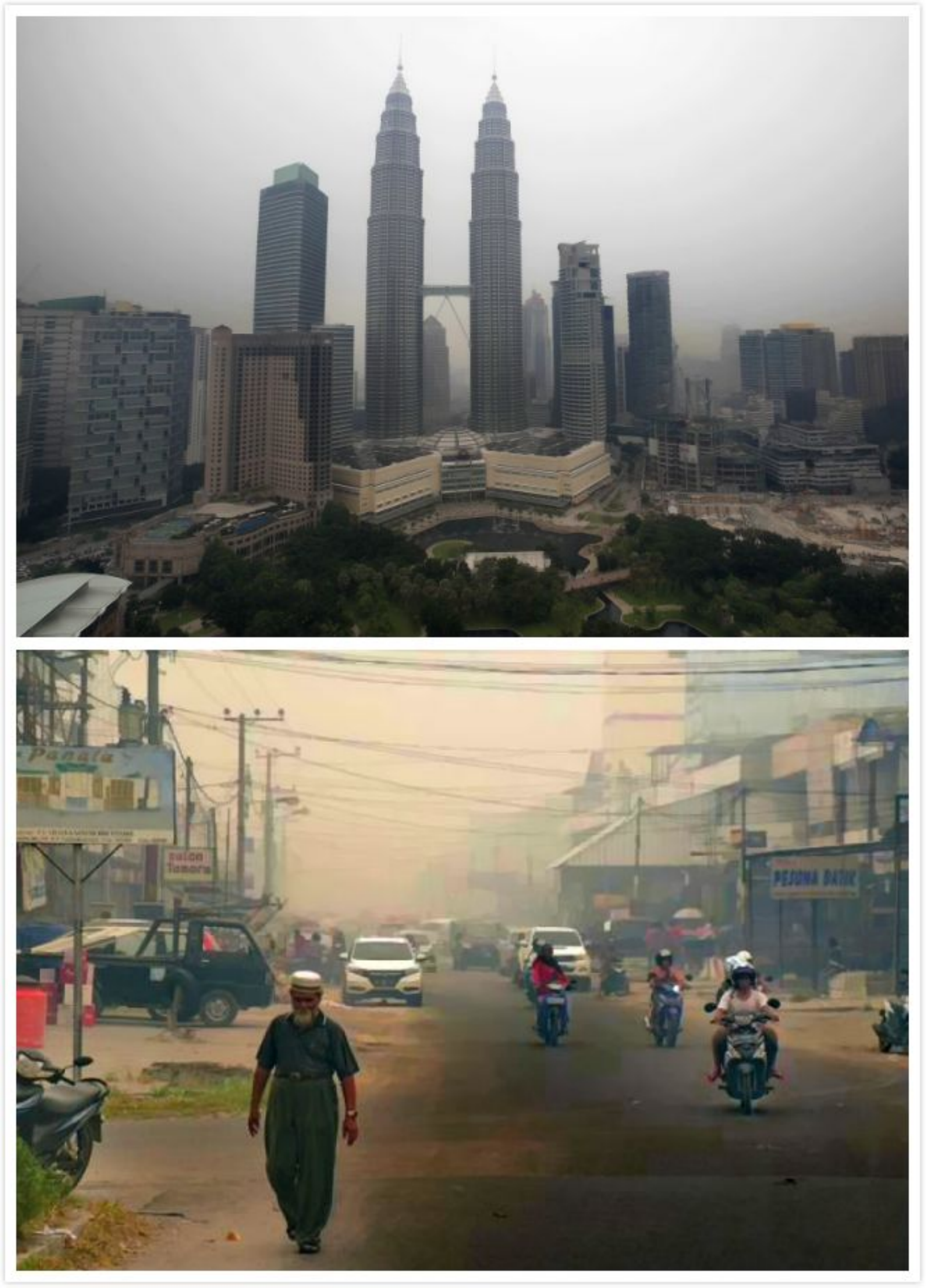}		}\\
		\subfigure[CAP] {
			\includegraphics[width=\textwidth]{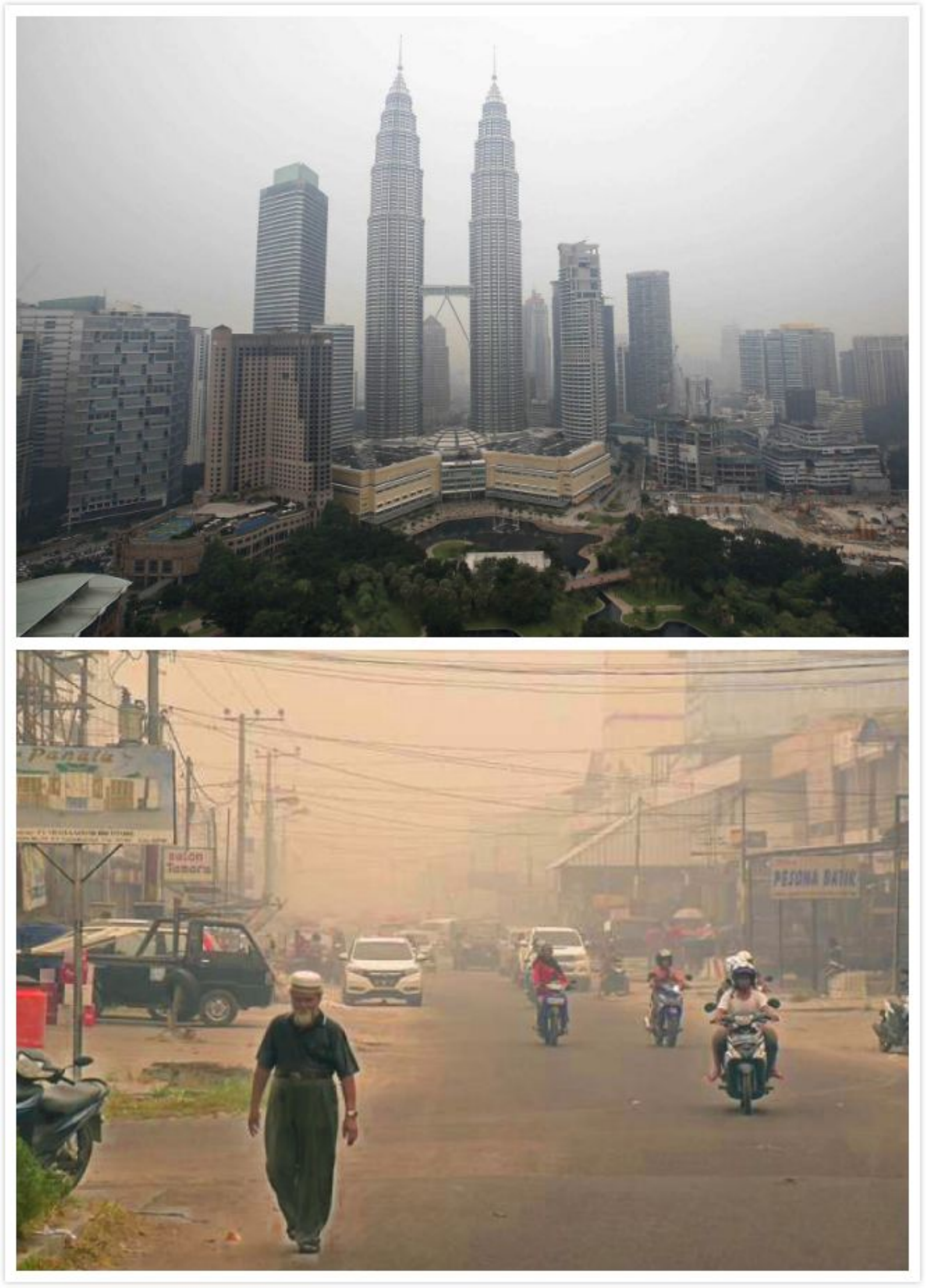}	}\end{minipage}
	\begin{minipage}{0.18\textwidth}
		\centering \subfigure[NLD] {
			\includegraphics[width=\textwidth]{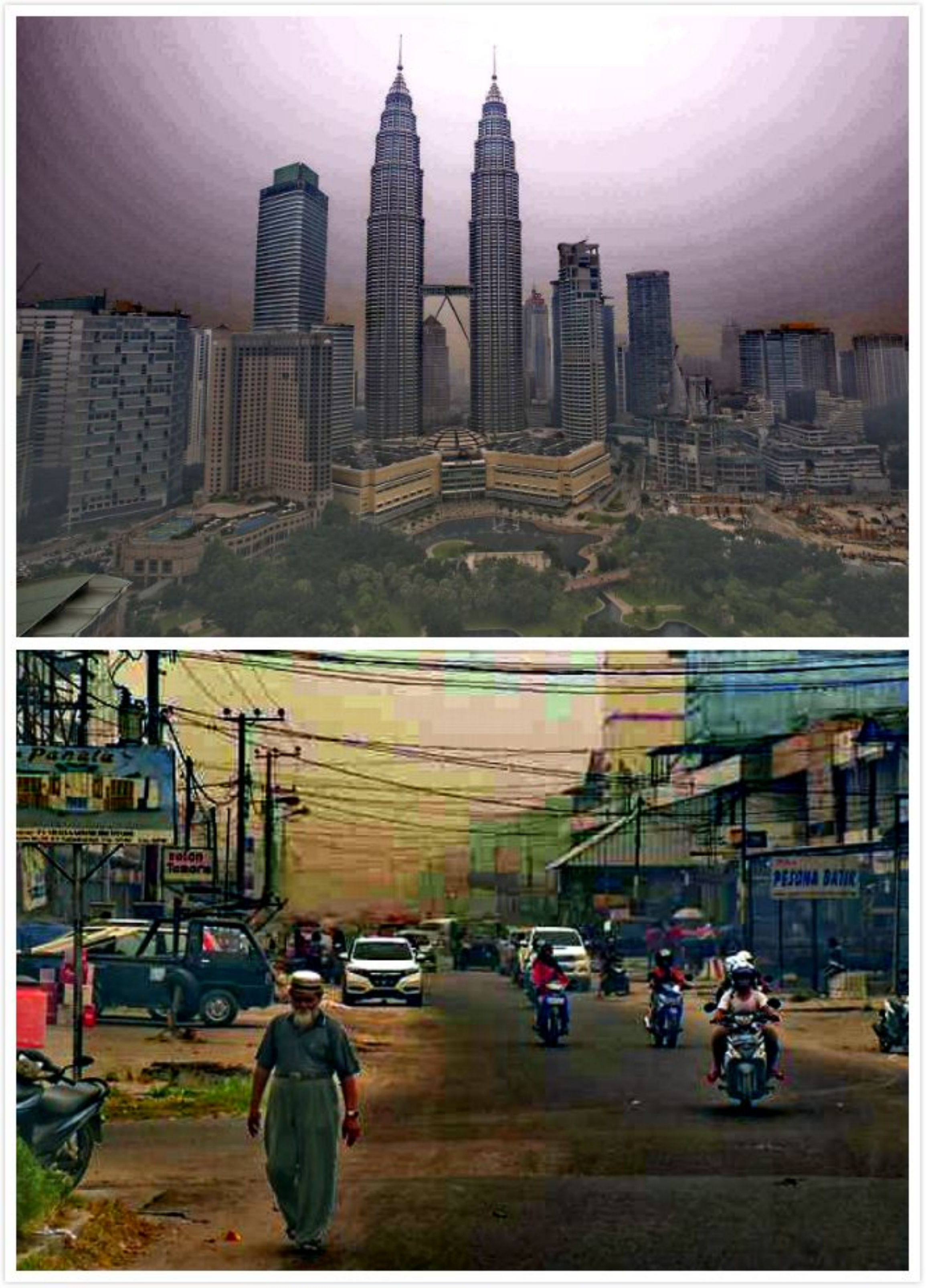}		}\\
		\subfigure[DehazeNet] {
			\includegraphics[width=\textwidth]{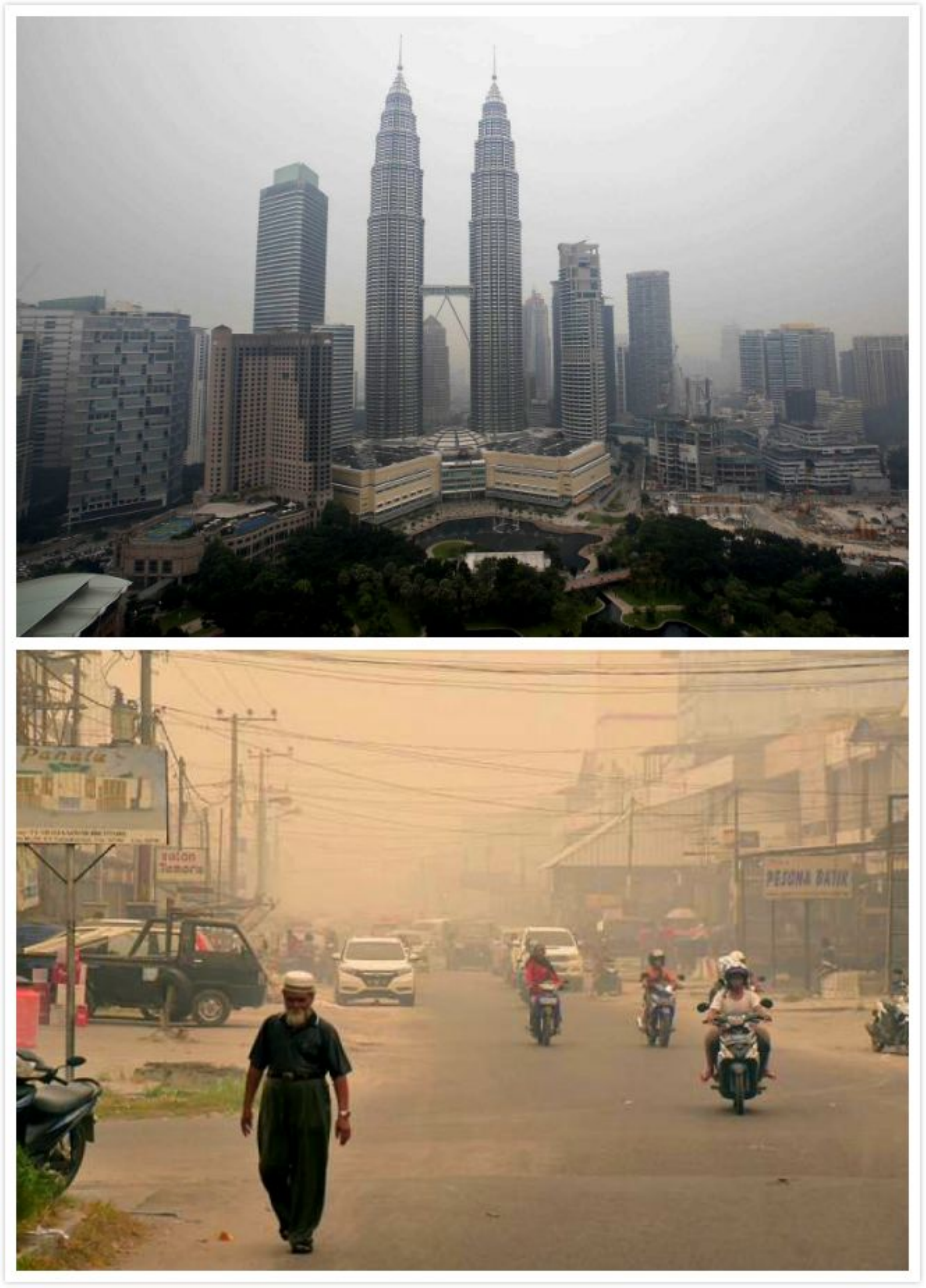}	}\end{minipage}
	\begin{minipage}{0.18\textwidth}
		\centering \subfigure[MSCNN] {
			\includegraphics[width=\textwidth]{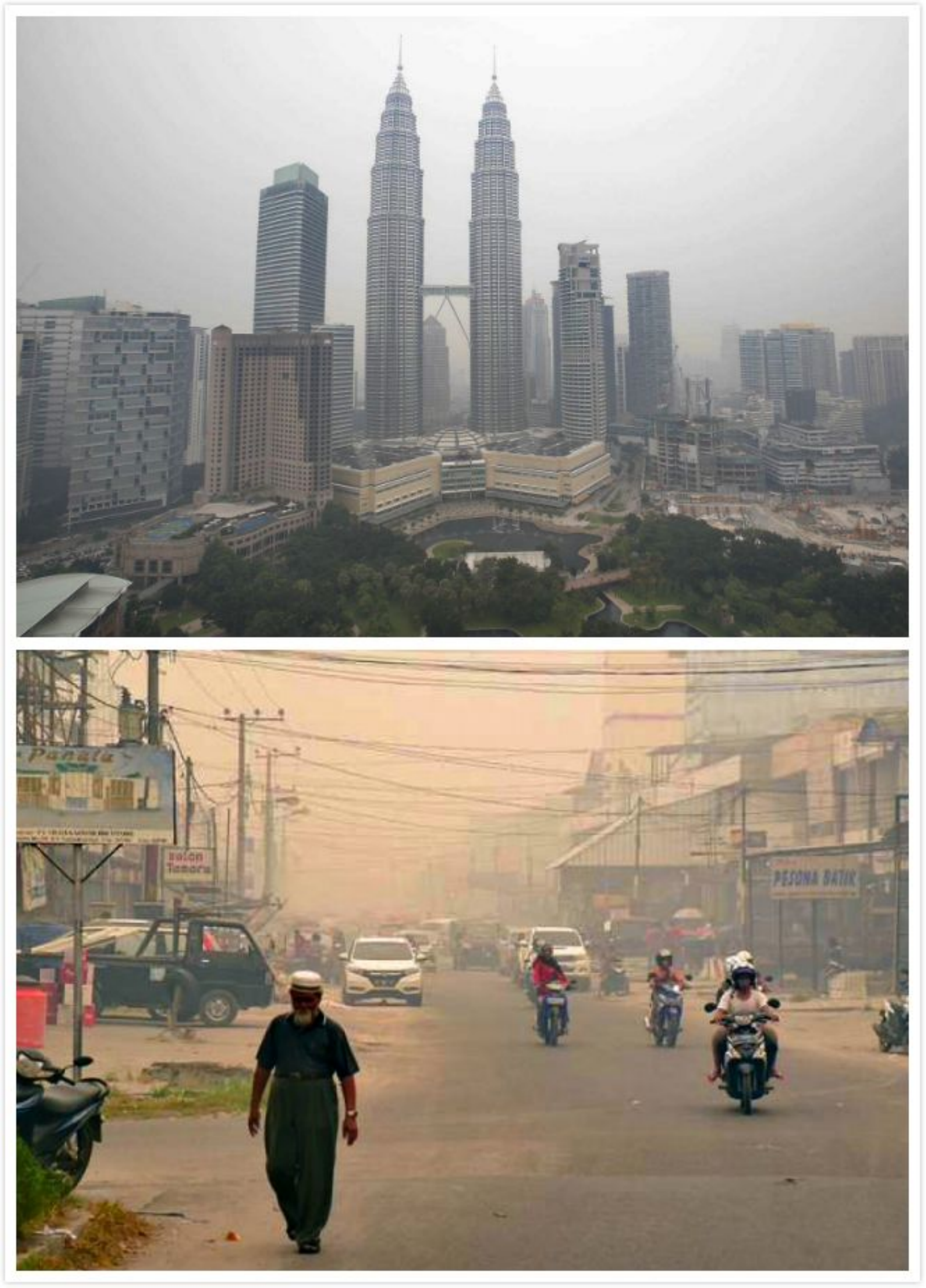}		}\\
		\subfigure[AOD-Net] {
			\includegraphics[width=\textwidth]{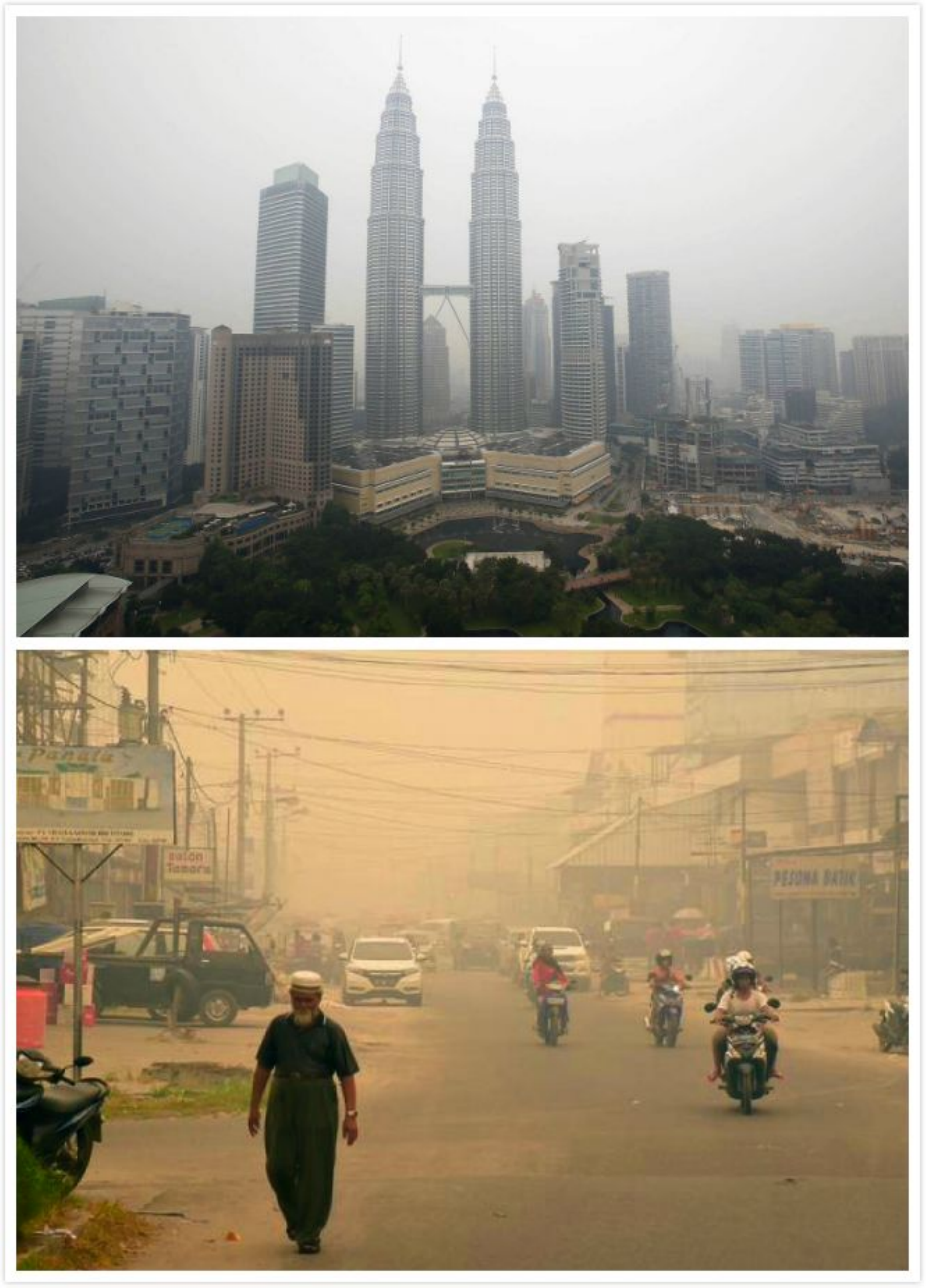}	}\end{minipage}
	\caption{Examples of dehazed results on a real-world hazy image from HSTS.}
	\label{fig:sub_img2}
\end{figure*}

Figures~\ref{fig:sub_img1} and~\ref{fig:sub_img2} show qualitative examples of dehazed results on a synthetic and a real-world image, respectively.
Quantitative results can be found in Table \ref{tab-clearness_authenticity} and the trends are visualized in Figure \ref{fig:realsyn_clearness_authenticity}. We also compute the full- and no-reference metrics on synthetic images to examine their consistency with the subjective scores.

A few interesting observations could be drawn:
\begin{itemize}
	\item The subjective qualities of various algorithms' results show different trends on synthetic and real hazy images. On the 10 synthetic images of HSTS, DCP~\cite{he2011singlecvpr} receives the best clearness score and DehazeNet is the best in authenticity score. On the 10 real images, CNN-based methods~\cite{cai2016dehazenet,ren2016single,li2017aod} rank top-3 in terms of both clearness and authenticity, in which MSCNN~\cite{ren2016single} achieves the best according to both scores. 
	\item The clearness and authenticity scores of the same image are often not aligned. As can be seen from Figure \ref{fig:realsyn_clearness_authenticity}, the two subjective scores are hardly correlated on synthetic images; their correlation shows better on real images. That reflects the complexity and multi-facet nature of subjective perceptual evaluation.
	\item From Table \ref{tab-clearness_authenticity}, we observe the divergence between subjective and objective (both full- and no-reference) evaluation results. For the best performer in subjective evaluation, MSCNN~\cite{ren2016single}, its PSNR/SSIM results on synthetic indoor images are quite low, while SSEQ/BLIINDS-II on both synthetic and outdoor images are moderate. As another example, GRM~\cite{chen2016robust} receives the highest SSEQ/BLIINDS-II scores on real HSTS images. However, both of its subjective scores rank only fifth among nine algorithms on the same set. 
\end{itemize}

\begin{table*}[!ht]
	\caption{Comparison of average per-image running time (second) on synthetic indoor images in SOTS.}
	\begin{center}\footnotesize{
			\begin{tabular}{c|c|c|c|c|c|c|c|c|c}
				\hline
				& DCP~\cite{he2011singlecvpr}  & FVR~\cite{tarel2009fast} & BCCR~\cite{meng2013efficient} & GRM~\cite{chen2016robust} & CAP~\cite{zhu2015fast} & NLD~\cite{berman2016non} & DehazeNet~\cite{cai2016dehazenet} & MSCNN~\cite{ren2016single} & AOD-Net~\cite{li2017aod}\\
				\hline
				Time &\textcolor{blue}{1.62}&6.79&3.85 & 83.96  & \textcolor{cyan}{0.95}  & 9.89 & 2.51 &  2.60  &   \textcolor{red}{0.65} \\
				\hline
		\end{tabular}}
		\label{tab-time}
	\end{center}
\end{table*}

\subsection{Running Time}
Table~\ref{tab-time} reports the per-image running time of each algorithm, averaged over the synthetic indoor images ($620\times 460$) in SOTS, using a machine with 3.6 GHz CPU and 16G RAM. All methods are implemented in MATLAB, except AOD-Net by Pycaffe. However, it is fair to compare AOD-Net with other methods since MATLAB implementation has superior
efficiency than Pycaffe as shown in~\cite{li2017aod}. AOD-Net shows a clear advantage over others in efficiency, thanks to its light-weight feed-forward structure.

\section{What are Beyond: From RESIDE to RESIDE-$\beta$}

RESIDE serves as a sufficient benchmark for evaluating single image dehazing as a traditional image restoration problem: either to ensure signal fidelity or to please \textit{human vision}. However, dehazing is increasingly demanded in \textit{machine vision} systems in outdoor environments, whose requirement is not naturally met by taking an image restoration viewpoint. To identify and eliminate the gaps between \textit{current dehazing research} and the \textit{practical application need}, we introducing the RESIDE-$\beta$ part, as an exploratory and supplementary part of the RESIDE benchmark, including two innovative explorations on solving two hurdles, on training data content and evaluation criteria, respectively. Being our novel try, RESIDE-$\beta$ has a ``beta stage'' nature and is meant to inspire more followers.

\begin{figure}[t]\footnotesize
	\begin{center}
		\begin{tabular}{@{}cccc@{}}
			\includegraphics[width = 0.11\textwidth]{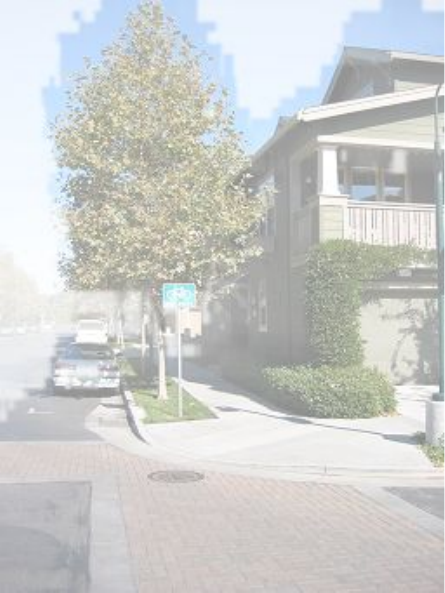} & \hspace{-0.4cm}
			\includegraphics[width = 0.11\textwidth]{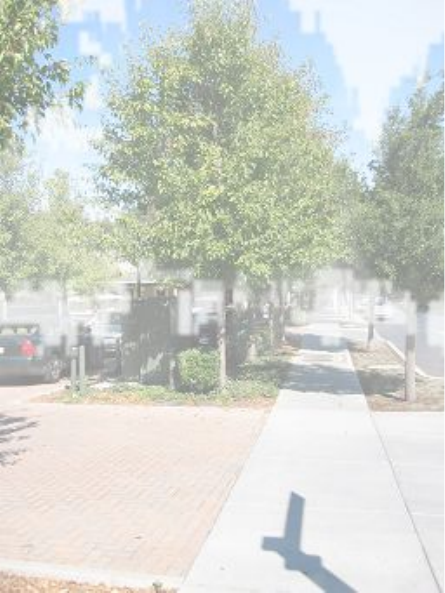} & \hspace{-0.4cm}
			\includegraphics[width = 0.11\textwidth]{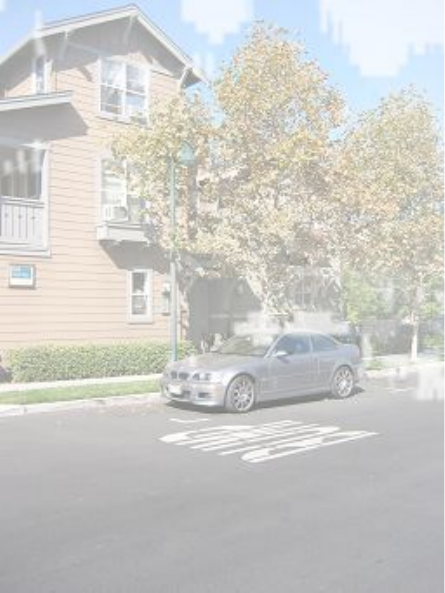} & \hspace{-0.4cm}
			\includegraphics[width = 0.11\textwidth]{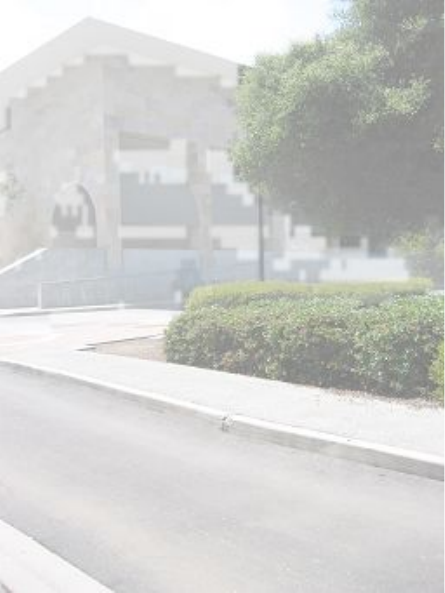} \\
			\includegraphics[width = 0.11\textwidth]{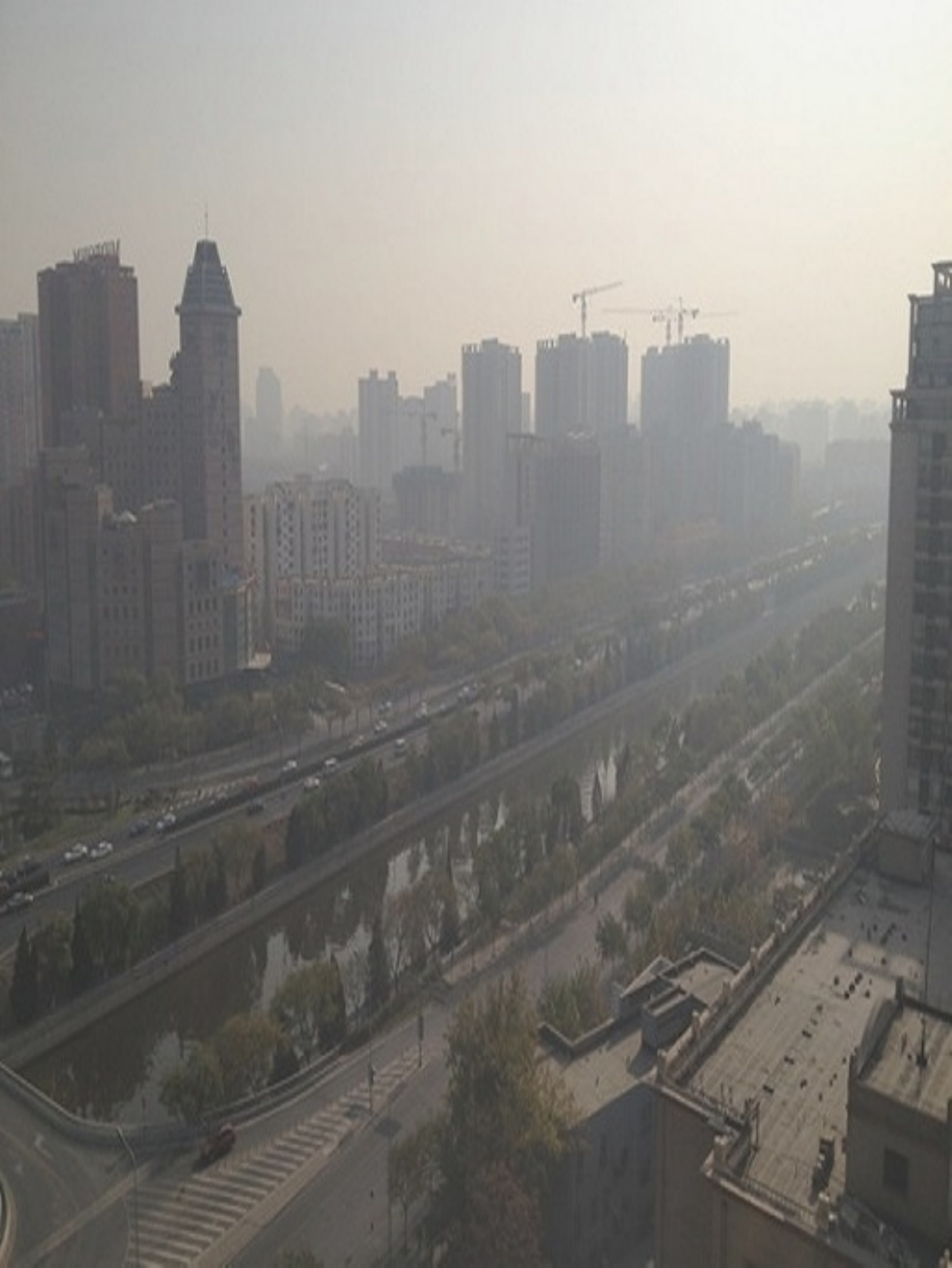} & \hspace{-0.4cm}
			\includegraphics[width = 0.11\textwidth]{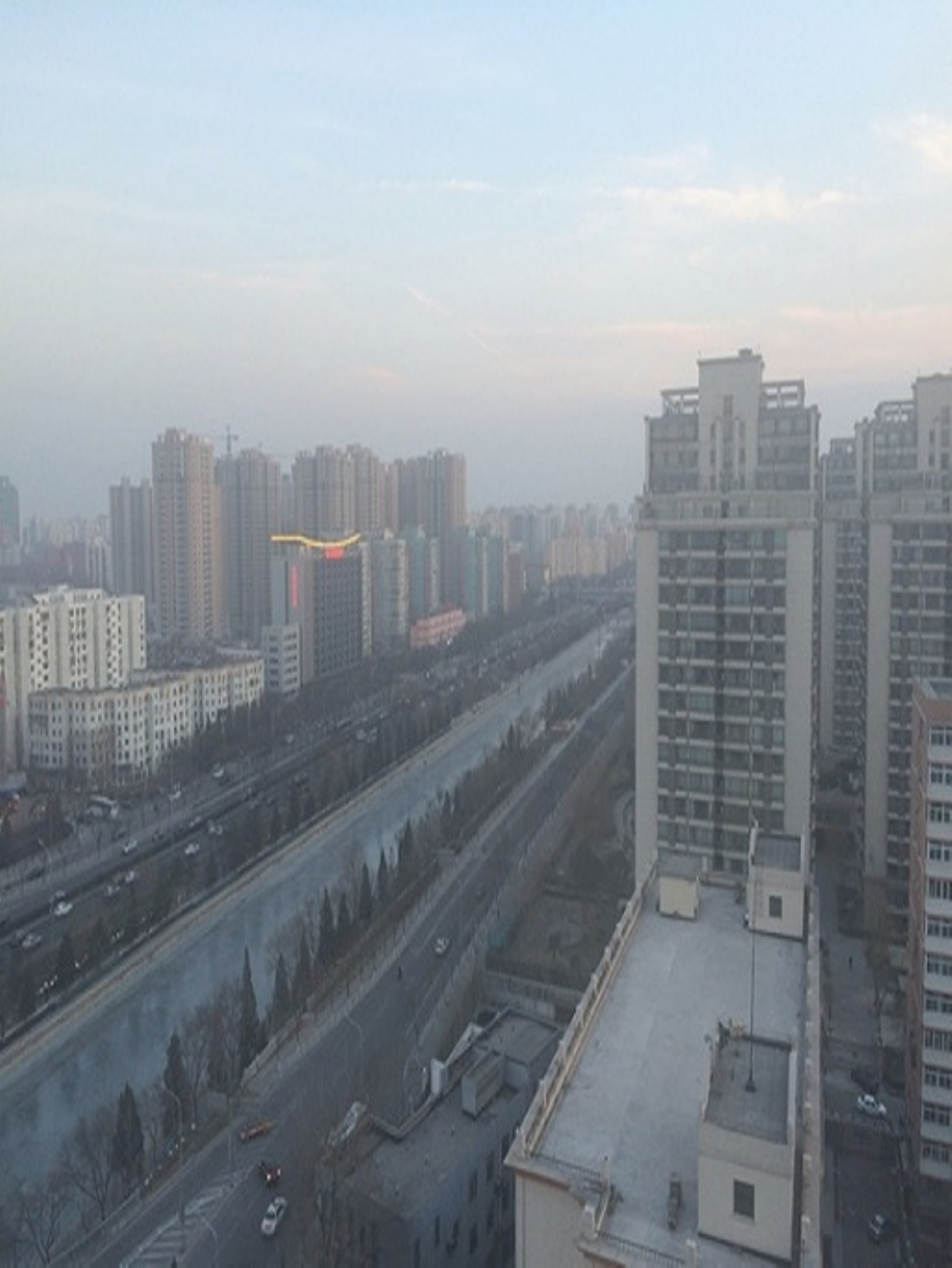} & \hspace{-0.4cm}
			\includegraphics[width = 0.11\textwidth]{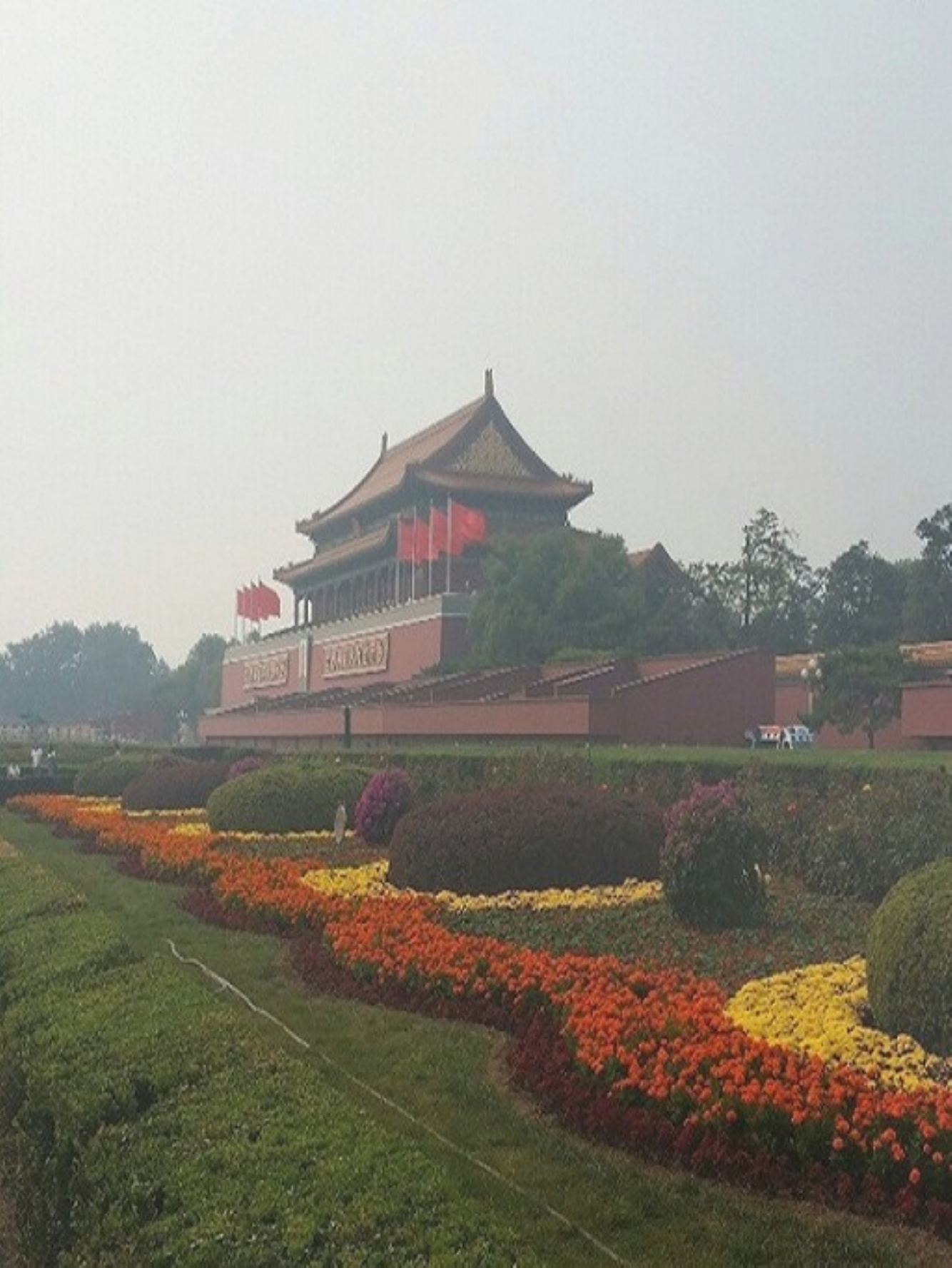} & \hspace{-0.4cm}
			\includegraphics[width = 0.11\textwidth]{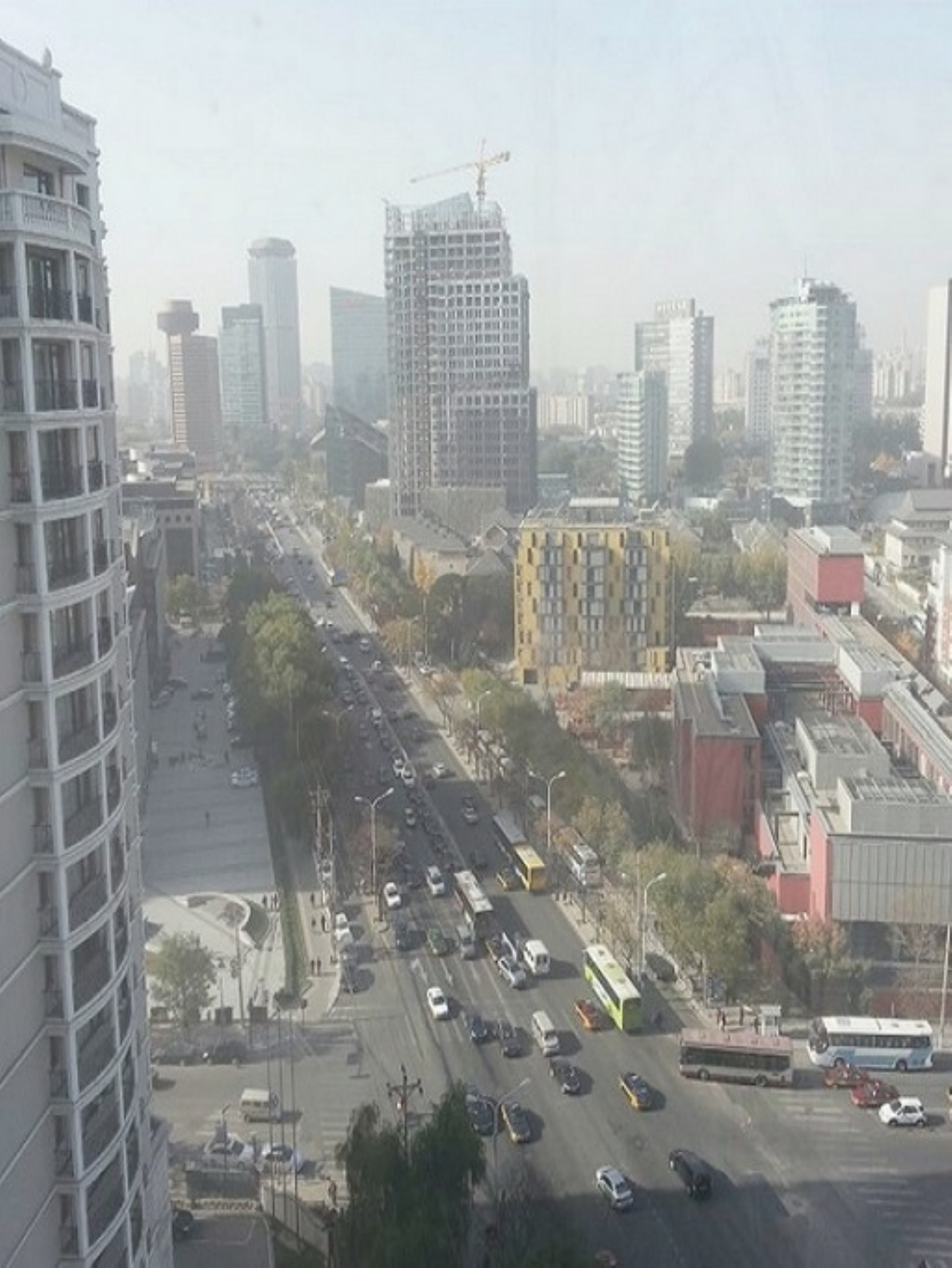}
		\end{tabular}
	\end{center}
	\caption{Visual comparison between the synthetic hazy images directly generated from Make3D (first row) and from OTS (second row).
	}
	\label{fig-3dout}
\end{figure}

\subsection{Indoor versus Outdoor Training Data}

Up to our best knowledge, almost all data-driven dehazing models have been utilizing synthetic training data, because of the prohibitive difficulty of simultaneously collecting real-world hazy RGB images and their ``hazy-free'' ground truth. Most outdoor scenes contain object movements from time to time, e.g. traffic surveillance and autonomous driving. Even in a static outdoor scene, the change of illumination conditions etc. along time is inevitable. Despite their positive driving effects in the development of dehazing algorithms, those synthetic images are collected from indoor scenes \cite{silberman2012indoor,scharstein2003high}, while dehazing is applied to outdoor environments. 

The content of training data thus significantly diverges from the target subjects in real dehazing applications. Such a mismatch might undermine the practical effectiveness of the trained dehazing models. 
\cite{zhang2017hazerd} collected 14 outdoor clean images with accurate depth information, and proposed to generate hazy images from them with parameters that are chosen to be physically realistic. Their meaningful and delicate efforts are however not straightforward to scale up and generate large-scale training sets. 

 Aiming for automatic generation of large-scale realistic outdoor hazy images, we first examine the possibility of utilizing existing outdoor depth datasets. While several such datasets, e.g., Make3D \cite{saxena2009make3d} and KITTI \cite{geiger2012we}, have been proposed, their depth information is less precise and incomplete compared to indoor datasets. For example, due to the limitations of RGB-based depth cameras, the Make3D dataset suffer from at least 4 meters of average root mean squared error in the predicted depths, and the KITTI dataset has at least 7 meters of average error \cite{ma2017sparse}. In comparison, the average depth errors in indoor datasets, e.g., NYU-Depth-v2 ~\cite{silberman2012indoor}, are usually as small as 0.5 meter. For the outdoor depth maps can also contain a large amount of artifacts and large holes, which renders it inappropriate for direct use in haze simulation. We choose Make3D to synthesize hazy images in the same way as we did for RESIDE training set, a number of examples being displayed at the first row of Figure \ref{fig-3dout}. It can be easily seen that they suffer from unrealistic artifacts (e.g., notice the ``blue'' regions around the tree), caused by inaccurate depth map. A possible remedy is to adopt recent approaches of depth map denoising and in-painting \cite{wang2008stereoscopic,sakaridis2017semantic}, which we leave for future. 

Another option is to estimate depth from outdoor images and then synthesizing hazy images. After comparing different depth estimation methods, we find the algorithm in \cite{liu2016learning} to produce fewest visible depth errors and to cause much less visual artifacts on natural outdoor images, same as \cite{li2017aod} observed. We display a few synthetic hazy examples generated by using \cite{liu2016learning} for depth estimation, in the second row of Figure \ref{fig-3dout}. By comparing them with the first row (Make3D), one can see that using depth estimation \cite{liu2016learning} leads to much more visually plausible results. 

We thus extend to a large scale effort, collecting 2, 061 real world outdoor images from \cite{BeijingImages}, among which we carefully excluded those originally with haze and ensure their scenes to be as diverse as possible. We use \cite{liu2016learning} to estimate the depth map for each image, with which we finally synthesize 72, 135 outdoor hazy images with $\beta$ in [0.04, 0.06, 0.08, 0.1, 0.12, 0.16, 0.2] and A in [0.8, 0.85, 0.9, 0.95, 1]. This new set, called Outdoor Training Set (OTS), consists of paired clean outdoor images and generated hazy ones. It is included as a part of RESIDE-$\beta$, and could be used for training. Despited that depth estimation could potentially be noisy, we visually inspect the new set and find most generated hazy images to be free of noticeable artifacts (and much better than generating using Make3D). As we observed from preliminary experiments, including this outdoor set for training  performed in general similarly on SOTS in the sense of PSNR/SSIM, but improved the generalization performance on real-world images, in terms of visual quality.

\subsection{Restoration versus High-Level Vision}
It has been recognized that the performance of high-level computer vision tasks, such as object detection and recognition, will deteriorate in the presence of various degradations, and is thus largely affected by the quality of image restoration and enhancement. Dehazing could be used as pre-processing for many computer vision tasks executed in the wild, and the resulting task performance could in turn be treated as an indirect indicator of the dehazing quality. Such a ``task-driven'' evaluation way has received little attention so far, despite its great implications for outdoor applications. 

A relevant preliminary effort was presented in \cite{li2017aod}, where the authors compared a few CNN-based dehazing models by placing them in an object detection pipeline, but their tests were on synthetic hazy data with bounding boxes. \cite{sakaridis2017semantic} created a relatively small dataset of 101 real-world images depicting foggy driving scenes, which came with ground truth annotations for evaluating semantic segmentation and object detection. 
We notice that \cite{sakaridis2017semantic} investigated detection and segmentation problems in hazy images as well, evaluated on a small image set with only three dehazing methods. 

\subsubsection{Full-Reference Perceptual Loss Comparison on SOTS}

Since dehazed images are often subsequently fed for automatic semantic analysis tasks such as recognition and detection, we argue that the optimization target of dehazing in these tasks is neither pixel-level or perceptual-level quality, but the utility of the dehazed images in the given semantic analysis task \cite{liu2017recognizable}. The perceptual loss \cite{johnson2016perceptual} was proposed to measure the semantic-level similarity of images, using the VGG recognition model\footnote{Public available at \url{http://www.robots.ox.ac.uk/~vgg/software/very_deep/caffe/VGG_ILSVRC_16_layers.caffemodel}} pre-trained on ImageNet dataset\cite{russakovsky2015imagenet}. Here, we compared the Euclidean distance between clean images and dehazed images with different level features including relu2\_2, relu3\_3, relu4\_3 and relu5\_3. Since it is a full-reference metric, we compute the perceptual loss on the SOTS dataset, as listed in Table \ref{tab-p_loss}. We also compute the perceptual loss on the 10 synthetic images in HSTS, to examine how well it agrees with the perceptual quality, as seen from Table \ref{tab-p_loss_hsts}. DehazeNet and CAP consistently lead to the lowest perceptual loss differences on both sets, which seem to be in general aligned with PSNR results, but not SSIM or other two no-reference metrics. 

On HSTS synthetic images, we observe the perceptual loss to be correlated to the authenticity score to some extent (e.g., DehazeNet and AOD-Net perform well under both), but hardly correlated to the clearness. It might imply that for preserving significant semantical similarities for recognition, it is preferable to keep a realistic visual look than to thoroughly remove haze. In other words, ``under-dehazed'' images might be preferred over ``over-dehazed'' images, the latter potentially losing details and suffering from method artifacts.

\begin{table*}[!ht]
	\caption{perceptual loss on SOTS　indoor images.}
	\begin{center}\footnotesize{
		\begin{tabular}{c|c|c|c|c|c|c|c|c|c|c}
			\hline
			& Haze & DCP~\cite{he2011singlecvpr}  & FVR~\cite{tarel2009fast} & BCCR~\cite{meng2013efficient} 
            & GRM~\cite{chen2016robust} & CAP~\cite{zhu2015fast} & NLD~\cite{berman2016non} 
            & DehazeNet~\cite{cai2016dehazenet} & MSCNN~\cite{ren2016single} & AOD-Net~\cite{li2017aod} \\
			\hline
            Relu2\_2 & 0.0558 & 0.0473 & 0.0601 & 0.0593 & 0.0395 & \textcolor{cyan}{0.0380} & 0.0523 & \textcolor{red}{0.0314} & 0.0417 & \textcolor{blue}{0.0394} \\
            \hline
			Relu3\_3 & 0.0814 & 0.0731 & 0.0988 & 0.0885 & \textcolor{blue}{0.0626} & \textcolor{cyan}{0.0617} & 0.0805 & \textcolor{red}{0.0520} & 0.0651 & 0.0634 \\
			\hline
			Relu4\_3 & 0.0205 & 0.0190 & 0.0256 & 0.0217 & \textcolor{blue}{0.0168} & \textcolor{cyan}{0.0165} & 0.0206 & \textcolor{red}{0.0143} & 0.0172 & 0.0186 \\
            \hline
            Relu5\_3 & 0.0264 & \textcolor{cyan}{0.0158} & 0.0280 & 0.0188 & \textcolor{blue}{0.0161} & 0.0195 & 0.0186 & \textcolor{red}{0.0151} & 0.0204 & 0.0173 \\
			\hline
		\end{tabular}}
		\label{tab-p_loss}
	\end{center}
\end{table*}

\begin{table*}[!ht]
	\caption{perceptual loss on HSTS 10 synthetic outdoor images.}
	\begin{center}\footnotesize{
		\begin{tabular}{c|c|c|c|c|c|c|c|c|c|c}
			\hline
			& Haze & DCP~\cite{he2011singlecvpr}  & FVR~\cite{tarel2009fast} & BCCR~\cite{meng2013efficient} 
            & GRM~\cite{chen2016robust} & CAP~\cite{zhu2015fast} & NLD~\cite{berman2016non} 
            & DehazeNet~\cite{cai2016dehazenet} & MSCNN~\cite{ren2016single} & AOD-Net~\cite{li2017aod} \\
			\hline
            Relu2\_2 & 0.0595 & 0.0544 & 0.0593 & 0.0635 & 0.0443 & \textcolor{cyan}{0.0334} & 0.0541 & \textcolor{red}{0.0233} & 0.0452 & \textcolor{blue}{0.0356} \\
            \hline
			Relu3\_3 & 0.0918 & 0.0838 & 0.0973 & 0.0944 & 0.0659 & \textcolor{cyan}{0.0538} & 0.0829 & \textcolor{red}{0.0392} & 0.0728 & \textcolor{blue}{0.0596} \\
			\hline
			Relu4\_3 & 0.0234 & 0.0213 & 0.0274 & 0.0240 & 0.0183 & \textcolor{cyan}{0.0145} & 0.0217 & \textcolor{red}{0.0108} & 0.0264 & \textcolor{blue}{0.0165} \\
            \hline
            Relu5\_3 & 0.0347 & 0.0184 & 0.0320 & 0.0207 & 0.0196 & \textcolor{blue}{0.0181} & 0.0213 & \textcolor{red}{0.0122} & 0.0192 & \textcolor{cyan}{0.0178} \\
			\hline
		\end{tabular}}
		\label{tab-p_loss_hsts}
	\end{center}
\end{table*}

\begin{table}
    \caption{Detailed classes information of RTTS.}
    \begin{center}
       \begin{tabular}{c|c|c|c|c|c|c}
          \hline
          Category & \textit{person} & \textit{bicycle} & \textit{car} & \textit{bus} & \textit{motorbike} & Total \\
          \hline
          Normal & 7,950 & 534 & 18,413 & 1,838 & 862 & 29,597 \\
          \hline
          Difficult & 3,416 & 164 & 6,904 & 752 & 370 & 11,606 \\
          \hline
          Total & 11,366 & 698 & 25,317 & 2,590 & 1,232 & 41,203 \\
          \hline
       \end{tabular}
       \label{tab-rtts-details}
    \end{center}
\end{table}

\subsubsection{No-Reference Task-driven Comparison on RTTS} For real-world images without ground-truth, following \cite{li2017aod}, we adopt a task-driven evaluation scheme for dehazing algorithms, by studying the object detection performance on their dehazed results. Specially, we used several state-of-the-art pre-trained object detection models, including Faster R-CNN (FRCNN)~\cite{NIPS2015_5638}, YOLO-V2~\cite{redmon2017yolo9000}, SSD-300 and SSD-512~\cite{liu2016ssd}\footnote{Here we use \href{https://github.com/rbgirshick/py-faster-rcnn}{py-faster-rcnn} and its model is trained on VOC2007\_trainval, while official implementations are used for \href{https://github.com/pjreddie/darknet}{YOLO-V2} and \href{https://github.com/weiliu89/caffe/tree/ssd}{SSDs} and their models are trained on both VOC2007\_trainval and VOC2012\_trainval}, to detect objects of interests from the dehazed images, and rank all algorithms via the mean Average Precision (mAP) results achieved.

For that purpose, we collect a \textit{Real-world Task-driven Testing Set} (RTTS), consisting of 4, 322 real-world hazy images crawled from the web, covering mostly traffic and driving scenarios. Each image is annotated with object categories and bounding boxes, and RTTS is organized in the same form as VOC2007~\cite{pascal-voc-2007}. We currently focus on five traffic-related categories: car, bicycle, motorbike, person, bus. We obtain  41, 203 annotated bounding boxes, 11, 606 of which are marked as ``difficult'' and not used in this paper's experiments. The class details of RTTS are shown in Table \ref{tab-rtts-details}. Additionally, we also collect 4,807 unannotated real-world hazy images, which are not exploited in this paper, but may potentially be used for domain adaption in future, etc. The RTTS set is the largest annotated set of its kind.

\begin{figure*}[t]
    \centering    
    \begin{minipage}{0.15\textwidth}
    	\centering
        \subfigure[Ground Truth] {
        	\includegraphics[width=1.0\textwidth,height=1.5in]{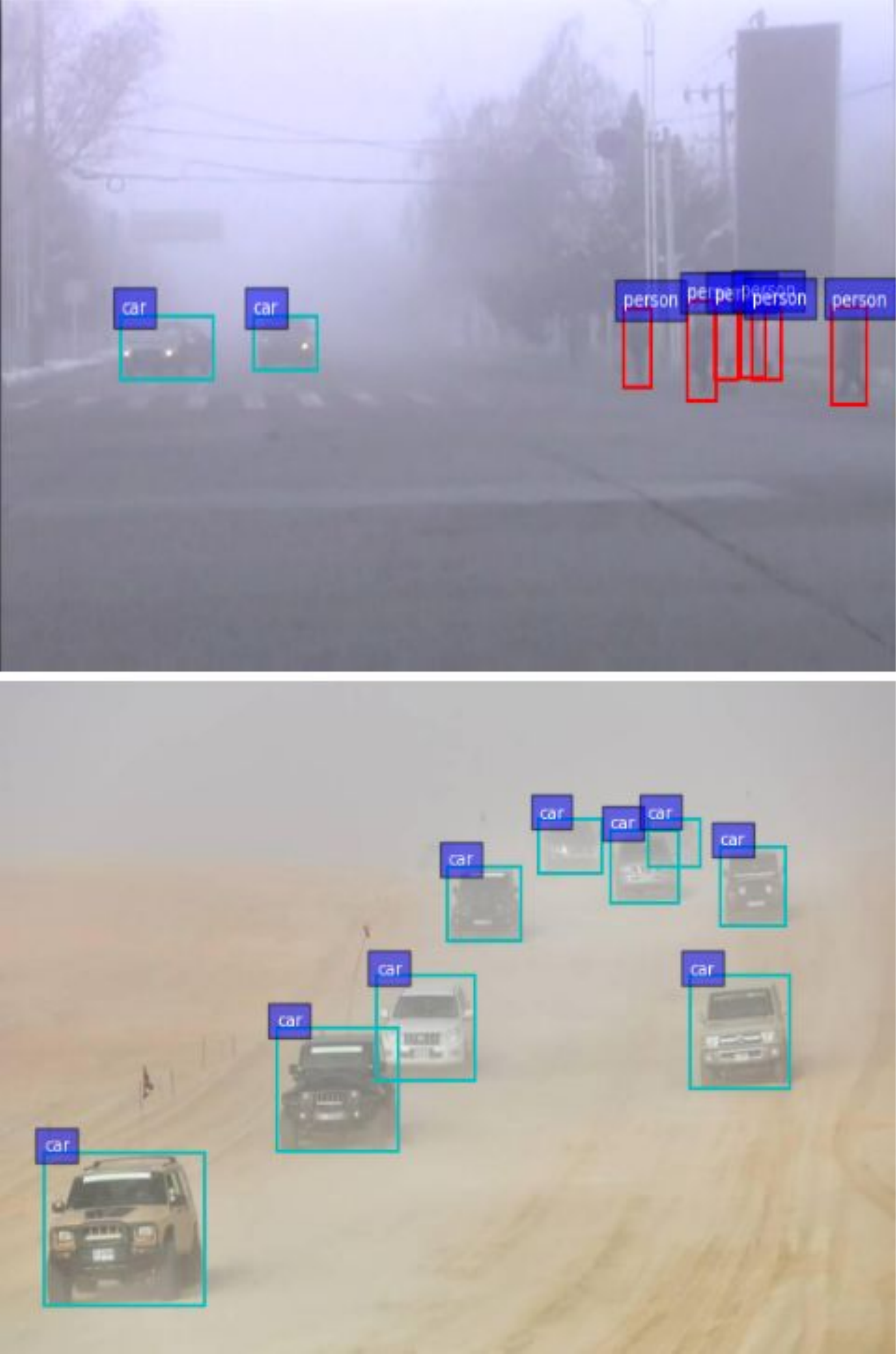}}     
    \end{minipage}
    \begin{minipage}{0.15\textwidth}
        \centering 
        \subfigure[RawHaze] {
            \includegraphics[width=\textwidth,height=1.5in]{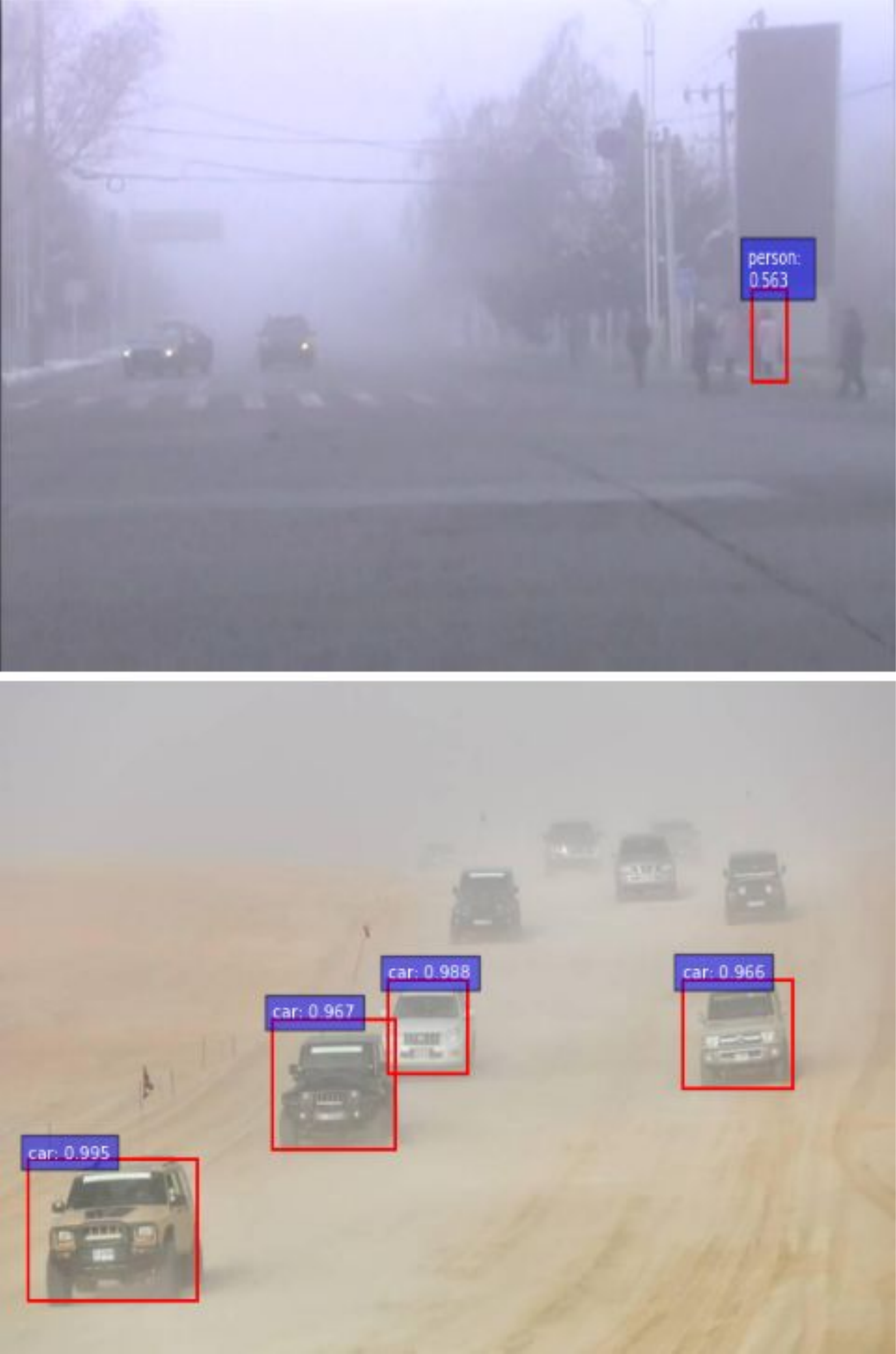}}\\
        \subfigure[DCP] {
            \includegraphics[width=\textwidth,height=1.5in]{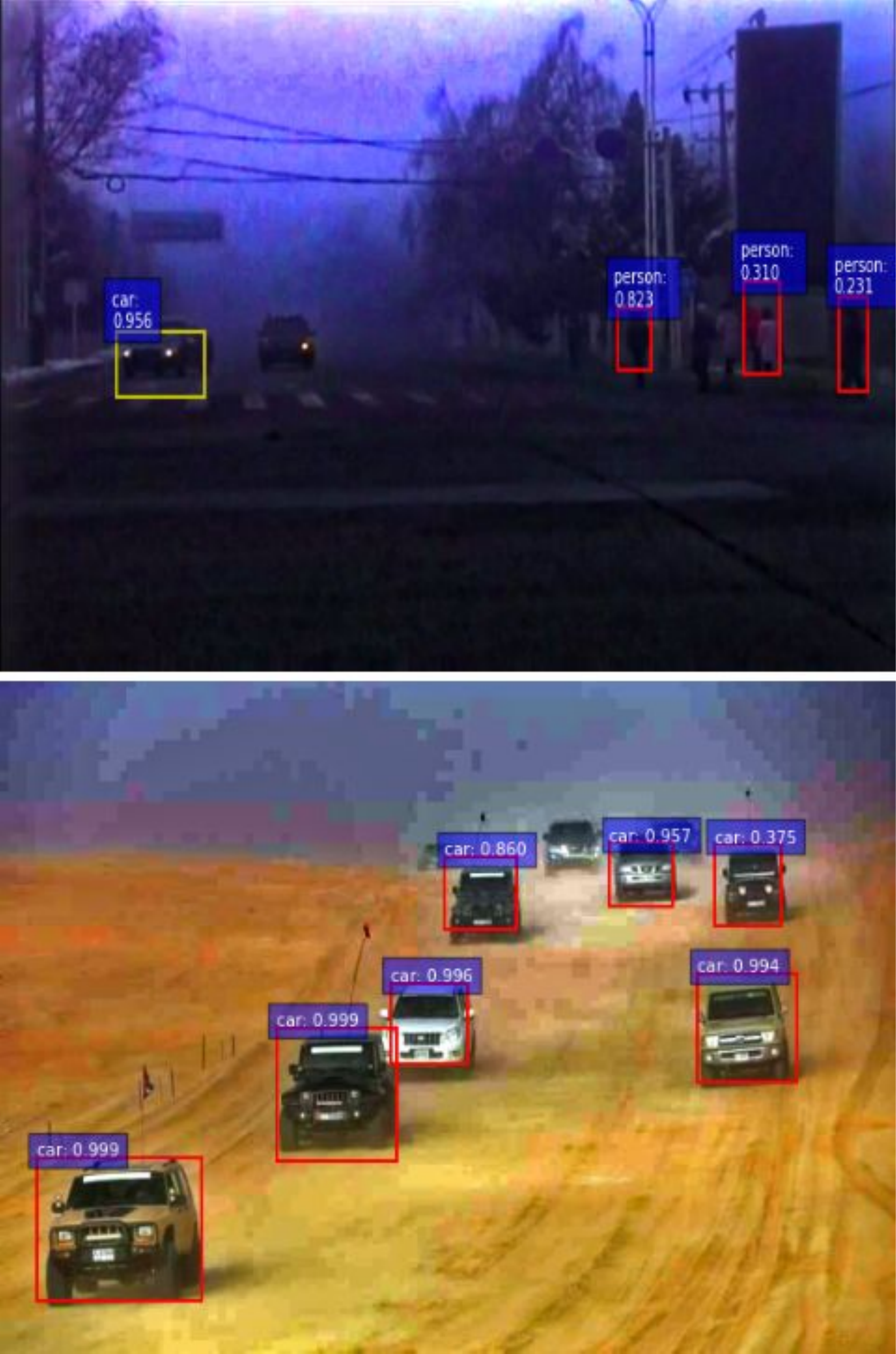}}
    \end{minipage}
    \begin{minipage}{0.15\textwidth}
        \centering 
        \subfigure[FVR] {
            \includegraphics[width=\textwidth,height=1.5in]{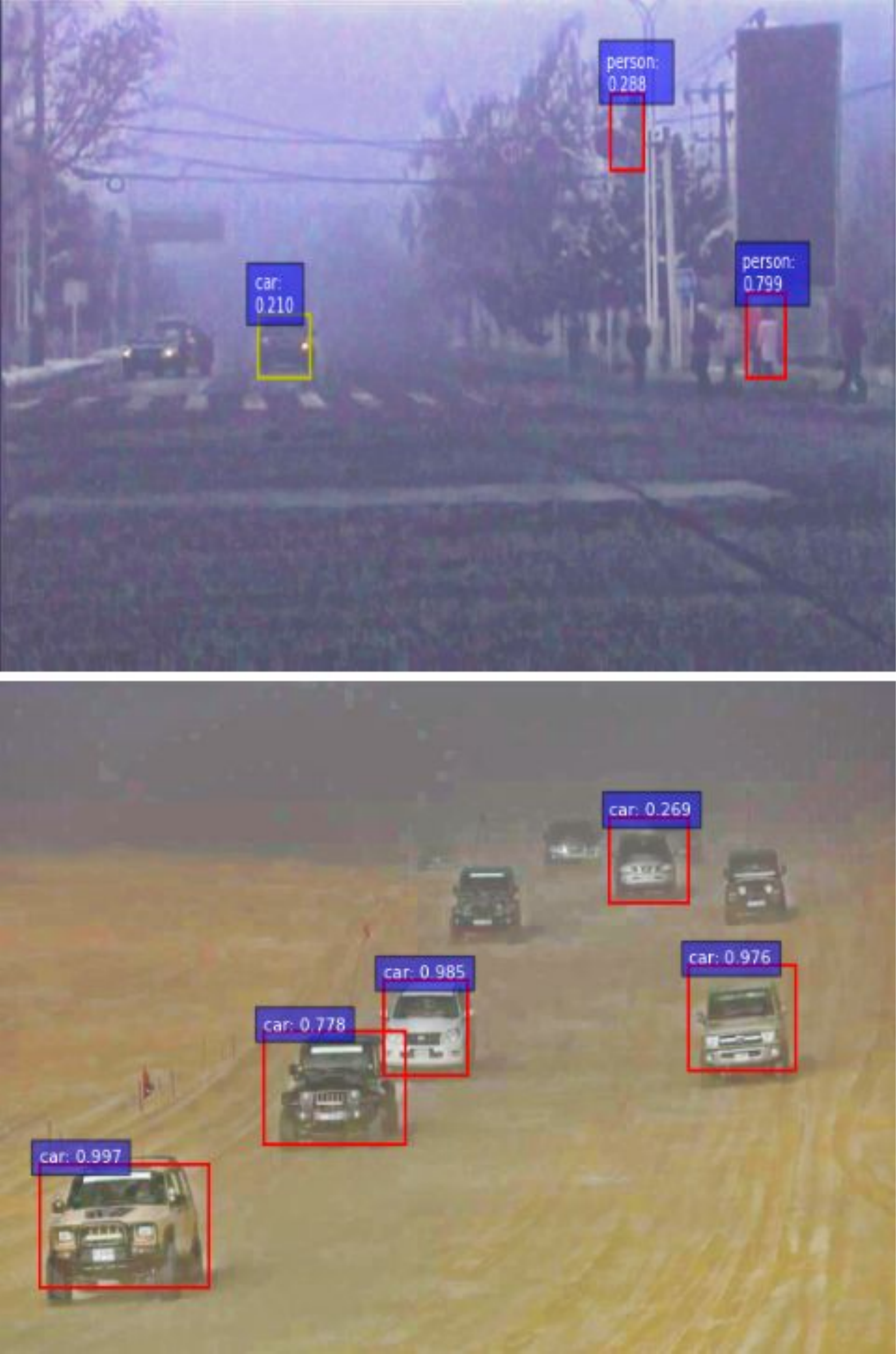}}\\
        \subfigure[BCCR] {
            \includegraphics[width=\textwidth,height=1.5in]{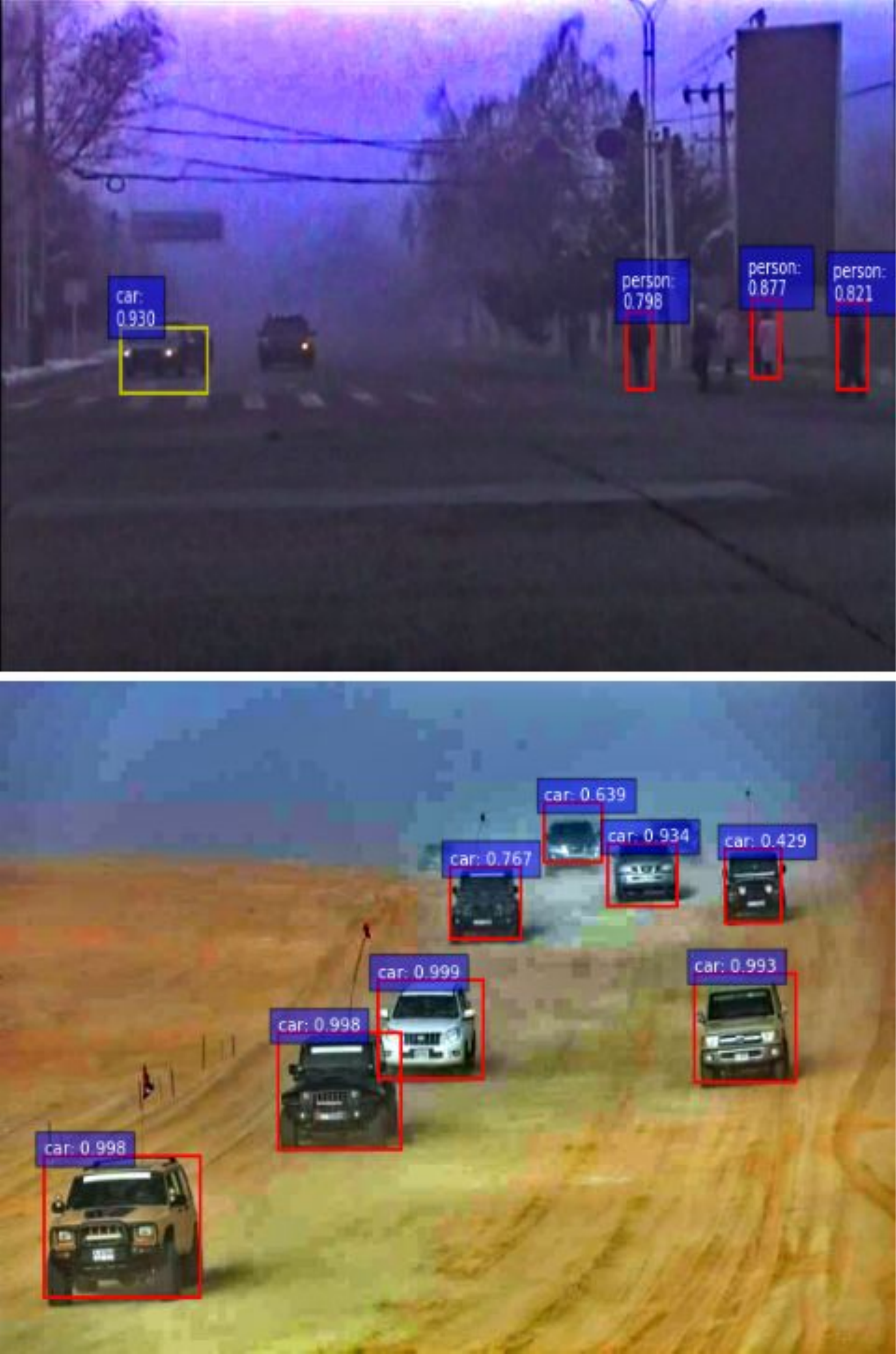}}
    \end{minipage}
    \begin{minipage}{0.15\textwidth}
        \centering 
        \subfigure[GRM] {
            \includegraphics[width=\textwidth,height=1.5in]{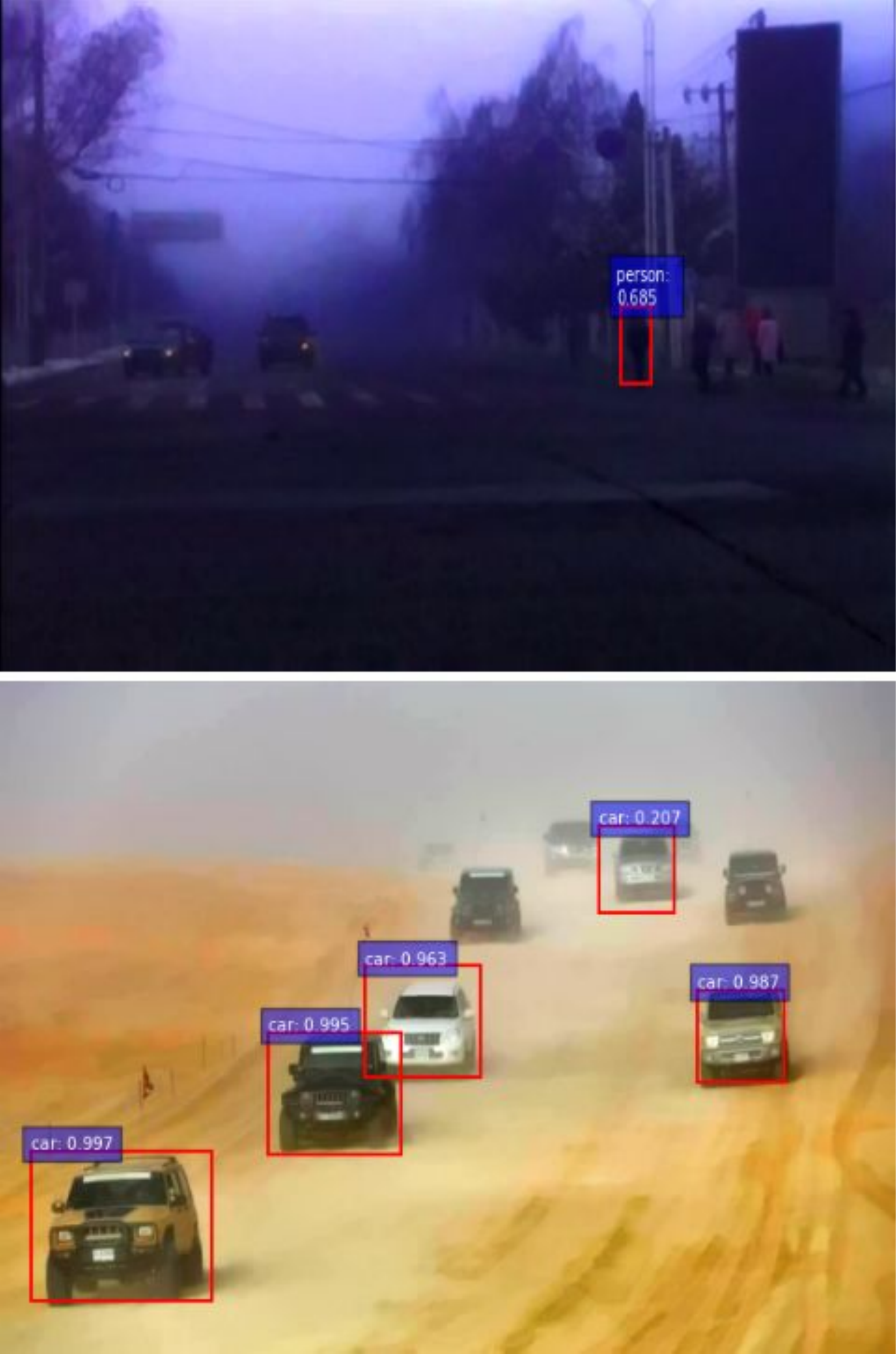}}\\
        \subfigure[CAP] {
            \includegraphics[width=\textwidth,height=1.5in]{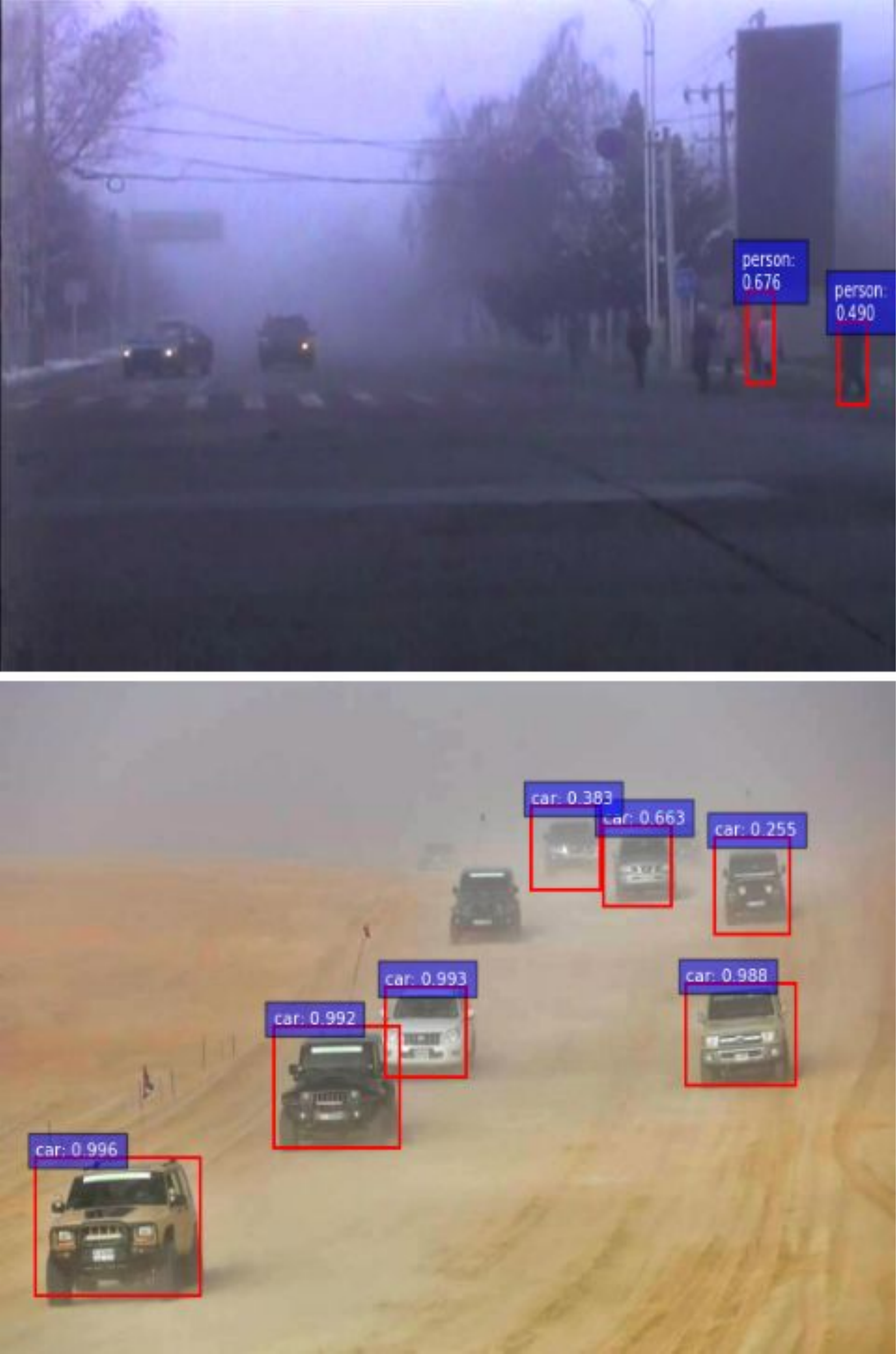}}
    \end{minipage}
    \begin{minipage}{0.15\textwidth}
        \centering 
        \subfigure[NLD] {
            \includegraphics[width=\textwidth,height=1.5in]{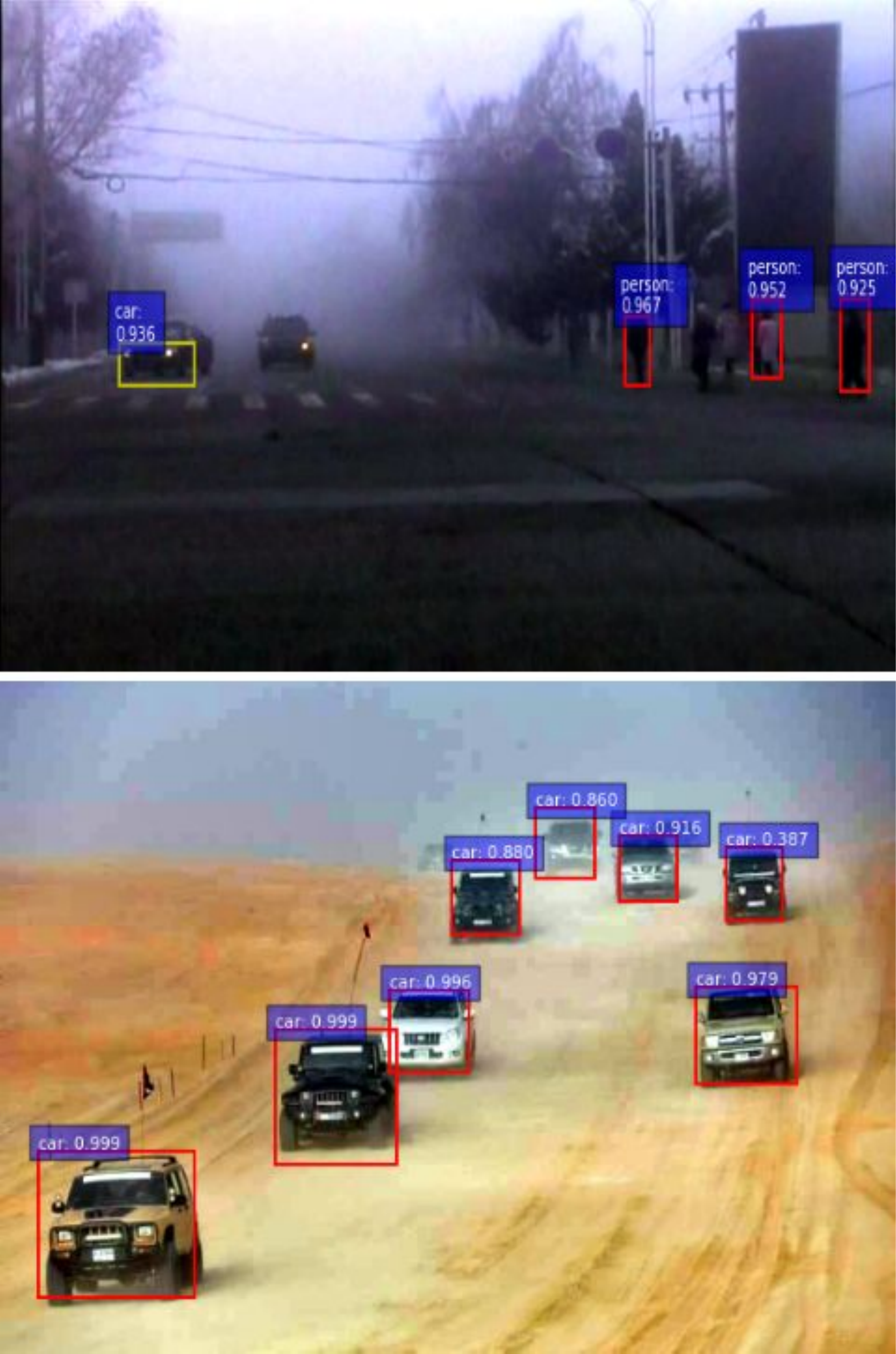}}\\
        \subfigure[DehazeNet] {
            \includegraphics[width=\textwidth,height=1.5in]{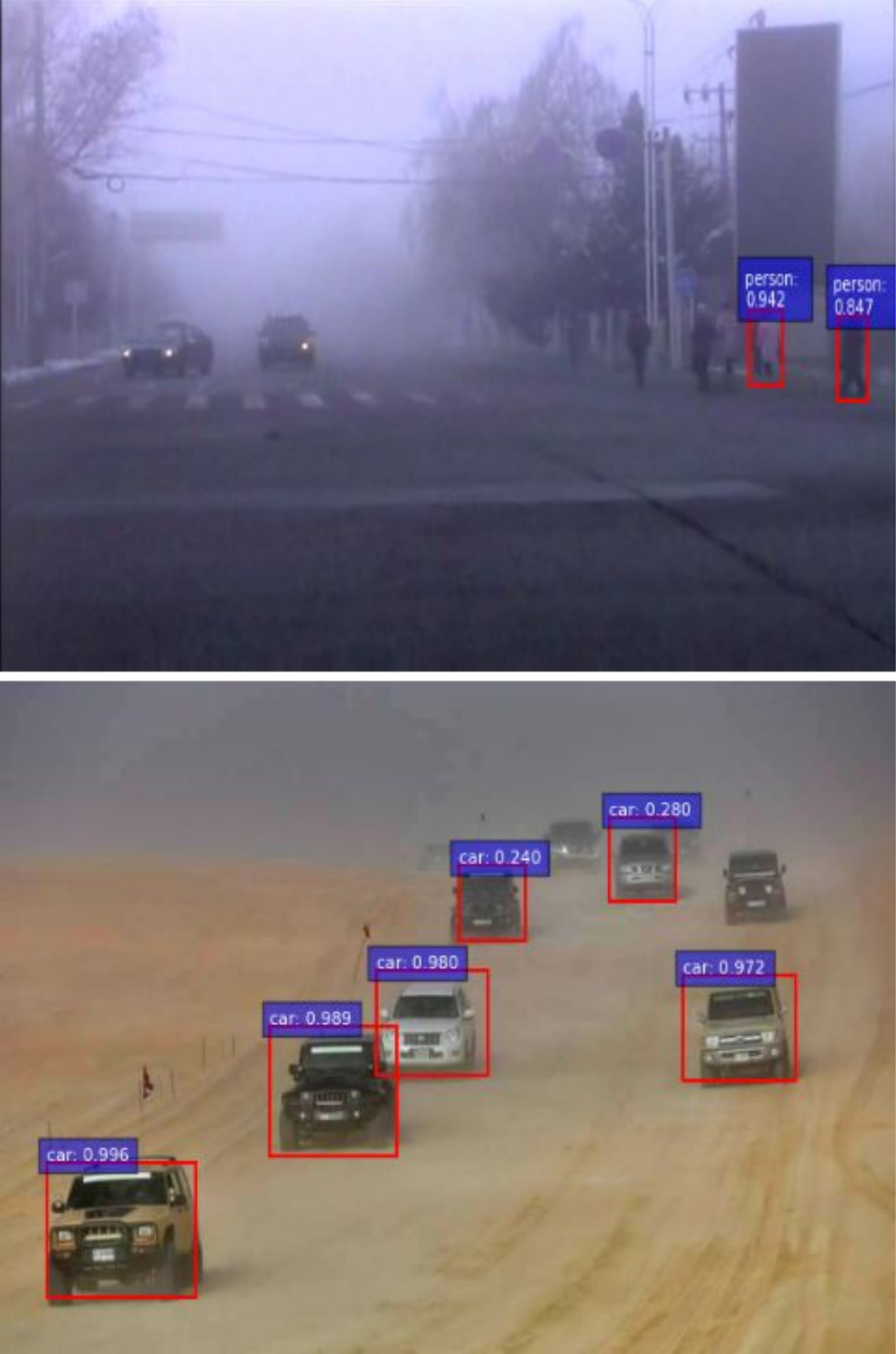}}
    \end{minipage}
    \begin{minipage}{0.15\textwidth}
        \centering 
        \subfigure[MSCNN] {
            \includegraphics[width=\textwidth,height=1.5in]{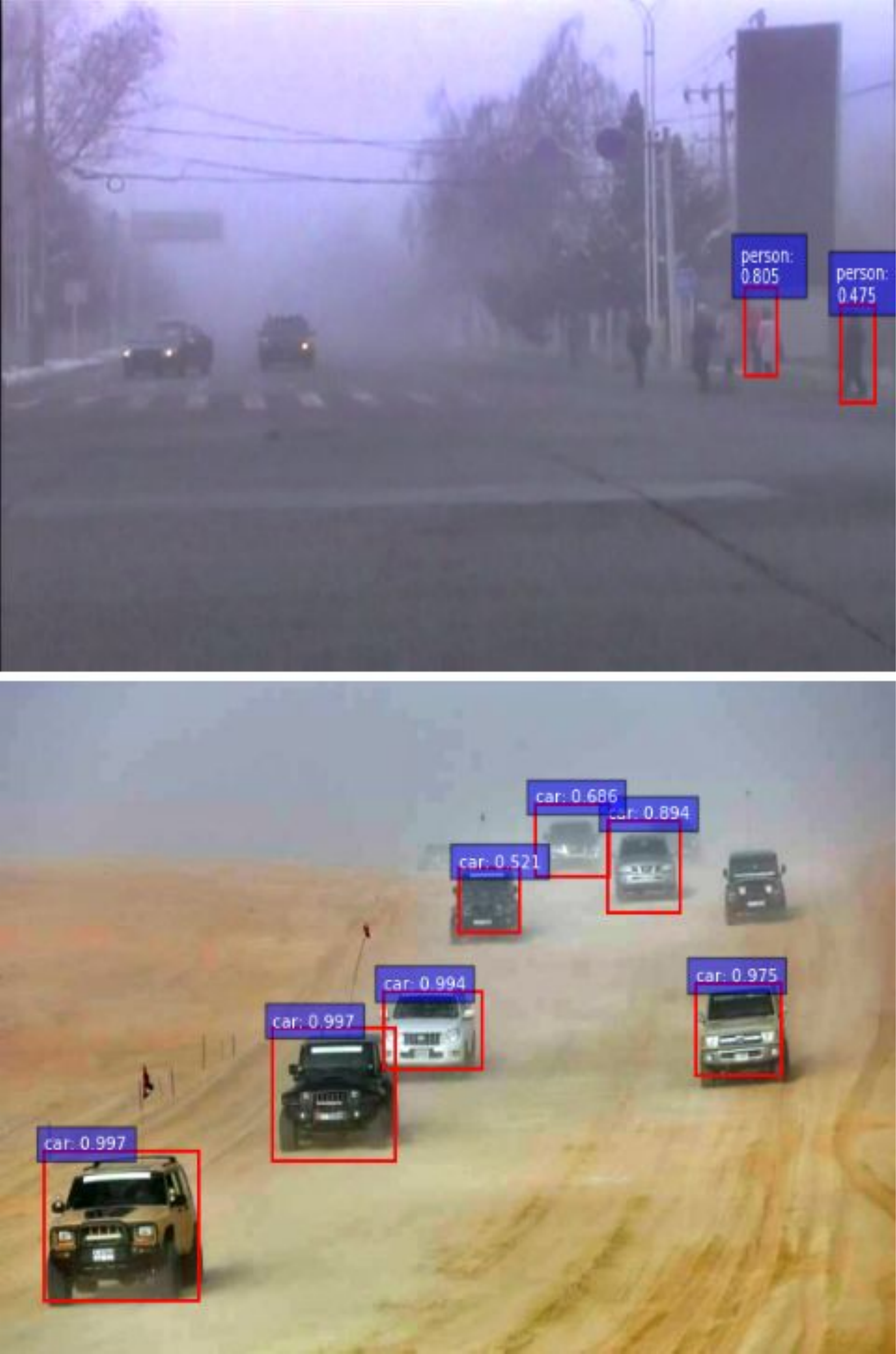}}\\
        \subfigure[AOD-Net] {
            \includegraphics[width=\textwidth,height=1.5in]{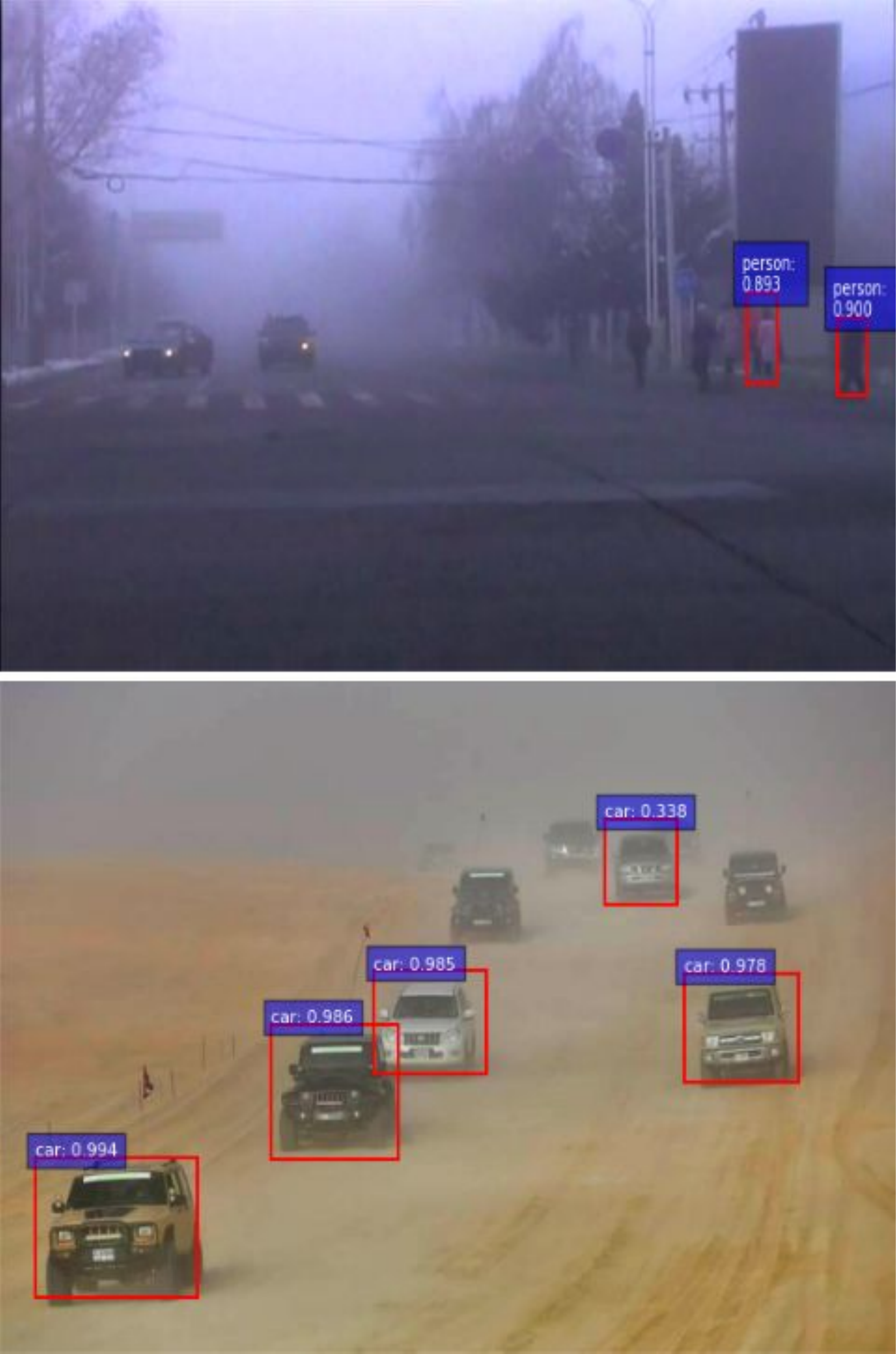}}
    \end{minipage}
    \caption{Visualization of two RTTS images' object detection results after applying different dehazing algorithms.}
    \label{fig:demo_det}
\end{figure*}

Table~\ref{tab-det} compares all mAP results\footnote{For FVR, only 3,966 images are counted, since for the remaining 356 FVR fails to provide any reasonable result.}. The results are not perfectly consistent among four different detection models, the overall tendency clearly shows that MSCNN, BCCR, and DCP are the top-3 choices that are most favored by detection tasks on RTTS. If comparing the ranking of detection mAP with the no-reference results on the same set (see Table \ref{tab-no-ref}), we can again only observe a weak correlation. For example, BCCR~\cite{meng2013efficient} achieves highest BLIINDS-II value, but MSCNN has lower SSEQ and BLIINDS-II scores than most competitors. We further notice that MSCNN also achieved the best clearness and authenticity on HSTS real-world images (see Table \ref{tab-clearness_authenticity}). Figure~\ref{fig:demo_det} display the object detection results using FRCNN on an RTTS hazy image and after applying nine different dehazing algorithms.

\subsubsection{Discussion: Optimizing Detection Performance in Haze?} \cite{li2017aod} for the first time reported the promising performance on detecting objects in the haze, by concatenating and jointly tuning AOD-Net with FRCNN as one unified pipeline, similar to other relevant works \cite{wang2016studying,ACII17,liu2018learning}. The authors trained their detection pipeline using an annotated dataset of synthetic hazy images, generated from VOC2007~\cite{pascal-voc-2007}. Due to the absence of annotated realistic hazy images, they only reported quantitative performance on a separate set of synthetic annotated images. While their goal is different from the scope of RTTS (where a fixed FRCNN is applied on dehazing results for fair comparison), we are interested to explore whether we could further boost the detection mAP on RTTS realistic hazy images using such a joint pipeline. We also point to other recent works utilizing domain adaptation, e.g., \cite{liu2018improved}.

In order for further enhancing the performance of such a dehazing + detection joint pipeline in realistic hazy photos or videos, there are at least two other noteworthy potential options as we can see for future efforts:
\begin{itemize}
\item Developing \textit{photo-realistic} simulation approaches of generating hazy images from clean ones \cite{shrivastava2016learning,li2017photo}. That would resolve the bottleneck of handle-labeling and supply large-scale annotated training data with little mismatch. The technique of haze severity estimation \cite{li2015using} may also help the synthesis, by first estimating the haze level from (unannotated) testing images and then generating training images accordingly. 
\item If we view the synthetic hazy images as the source domain (with abundant labels) and the realistic ones as the target domain (with scarce labels), then the unsupervised domain adaption can be performed to reduce the domain gap in low-level features, by exploiting unannotated realistic hazy images. For example, \cite{wang2015deepfont} provided an example of pre-training the robust low-level CNN filters using unannotated data from both source and target domains, leading to much improved robustness when applied to testing on the target domain data. For this purpose, we have included 4,322 unannotated realistic hazy images in RESIDE that might help build such models. 

\end{itemize}
Apparently, the above discussions can be straightforwardly applied to other high-level vision tasks in uncontrolled outdoor environments (e.g., bad weathers and poor illumination), such as tracking, recognition, semantic segmentation, etc. 

\begin{table*}[!ht]\scriptsize
	\caption{All detection results on RTTS(in \%), Please note that, the model used in FRCNN is trained on VOC2007\_trainval dataset, while the models used in YOLO-V2 and SSDs are trained on VOC2007\_trainval + VOC2012\_trainval.}
	\begin{center}\scriptsize{
		\begin{tabular}{c|c|c|c|c|c|c|c|c|c|c|c}
		\hline
		\multicolumn{2}{c|}{}   
        	& Haze & DCP~\cite{he2011singlecvpr} & FVR~\cite{tarel2009fast} & BCCR~\cite{meng2013efficient} 
        	& GRM~\cite{chen2016robust} & CAP~\cite{zhu2015fast} & NLD~\cite{berman2016non} 
            & DehazeNet~\cite{cai2016dehazenet} & MSCNN~\cite{ren2016single} & AOD~\cite{li2017aod}\\
        \hline
        \multirow{4}{*}{mAP} 
        	& FRCNN~\cite{NIPS2015_5638}        & 37.58 & \textcolor{blue}{40.58} & 35.01 & \textcolor{red}{41.56} & 28.90 & 39.63 & 40.03 & 40.54 & \textcolor{cyan}{41.34} & 37.47 \\
		    & YOLO-V2~\cite{redmon2017yolo9000} & 40.37 & 39.81 & 38.06 & \textcolor{cyan}{40.65} & 29.41 & 39.80 & 39.93 & 40.10 & \textcolor{red}{40.76} & \textcolor{blue}{40.53} \\
		    & SSD-300~\cite{liu2016ssd}         & 50.26 & 49.40 & 47.04 & \textcolor{cyan}{51.57} & 35.59 & \textcolor{blue}{50.31} & 49.84 & 50.14 & \textcolor{red}{51.82} & 49.77 \\
		    & SSD-512~\cite{liu2016ssd}         & 55.55 & \textcolor{blue}{55.71} & 52.29 & \textcolor{red}{57.17} & 39.18 & 55.70 & 54.99 & 55.40 & \textcolor{cyan}{56.88} & 55.29 \\
		\hline
        \multirow{4}{*}{Person}         
        	& FRCNN~\cite{NIPS2015_5638} 		& 60.84 & \textcolor{red}{61.54} & 57.72 & \textcolor{cyan}{64.51} & 50.22 & 61.29 & 60.53 & 61.40 & \textcolor{blue}{61.43} & 61.22 \\
            & YOLO-V2~\cite{redmon2017yolo9000} & 61.24 & 61.14 & 60.00 & 61.16 & 50.13 & \textcolor{cyan}{61.24} & 60.49 & 61.16 & \textcolor{red}{61.30} & \textcolor{blue}{61.20} \\
            & SSD-300~\cite{liu2016ssd}     	& 68.60 & 68.18 & 66.36 & \textcolor{cyan}{69.12} & 53.91 & \textcolor{blue}{68.78} & 66.96 & 68.18 & \textcolor{red}{69.20} & 68.28 \\
            & SSD-512~\cite{liu2016ssd}     	& 72.58 & \textcolor{blue}{72.72} & 69.45 & \textcolor{red}{73.34} & 56.74 & 72.50 & 71.20 & 72.34 & \textcolor{cyan}{73.13} & 72.62 \\
		\hline
        \multirow{4}{*}{Bicycle}        
        	& FRCNN~\cite{NIPS2015_5638} 		& 40.72 & \textcolor{blue}{40.77} & 38.76 & \textcolor{red}{44.57} & 30.71 & 40.48 & 40.21 & 40.68 & \textcolor{cyan}{41.69} & 40.33 \\
            & YOLO-V2~\cite{redmon2017yolo9000} & 44.63 & 43.39 & 40.08 & \textcolor{cyan}{43.66} & 28.81 & 42.65 & \textcolor{blue}{43.56} & 42.34 & 43.53 & \textcolor{red}{44.55} \\
            & SSD-300~\cite{liu2016ssd}     	& 54.92 & 51.36 & 49.35 & 53.33 & 34.48 & 53.38 & \textcolor{blue}{53.42} & 53.08 & \textcolor{red}{55.73} & \textcolor{cyan}{54.18} \\
            & SSD-512~\cite{liu2016ssd}     	& 58.45 & 56.70 & 54.57 & \textcolor{cyan}{58.57} & 36.70 & 57.49 & 56.38 & 57.50 & \textcolor{red}{58.76} & \textcolor{blue}{57.91} \\
		\hline
        \multirow{4}{*}{Car}            
        	& FRCNN~\cite{NIPS2015_5638} 		& 35.18 & 42.15 & 34.74 & \textcolor{red}{42.69} & 26.30 & 41.52 & \textcolor{blue}{42.30} & 41.74 & \textcolor{cyan}{42.61} & 35.13 \\
            & YOLO-V2~\cite{redmon2017yolo9000} & 39.39 & 38.93 & 37.22 & \textcolor{cyan}{39.88} & 29.91 & 39.03 & 38.96 & 39.35 & \textcolor{red}{40.00} & \textcolor{blue}{39.49} \\
            & SSD-300~\cite{liu2016ssd}     	& 54.14 & 54.98 & 50.81 & \textcolor{cyan}{56.32} & 40.21 & 55.08 & 54.98 & \textcolor{blue}{55.27} & \textcolor{red}{56.32} & 54.62 \\
            & SSD-512~\cite{liu2016ssd}     	& 63.05 & 64.95 & 61.54 & \textcolor{red}{65.80} & 47.79 & 64.15 & \textcolor{blue}{65.04} & 64.21 & \textcolor{cyan}{65.22} & 64.05 \\
		\hline
        \multirow{4}{*}{Bus}            
        	& FRCNN~\cite{NIPS2015_5638} 		& 20.90 & 24.18 & 19.06 & 24.66 & 14.81 & \textcolor{blue}{24.74} & 23.74 & \textcolor{cyan}{25.20} & \textcolor{red}{25.25} & 20.56 \\
            & YOLO-V2~\cite{redmon2017yolo9000} & 20.57 & 19.34 & \textcolor{blue}{19.42} & \textcolor{red}{20.01} & 12.86 & 18.90 & 18.22 & 19.07 & \textcolor{cyan}{19.63} & 19.09 \\
            & SSD-300~\cite{liu2016ssd}     	& 30.13 & 30.87 & \textcolor{blue}{30.98} & \textcolor{red}{33.70} & 19.72 & 30.90 & 30.43 & 30.86 & \textcolor{cyan}{32.26} & 29.42 \\
            & SSD-512~\cite{liu2016ssd}     	& 34.60 & \textcolor{blue}{36.51} & 33.47 & \textcolor{red}{37.69} & 22.81 & 35.47 & 34.31 & 35.18 & \textcolor{cyan}{37.42} & 34.13 \\
		\hline
        \multirow{4}{*}{Motorbike}      
        	& FRCNN~\cite{NIPS2015_5638} 		& 30.24 & \textcolor{blue}{34.25} & 24.78 & \textcolor{cyan}{34.34} & 22.44 & 30.10 & 33.36 & 33.70 & \textcolor{red}{35.72} & 30.09 \\
            & YOLO-V2~\cite{redmon2017yolo9000} & 37.84 & 36.23 & 33.59 & \textcolor{blue}{38.54} & 25.33 & 37.10 & 38.40 & \textcolor{cyan}{38.59} & \textcolor{red}{39.33} & 38.31 \\
            & SSD-300~\cite{liu2016ssd}     	& 43.48 & 41.61 & 37.72 & \textcolor{cyan}{45.38} & 29.63 & \textcolor{blue}{43.41} & 43.40 & 43.30 & \textcolor{red}{45.60} & 42.35 \\
            & SSD-512~\cite{liu2016ssd}     	& 49.08 & 47.69 & 42.40 & \textcolor{red}{50.46} & 31.85 & \textcolor{blue}{48.89} & 48.04 & 47.79 & \textcolor{cyan}{49.87} & 47.76 \\
		\hline
	\end{tabular}}
    \label{tab-det}
	\end{center}
\end{table*}

\begin{table*}[!ht]
	\caption{Average no-reference metrics of dehazed results on RTTS.}
	\begin{center}\footnotesize{
			\begin{tabular}{c|c|c|c|c|c|c|c|c|c}
				\hline
				& DCP~\cite{he2011singlecvpr}  & FVR~\cite{tarel2009fast} & BCCR~\cite{meng2013efficient} & GRM~\cite{chen2016robust} & CAP~\cite{zhu2015fast} & NLD~\cite{berman2016non} & DehazeNet~\cite{cai2016dehazenet} & MSCNN~\cite{ren2016single} & AOD-Net~\cite{li2017aod}\\
				\hline
				SSEQ &  62.87  &  \textcolor{cyan}{63.59}  &  \textcolor{blue}{63.31} & 58.64  &  60.66  & 59.37 &  60.01 &  62.31  &   \textcolor{red}{65.35} \\
				\hline
				BLIINDS-II & \textcolor{blue}{68.34}  & 67.68 &  \textcolor{red}{74.07}  &  54.54 & 65.15 &  68.32 & 52.54  & 56.59  &  \textcolor{cyan}{71.05} \\
				\hline
		\end{tabular}}
		\label{tab-no-ref}
	\end{center}
\end{table*}

\section{Conclusions and Future Work}
In this paper, we systematically evaluate the state-of-the-arts in single image dehazing. From the results presented, there seems to be no single-best dehazing model for all criteria: AOD-Net and DehazeNet are favored by PSNR and SSIM; DCP, FVR and BCCR are more competitive in terms of no-reference metrics; DehazeNet performs best in terms of perceptual loss; MSCNN shows to have the most appreciated subjective quality and superior detection performance on real hazy images; and AOD-Net is the most efficient among all. The reason why each dehazing method might succeed or fail in each evaluation case is certainly complicated, e.g., depending on the prior it uses or the model's design choices. Some overall remarks and empirical hypotheses made by the authors are:
\begin{itemize}
\item Deep learning methods \cite{cai2016dehazenet,ren2016single,li2017aod}, especially with the end-to-end optimization towards reconstruction loss \cite{li2017aod}, are advantageous under traditional PSNR and SSIM metrics. However, the two metrics do not necessarily reflect human perceptual quality, and those models may not always generalize well on real-world hazy images.
\item Classical prior-based methods \cite{he2011single,tarel2009fast,meng2013efficient} seem to generate results favored more by human perception. That is probably because their priors explicitly emphasized illumination, contrasts, or edge sharpness, to which human eyes are particularly sensitive. On the other hand, the typical MSE loss used in deep learning methods tend to over-smooth visual details in results, which are thus less preferred by human viewers. We refer the readers to a later manuscript \cite{liu2018improved} for more related discussions. 
\item The detection results on RTTS endorse MSCNN \cite{ren2016single} in particular, which is aligned with the current trend in object detection to use multi-scale features \cite{lin2017feature}. 
\end{itemize}

Based on the RESIDE study and its extensions, we see the highly complicated nature of the dehazing problem, in both real-world generalization and evaluation criteria. For future research, we advocate to be evaluate and optimize dehazing algorithms towards more dedicated criteria (e.g., subjective visual quality, or high-level target task performance), rather than solely PSNR/SSIM, which are found to be poorly aligned with other metrics we used. In particular, correlating dehazing with high-level computer vision problems will likely lead to innovative robust computer vision pipelines that will find many immediate applications. Another blank to fill is developing no-reference metrics that are better correlated with human perception, for evaluating dehazing results. That progress will accelerate the needed shift from current full-reference evaluation on only synthetic images, to the more realistic evaluation schemes with no ground truth.

\section*{Acknowledgement}

Wenqi Ren's research work is supported by the National Natural Science Foundation of China No. 61802403, the Open Projects Program of National Laboratory of Pattern Recognition and the CCF-Tencent Open Fund. Dan Feng's research work is supported by the National Natural Science Foundation of China No.U1705261 and No. 61772222; Engineering Research Center of data storage systems and Technology, Ministry of Education, China. Dacheng Tao's research work is supported by Australian Research Council Projects FL-170100117 and DP-180103424. Zhangyang Wang's research work is supported by the National Science Foundation under Grant No. 1755701. We appreciate the support from the authors of \cite{berman2016non,fattal2008single}. We also acknowledge Dr. Changxing Ding, South China University of Technology, for his indispensable support to our data collection and cleaning.

\vspace{+10em}
    
\bibliographystyle{IEEEtran}
\bibliography{egbib}
\end{document}